\documentclass[a4wide]{article}

\usepackage[utf8]{inputenc}
\usepackage{wrapfig,dsfont}
\usepackage[textsize=tiny]{todonotes}
\usepackage{booktabs} 
\usepackage{graphicx}
\usepackage[vlined,linesnumbered,ruled,titlenotnumbered]{algorithm2e}
\usepackage{mathtools}
\usepackage{hyperref}
\usepackage{balance}
\usepackage{xspace}
\usepackage{algpseudocode,booktabs,amsmath}
\usepackage{multirow, rotating}
\usepackage{mwe,subfigure}
\usepackage{color}
\usepackage{blindtext}
\usepackage{amssymb}
\usepackage{amsthm}
\usepackage{rotating}
\usepackage{hyperref}
\usepackage{balance}
\usepackage{tikz}
\usetikzlibrary{shapes,decorations,patterns,positioning}
\usetikzlibrary{plotmarks}
\usepackage{url}
\usepackage{booktabs}
\usepackage{multirow}
\usepackage{verbatim}
\usepackage{lscape}

\DeclareSymbolFont{symbolsC}{U}{pxsyc}{m}{n}
\DeclareMathSymbol{\colonequals}{\mathrel}{symbolsC}{"42}

\newcommand{\R}{\mathbb{R}}
\newcommand{\HYPR}{HYP-2D\xspace}

\newcommand{\ignore}[1]{}
\begin{document} 

\title{Evolutionary Diversity Optimization Using Multi-Objective Indicators}

\author{
Aneta Neumann\\
Optimisation and Logistics\\
School of Computer Science\\
The University of Adelaide\\
Adelaide, Australia
\and 
Wanru Gao\\
Optimisation and Logistics\\
School of Computer Science\\
The University of Adelaide\\
Adelaide, Australia
\and  
\\
Markus Wagner\\
Optimisation and Logistics\\
School of Computer Science\\
The University of Adelaide\\
Adelaide, Australia
 \and
 \\
Frank Neumann\\
Optimisation and Logistics\\
School of Computer Science\\
The University of Adelaide\\
Adelaide, Australia
}

\maketitle
\begin{abstract}
Evolutionary diversity optimization aims to compute a diverse set of solutions where all solutions meet a given quality criterion. With this paper, we bridge the areas of evolutionary diversity optimization and evolutionary multi-objective optimization. We show how popular indicators frequently used in the area of multi-objective optimization can be used for evolutionary diversity optimization.
Our experimental investigations for evolving diverse sets of TSP instances and images according to various features show that two of the most prominent multi-objective indicators, namely the hypervolume indicator and the inverted generational distance, provide excellent results in terms of visualization and various diversity indicators.
\end{abstract}


\section{Introduction}

Evolutionary algorithms have been used for a wide range of optimization problems and to discover novel designs for various engineering problems. Diversity plays a crucial role when designing evolutionary algorithms as it often prevents the algorithms from premature convergence. In recent years, evolutionary diversity optimization has gained increasing attention~\cite{DBLP:conf/gecco/UlrichT11,DBLP:conf/gecco/UlrichBZ10,DBLP:conf/ppsn/GaoNN16,DBLP:conf/gecco/AlexanderKN17,DBLP:journals/corr/abs-1802-05448}. Evolutionary diversity optimization uses an evolutionary algorithm in order to compute a diverse set of solutions that all fulfill given quality criteria. Presenting decision makers with such alternative designs that are all of good quality gives them a variety of design choices and helps to better understand the space of good solutions for the problem at hand.
Related to evolutionary diversity optimization is the concept of novelty search~\cite{DBLP:books/sp/StanleyL15,DBLP:conf/gecco/RisiVHS09}. Here evolutionary algorithms are used to discover new designs without focusing on an objective. The goal of novelty search is to explore designs that are different to the ones previously obtained. 
This paper focuses on evolutionary diversity optimization. We are interested in computing a diverse set of high quality solutions that can be presented to a decision maker.

Arguably, the most prominent area of evolutionary computation where a diverse set of solutions is sought is evolutionary multi-objective optimization~\cite{deb2001a}. Given a set of usually conflicting objective functions, the goal is to compute a set of solutions representing the different trade-offs of the considered functions. Evolutionary algorithms have been widely applied to multi-objective optimization problems and it is one of the key success areas for applying evolutionary algorithms. Over the years, many evolutionary multi-objective algorithms have been developed. Popular algorithms, among many others, are NSGA-II~\cite{DBLP:journals/tec/DebAPM02}, NSGA-III\cite{DBLP:journals/tec/LiDZK15}, MOEA/D~\cite{DBLP:journals/tec/ZhangL07}, and IBEA~\cite{DBLP:conf/ppsn/ZitzlerK04}. Making them applicable to the area of evolutionary diversity optimization provides huge potential for high performing evolutionary diversity optimization approaches.
With this paper, we bridge the areas of evolutionary diversity optimization and evolutionary multi-objective optimization. 
We consider popular indicators from the area of evolutionary multi-objective optimization and show how to make them  applicable in the area of evolutionary diversity optimization.

Ulrich and Thiele~\cite{DBLP:conf/gecco/UlrichT11} have introduced the framework for evolutionary diversity optimization. They studied how to evolve diverse sets of instances for single-objective problems to the underlying search space. Furthermore, this diversity optimization approach has been introduced into multi-objective search~\cite{DBLP:conf/gecco/UlrichBZ10}. In \cite{DBLP:journals/corr/GaoNN15}, an evolutionary diversity optimization process has been introduced to evolve instances of the Traveling Salesperson problem (TSP) based on given problem features. This approach evolves TSP instances that are hard or easy to solve for a given algorithm, and diversity is measured according to a weighted distribution in terms of the differences in feature values. Afterwards, the approach has been adapted in order to create variations of a given image that are close to it but differ in terms of the chosen image features~\cite{DBLP:conf/gecco/AlexanderKN17}. 

An important question that arises when using evolutionary diversity optimization for more than one criterion or feature is how to measure the diversity of a given set of solutions. The weighted contribution approach used in \cite{DBLP:journals/corr/GaoNN15,DBLP:conf/gecco/AlexanderKN17} has the disadvantage that it heavily depends on the chosen weightening of the features and does not distribute that well for two or three dimensions. In \cite{DBLP:journals/corr/abs-1802-05448}, an evolutionary diversity optimization approach has been introduced that aims to minimize the discrepancy of the solution set in the feature space. 
It has been shown that using the star discrepancy as a diversity measure achieves sets of higher diversity than the previous approaches using weighted contributions.

In this paper, we show how to use popular indicators from the area of evolutionary multi-objective optimization for evolutionary diversity optimization.
Indicators play a prominent role in the area of evolutionary multi-objective optimization and are frequently used to assess the quality of solution sets produced by evolutionary multi-objective algorithms~\cite{DBLP:conf/ppsn/ZitzlerK04,DBLP:journals/tec/ZitzlerTLFF03}. Based on the evaluation of this indicator the selection for survival is carried out. We show how to adapt popular indicators in the area of evolutionary multi-objective optimization to evolutionary diversity optimization. We study important indicators such as the hypervolume indicator (HYP), the inverted generational distance (IGD), and the additive epsilon approximation (EPS), and compare them in terms of their ability to lead to high quality and diverse sets of solutions.

We investigate these indicators for the problems of evolving TSP instances and constructing diverse sets of images as already studied in the literature. Our results show that HYP and IGD are well suited for evolutionary diversity optimization. They obtain the best results for the their respective indicator and also obtain sets of solutions of a better discrepancy when comparing them to the discrepancy-based approach given in~\cite{DBLP:journals/corr/abs-1802-05448}. 

The outline of the paper is as follows. 
First, we describe our approach in Section~\ref{sec:ind}. 
Then, in Sections~\ref{sec:images} and~\ref{sec:tsp}, we describe our diversity optimization for two problems: diverse sets of images and diverse sets of TSP instances. 
Finally, we draw some conclusions.

\section{Indicator-based Diversity Optimization}
\label{sec:ind}

Let $I \in X$ be a search point in a given search space $X$, $f\colon X \rightarrow \mathds{R}^d$ a function that assigns to each search point a feature vector and $q \colon X \rightarrow \mathds{R}$ be a function assigning a quality score to each $x \in X$~\cite{DBLP:journals/tcs/BerghammerFN12}. Diversity is defined in terms of a function $D \colon 2^X \rightarrow \mathds{R}$ which measures the diversity of a given set of search points. Considering evolutionary diversity optimization, the goal is to find a set $P=\{I_1, \ldots, I_{\mu}\}$ of $\mu$ solutions maximizing $D$ among all sets of $\mu$ solutions under the condition that $q(I) \geq \alpha$ holds for all $I\in P$, where $\alpha$ is a given quality threshold. Here $\mu$ is the size of the set that we are aiming for, which determines the parent population size in our evolutionary diversity optimization approach.

As already outlined, diversity has been optimized in a few different ways over the years. 
Of particular interest to us is the optimization of diversity in a given set of problem instances. We will use this domain as an application area to demonstrate that the general goal of diversity optimization with respect to multiple features is achievable.

If diversity is sought with respect to a single feature, then the generation of instances can focus on covering the range of values in some fashion. If two or more features are of interest, then covering this space evenly is not straightforward, as a metric is needed to assess the coverage. 

Recently,~\cite{DBLP:journals/corr/abs-1802-05448} have used the mathematical concept of ``discrepancy'' to measure the irregularities of distributions and used this measure for evolutionary diversity optimization. The used star-discrepancy uses axis-parallel boxes: ideally, the number of points inside the box is proportional to the size of the box. The computation of this metric is time consuming ($n^{1+d/2}$~\cite{DobkinEM96}) and the resulting distributions are counter-intuitive.

Here, we propose to use a very well-established concept, i.e., the use of indicators from multi-objective optimization.
In multi-objective optimization, a function $g\colon X \rightarrow \mathds{R}^d$ containing $d$ objectives is given and all objectives should be optimized at the same time. As the given objectives are usually conflicting, one is interested in the trade-offs with respect to the given objective functions.
Indicators in the area of multi-objective optimization have been used for many years to compare sets of solutions in the objective space, either for the purposes of comparing algorithm performance, or for use within an algorithm to drive a diversified search. 
Similarly to the diversity measure $D$ in evolutionary diversity optimization, an indicator $\mathcal{I} \colon 2^X \rightarrow \mathds{R}$ measures the quality of a set of solutions according to some indicator function $\mathcal{I}$. 
The immediate problem with applying multi-objective optimization indicators is that that diversity does not have a notion of dominance. In the context of multi-objective optimization, the optimal solutions are also referred to as non-dominated solutions.
A solution $x$ is called non-dominated (or Pareto optimal) if there is no other solution that is at least as good as $x$ with respect to every objective and better in at least one objective.
As multi-objective approaches aim to compute a set of non-dominated solutions, they reject dominated solutions over time. In evolutionary diversity optimization, every solution meeting the quality criteria is eligible and only the diversity among such solutions matters. Hence, we have to adapt the multi-objective indicators in a way that makes all solutions meeting the quality criterion non-dominated. 
 We do this by ensuring that all solutions are incomparable when applying these indicators. For a more comprehensive introduction to dominance we refer the interested reader to~\cite{Chand2015manyEmo}, which is present in a large number of multi-objective optimization indicators.

In the following, we will first present existing multi-objective optimization indicators and our transformations to deal with the dominance issue. Then, we introduce the generic $(\mu+\lambda)$-$EA_{D}$ and the concrete variants that will form the basis for our subsequent experimental studies on diversity optimization.

\subsection{Multi-objective optimization indicators for diversity optimization}\label{sec:moofordiversity}

In this article, we use three quality indicators evaluating the quality of a given set of  
objective vectors $S$. 
For a given set of search points $P$ (called the population) and a function $g\colon X \rightarrow \mathds{R}^d$, we define $S= \{g(x) \mid x \in P\}$ as the set of objective vectors of $P$.
\begin{itemize}

\item Hypervolume (HYP): HYP is the volume covered by the set of objective vectors $S$ with respect to a given reference point $r$.
The hypervolume indicator measures the volume of the dominated space of all solutions contained in a set $S \subseteq \R^d$. This space is measured with respect to a given reference point $r=\left( r_1, r_2, \ldots, r_d \right)$. The hypervolume $HYP(S,r)$ of a given set of objective vectors $S$ with respect to $r$ is then defined as $$HYP(S,r) = VOL \left( \cup_{(s_1,\ldots,s_d)\in S} \left[r_1,s_1\right] \times \cdots \left[r_d,s_d\right] \right)$$ with $VOL(\cdot)$ being the Lebesgue measure. 

\item Inverted generational distance (IGD): IGD measures $S$ with respect to a given reference set $R$. It calculates the average distance of objective vectors in $R$ to their closest points in $S$.
 We have $$IGD(R,S)=\frac{1}{\left|R\right|} \sum_{r \in R} \min_{s \in S} d(r,s),$$ where $d(r,s)$ is the Euclidean distance between $r$ and $s$ in the objective space. 

\item Additive epsilon approximation (EPS): EPS measures the approximation quality of the worst approximated point in $R$ that $S$ achieves.
For finite sets $S,R \subset \R^d$, the additive approximation of $S$ with respect to $R$ (assuming all objectives are to be minimized) is defined as 
$$\alpha(R,S) := \max_{r \in R} \min_{s \in S} \max_{1 \le i \le d} (s_i - r_i).$$
To get a sensitive indicator that can be used to guide the search, we consider instead the set $\{ \alpha(\{r\},S) \mid r \in R \}$ of all approximations of the points in $R$. We sort this set decreasingly and call the resulting sequence
$S_\alpha(R,S) := (\alpha_1, \ldots, \alpha_{|R|})$ (see~\cite{Wagner2015ageejor}).
\end{itemize}

While other indicators could also be used for driving diversity optimization, we do not intend to highlight differences of the indicators (which has been subject to many papers), but instead we will focus on demonstrating that they can in-fact be used as a tool out-of-the-box to explore the space of combinations of instance features.

\begin{figure*}[t]
\centering
\includegraphics[trim={0 0 0 0},clip,width=0.523\textwidth]{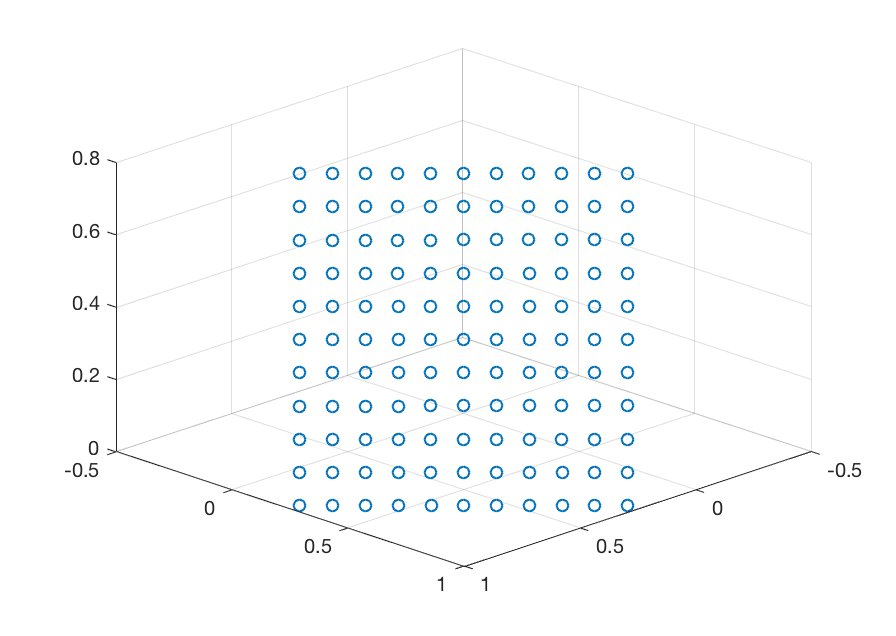}%
\includegraphics[trim={0 0 0 0},clip,width=0.523\textwidth]{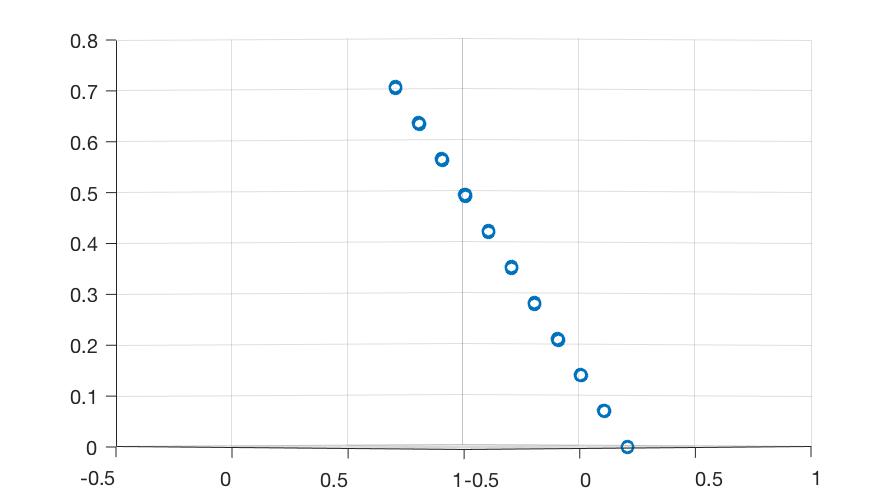}

\caption{Reference set in 3D using $11^2$ objective vectors. The normal vector that goes through the centre of the square goes through the origin. We use $101^2$ feature vectors in our experiments. 
}
\label{fig:refset}
\end{figure*}

These three indicators cannot be applied immediately, as there is no reference set (which some indicators require) and one has to deal with the issue of dominance as there is no preference of one feature value over the other. For example, let us consider two scaled features and visualize the combinations as points in a two-dimensional unit square. In this case, we would like to cover the entire square evenly, without preferring one region over the other, and in particular we cannot say that one area is preferred over another -- a naive multi-objective optimization setup for this two-dimensional problem might focus, for example, only on the area near the origin. 

We propose two approaches to deal with this challenge: (1) transformation of the two-dimensional problem into a three-dimensional problem, (2) doubling the number of dimensions.

\subsubsection{Problem Transformation}

When we are interested in covering a two-dimensional feature space, we can mitigate the problem of EPS-/HYP-preferred regions by transforming the two-dimensional problem into a three-dimensional one. We do so as follows:

\begin{enumerate}
\item We place the unit square with its original x/y-coordinates in the three-dimensional space using $z=0$.
\item We rotate it around the $x$ and $y$ axis by 45 degrees each time.
\item We translate it such that the center point of the transformed unit square is at $(sqrt(2)/4)^3$ (see Figure~\ref{fig:refset}). 
\end{enumerate}

After these steps, the normal vector that goes through the center of the unit square also goes through the origin. Note for the rotation, we use Java 1.8's method \textit{java.awt.geom.AffineTransform.getRotateInstance(...)}. This orientation allows us to use the wide spectrum of well-established quality indicators from the field of multi-objective optimization, designed for assessing various aspects of solutions sets, such as convergence and distribution -- and no modifications are needed at all. 
Especially for the volume- and dominance-based indicators our transformation has the important benefit that all features are of equal importance.

As we perform the same transformation with the instance set (i.e., our population) as well as the reference set (after rescaling it into the unit square based on known lower and upper values for the features), this means that the population is always on the Pareto front; this is a situation that is not that common in multi-objective optimization. Our goal is now to cover the reference set ``evenly'', as defined by the respective indicators.

\subsubsection{Dimension doubling}

\begin{figure}
\centering
\includegraphics[width=45mm]{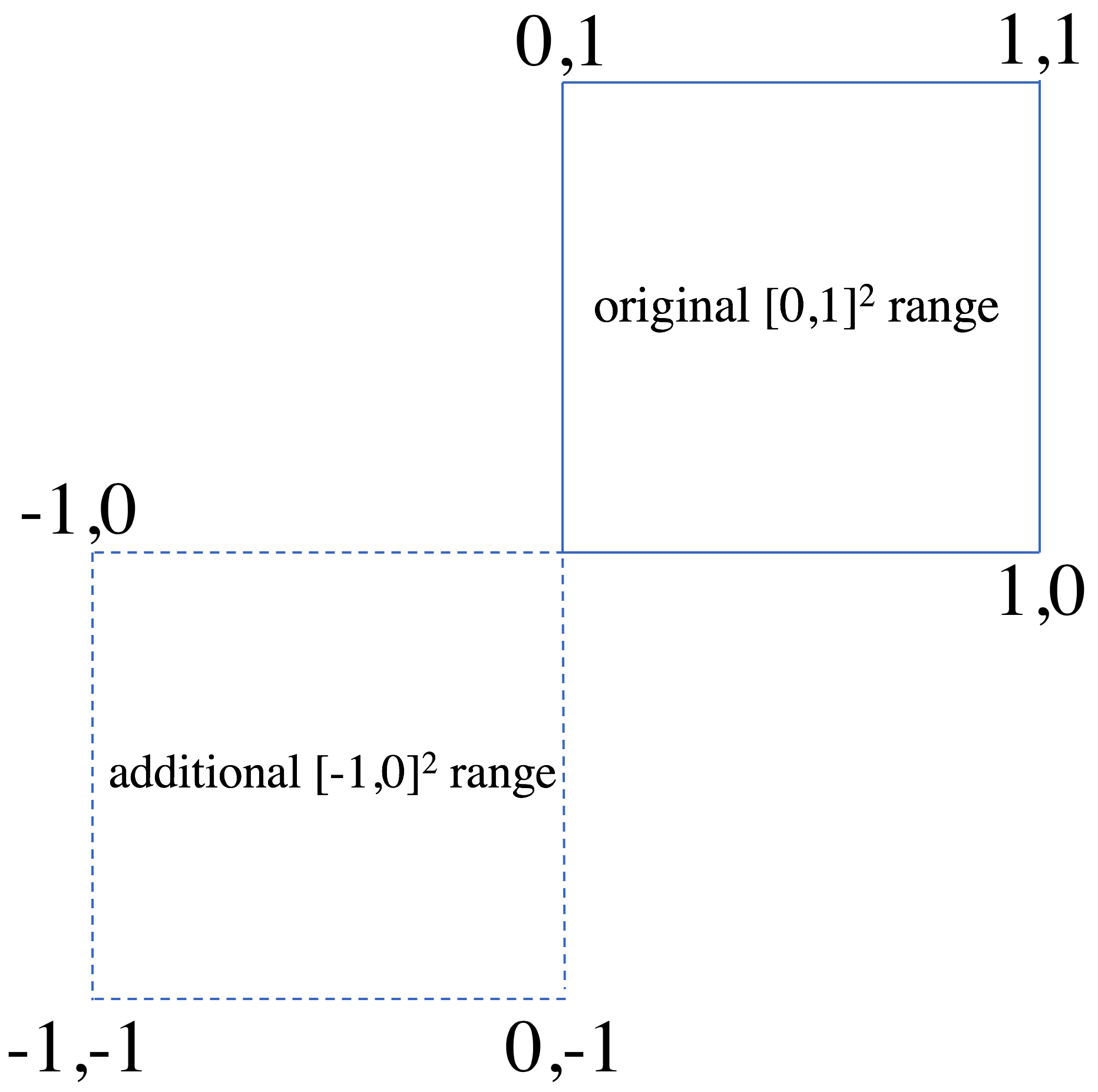}
\caption{Visualisation of the 2d-dimensional space.
}
\label{plot:dimensiondoubling}
\end{figure}

To avoid the dominance issue, we propose the following transformation. Given a feature vector $p=(p_1, p_2, \ldots, p_d)$ in the $d$-dimensional space, we project it into the $2d$-dimensional space by copying the original feature values and negating their copy, resulting in 
$$p'=(p_1, p_2, \ldots, p_d, -p_1, -p_2, \ldots,-p_d),$$ 
see Figure~\ref{plot:dimensiondoubling}. 
With this, dominance between solution vectors vanishes, and we can employ the hypervolume indicator without the need for any modifications. 

Because we work with rescaled value ranges in $\left[0,1\right]^d$, the necessary hypervolume reference point $r$ has to be adequately chosen in the 2d-dimensional space. For example. ($1^d$,$0^d$) would be based on the ranges' extreme values, and ($2^d$,$1^d$) would put an increased focus on maintaining extreme points in the population.

While this transformation mitigates the dominance issue, it remains an open problem how this can be made to work with the epsilon indicator as well. The challenge here is to define an evenly spread out reference set in the 2d-dimensional space given our dimension doubling.


\subsection{Evolutionary algorithm for optimizing diversity}

The algorithm used to optimize the feature-based population diversity follows the setting in~\cite{DBLP:conf/ppsn/GaoNN16} with modifications. 
Algorithm~\ref{EA} shows the evolutionary algorithm used for optimizing diversity. Let $I \in P$ be an individual in a population $P$. A problem specific feature vector $f(I) = (f_1(I), \ldots, f_d(I))$ is used to describe a potential solution. The indicators are calculated based on the feature vector.

Since the indicators introduced are defined in the space of $[0,1]^d$, the feature values are scaled before the calculation of indicators. 
Let $f_i^{\max}$ and $f_i^{\min}$ be the maximum and minimum value of a certain feature $f_i$ obtained from some initial experiments. 
The feature values are normalized based on the formula $$f_i'(I) = (f_i(I)-f_i^{\min})/(f_i^{\max}-f_i^{\min}).$$
Feature values outside the range $[f_i^{\min},f_i^{\max}]$ are set to $0$ or $1$, to allow the algorithm to work with non-anticipated features values.

\begin{algorithm}[!t]
\vspace{0.5mm}
{
   	Initialize the population $P$ with $\mu$ instances of quality at least $\alpha$.\\
	Let $C \subseteq P$ where $|C| = \lambda$.\\
	For each $I \in C$, produce an offspring $I'$ of $I$ by mutation. If $q(I')\geqslant\alpha$, add $I'$ to $P$. \\
	While $|P| > \mu$, remove an individual with the smallest loss to the diversity indicator $D$.\\ 
 	Repeat step 2 to 4 until termination criterion is reached.\\
} 
\vspace{0.5mm}
 \caption{$(\mu+\lambda)$-$EA_{D}$}
\label{EA}
\end{algorithm}

Based on this, we investigate the following diversity-optimizing algorithms in this study:
\begin{itemize}
\item EA$_{\text{\HYPR}}$ and EA$_{\text{EPS}}$ use the idea of transforming the two-dimensional problem into a three-dimensional one.
\item EA$_{\text{HYP}}$ uses the idea of doubling the dimensions.
\item EA$_{\text{IGD}}$ uses IGD, which can be used without the need to transform the feature vectors, as it does not consider concepts like dominance or volume like HYP and EPS.
\end{itemize}

In addition, we use EA$_{\text{DIS}}$ with discrepancy minimization, as used in \cite{DBLP:journals/corr/abs-1802-05448}.
As IGD and EPS require a reference set (e.g. solutions situated on the Pareto front), we use regular grids in the unit square and unit cube with a resolution of $101^2$ solutions and $11^3$ solutions. The necessary hypervolume reference point $r$ for EA$_{\text{\HYPR}}$ is set based on the extreme values of the reference set after the described rotations; for EA$_{\text{HYP}}$ it is set to ($2^d$,$1^d$) to increase the focus on extreme points.


\section{Images}
\label{sec:images}

\begin{figure}[!t]
\centering
\includegraphics[width=35mm]{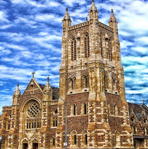}
\caption{Image $I^*$.}
\label{fig:ImageS}
\end{figure}

In this section, we aim to evolve a diverse set of images as described in~\cite{DBLP:conf/gecco/AlexanderKN17}. Given an image $I^*$, we want to compute
a diverse set of images $P=\{I_1, \ldots, I_{\mu}\}$ that agree on a given quality criteria $q(I)$ for each $I \in P$. 
We will use the image $I^*$ given in Figure~\ref{fig:ImageS} for our investigations.
An image $I$ fulfills the quality criteria $q(I)$ if the \emph{mean-squared error} in terms of the RGB-values of $I$ with respect to $I^*$ is less than 500.

Many different features have been widely applied to measurements of the properties of images. They often provide a good characterization of images. We select the set of features identified in~\cite{DBLP:conf/gecco/AlexanderKN17}. We carry out the indicator-based evolutionary optimization approach with respect to different multi-objective indicators and different sets of features. Our evolutionary algorithm evolves diverse populations of images for each indicator and for each feature combination.

In our experiments we used the following features:
standard-deviation-hue, mean-saturation, reflectional symmetry~\cite{den2014investigating}, hue~\cite{hughes2014computer}, Global Contrast Factor~\cite{matkovic2005global}, and  smoothness~\cite{Nixon:2008:FEI:1571711}.
\begin{table}[!t]
\centering
\vspace{1.8mm}
 \renewcommand{\arraystretch}{1.3}
\begin{tabular}{lllll}
\hline
      & Notation                                  & $f^{min}$ & $f^{max}$ & Description            \\ \hline
$f_1$ & SDHue                                     & 0.420 & 0.700 & standard deviation hue \\
$f_2$ & saturation  & 0.420 & 0.500 & mean saturation        \\
$f_3$ & symmetry                                       & 0.715 & 0.740 & reflectional symmetry\\
$f_4$ & hue                                       & 0.250 & 0.400 &  color descriptor \\
$f_5$ & GCF                                       & 0.024 & 0.027 & Global Contrast Factor~ \\
$f_6$ & smoothness                                      & 0.906 & 0.918 & smoothness  \\ \hline
\end{tabular} 
\vspace{3mm}
\caption{Description of features for images.}
\label{tab:features}
\end{table}
Instead of applying the star discrepancy~\cite{Thimard2001AnAT} to measure diversity we use the multi-objective indicators as previously introduced. Otherwise, the configuration of Algorithm~\ref{EA} is the same as in~\cite{DBLP:journals/corr/abs-1802-05448}. In order to produce a new solution the algorithm uses a self-adaptive offset random walk mutation introduced in~\cite{DBLP:journals/corr/abs-1802-05448}. Based on a random walk on the image this operator alters the RGB-values of the pixels visited in a slight way such that a new but similar image is obtained. Random walk lengths are increased in the case of a successful mutation and decreased in the case of unsuccessful ones. For details, we refer the reader to~\cite{DBLP:journals/corr/abs-1802-05448,DBLP:conf/gecco/NeumannSCN17}.
\hspace{-0.3cm}
\begin{figure}[!t]

\rotatebox{90}{\hspace{10mm}EA$_{\text{\HYPR}}$} \rotatebox{90}{\hspace{4mm}\rule{26mm}{1pt}}%
\hspace{0.1cm}
\includegraphics[trim={0.5cm 5.7cm 0.7cm 6.8cm},clip,width=0.3\textwidth]
{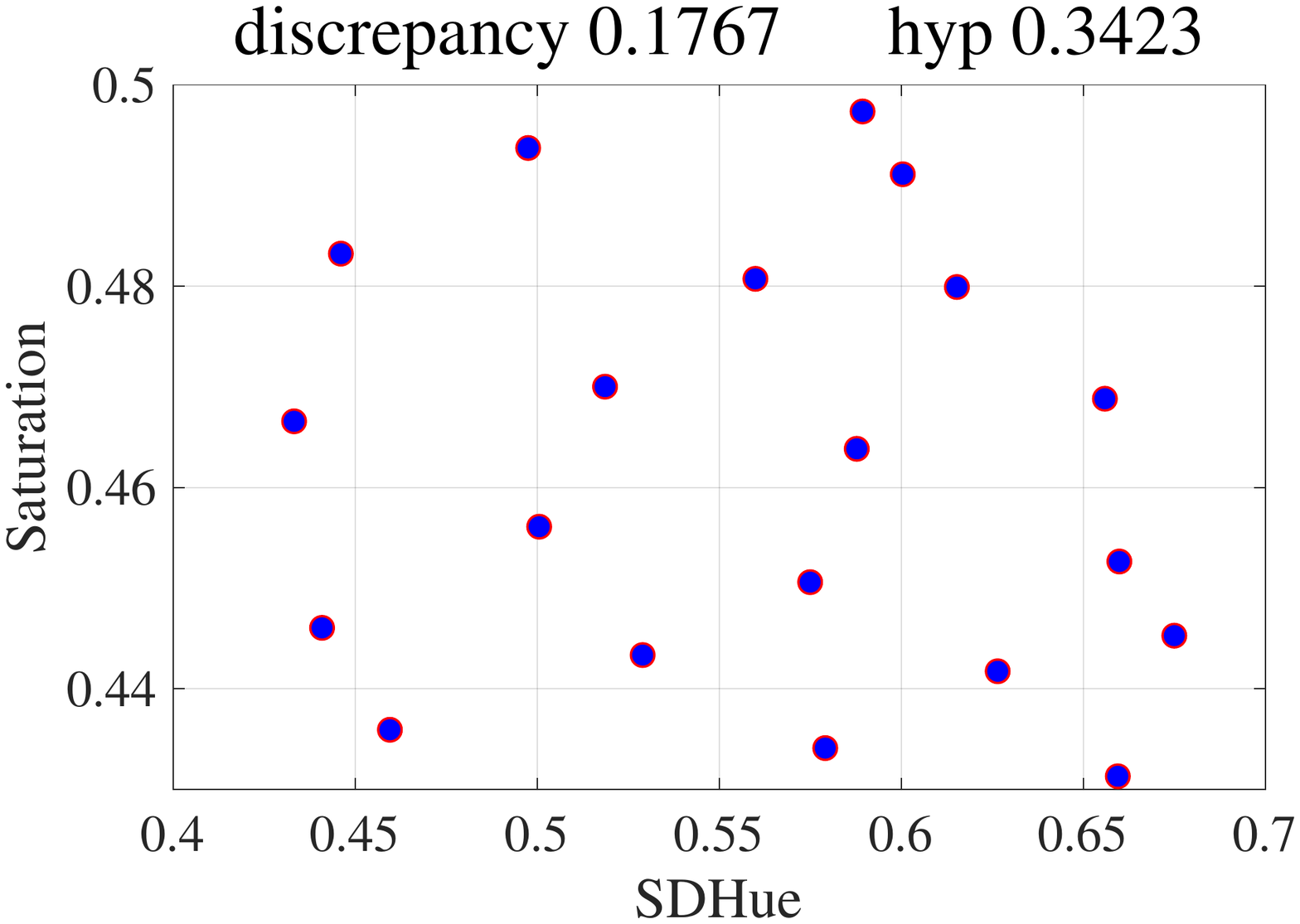}%
\includegraphics[trim={0.5cm 5.7cm 0.7cm 4.8cm},clip,width=0.3\textwidth]
{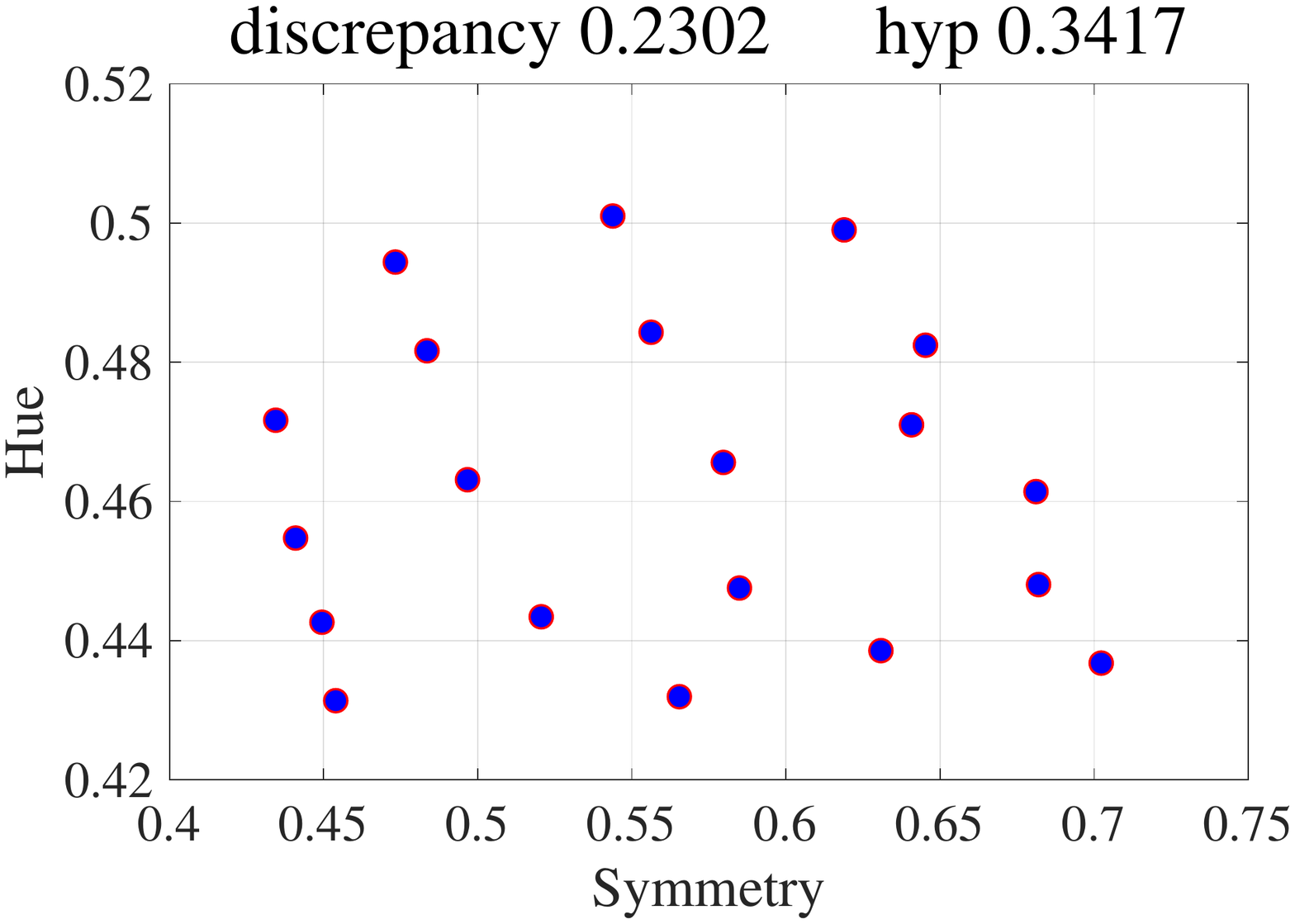}%
\includegraphics[trim={0.45cm 5.7cm 0.08cm 4.8cm},clip,width=0.308\textwidth]
{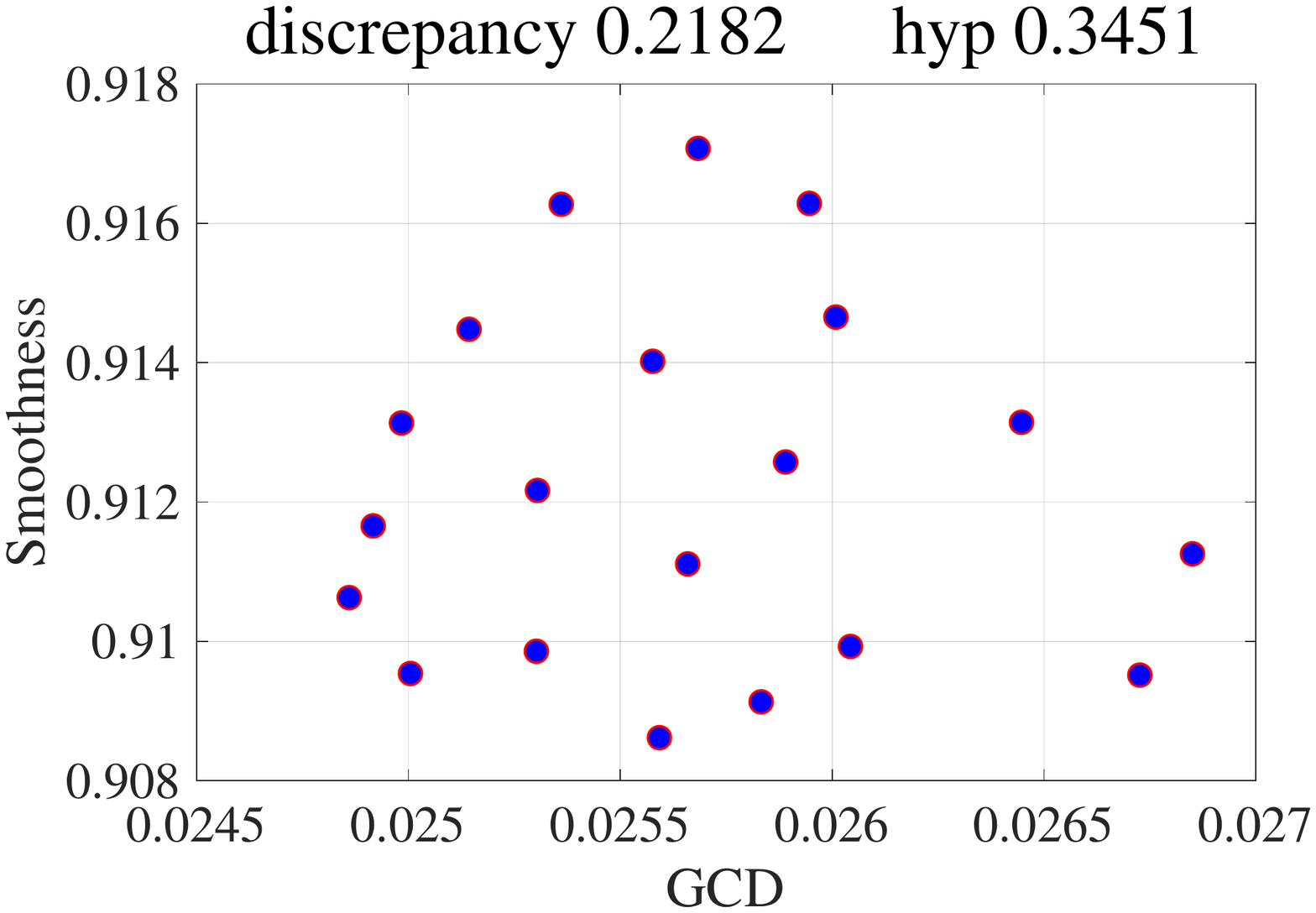}%

\rotatebox{90}{\hspace{12mm}EA$_{\text{HYP}}$} \rotatebox{90}{\hspace{4mm}\rule{25mm}{1pt}}%
\hspace{0.1cm}
\includegraphics[trim={0.5cm 5.7cm 0.47cm 6.8cm},clip,width=0.3\textwidth]
{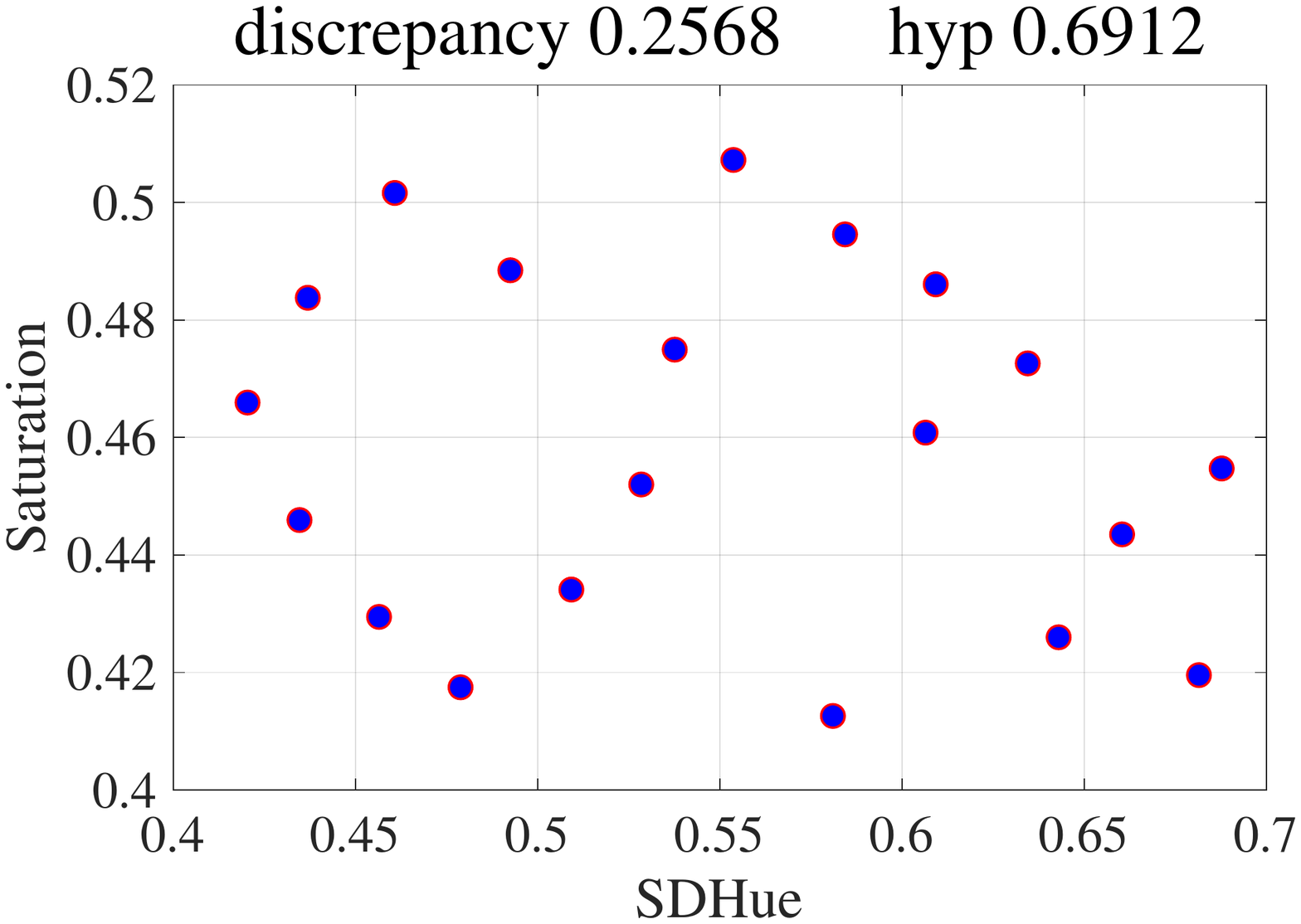}%
\includegraphics[trim={0.5cm 5.7cm 0.57cm 6.8cm},clip,width=0.3\textwidth]
{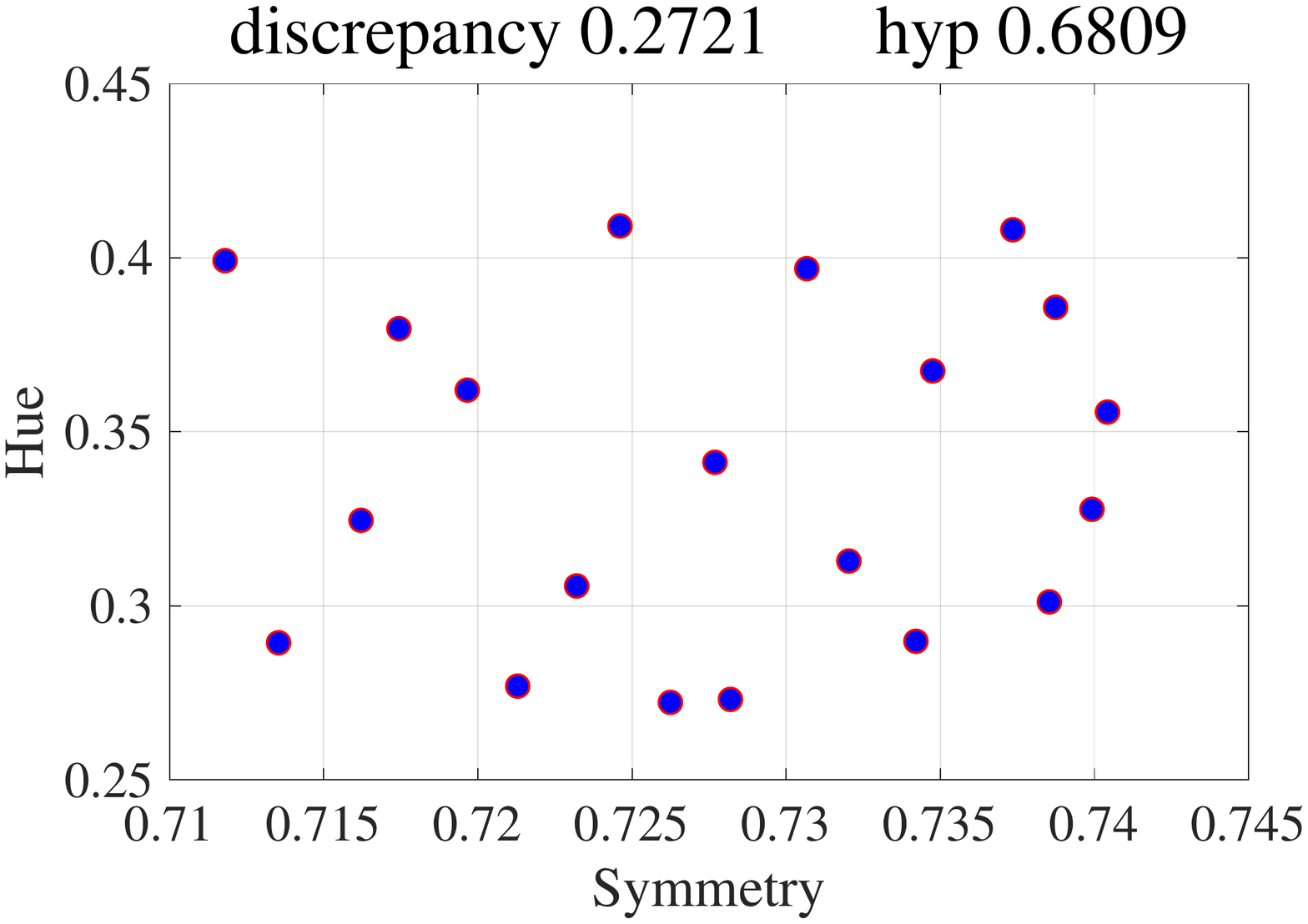}%
\includegraphics[trim={0.45cm 5.7cm 0.27cm 6.8cm},clip,width=0.31\textwidth]
{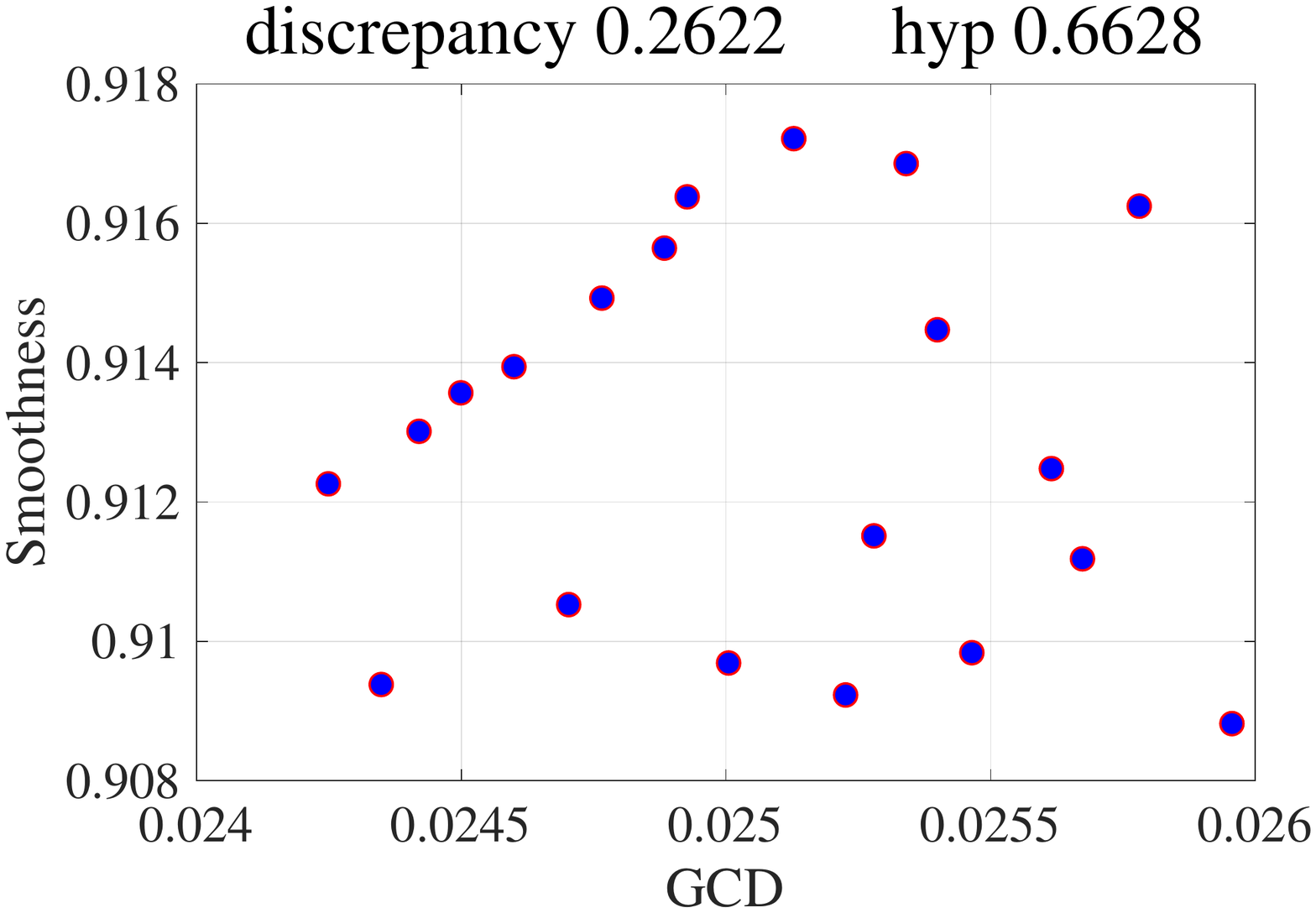}%

\rotatebox{90}{\hspace{12mm}EA$_{\text{IGD}}$} \rotatebox{90}{\hspace{4mm}\rule{26mm}{1pt}}%
\hspace{0.1cm}
\includegraphics[trim={0.05cm 4.7cm 0.002cm 4.8cm},clip,width=0.3\textwidth]{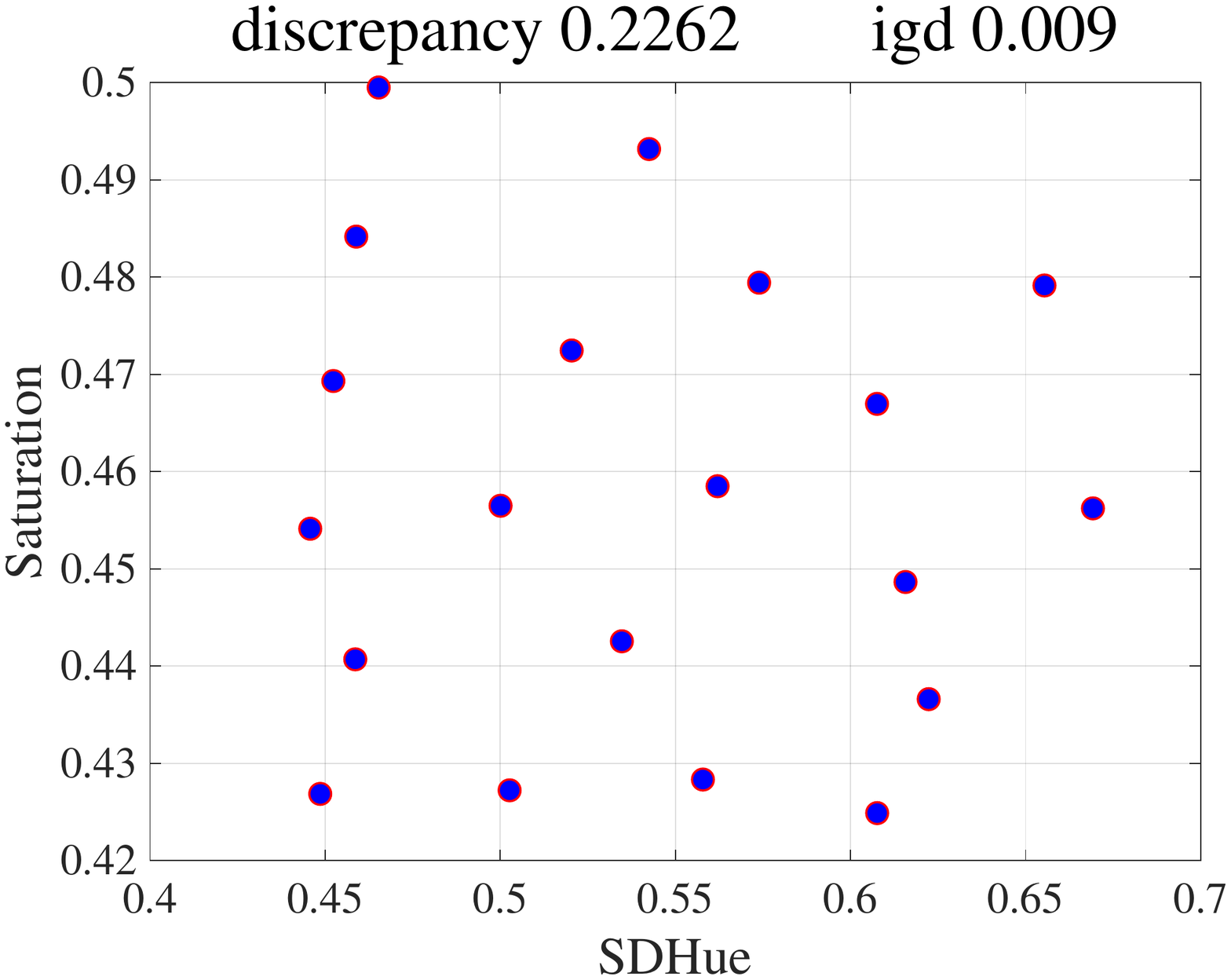}%
\includegraphics[trim={0.02cm 4.7cm 0.05cm 4.8cm},clip,width=0.3\textwidth]{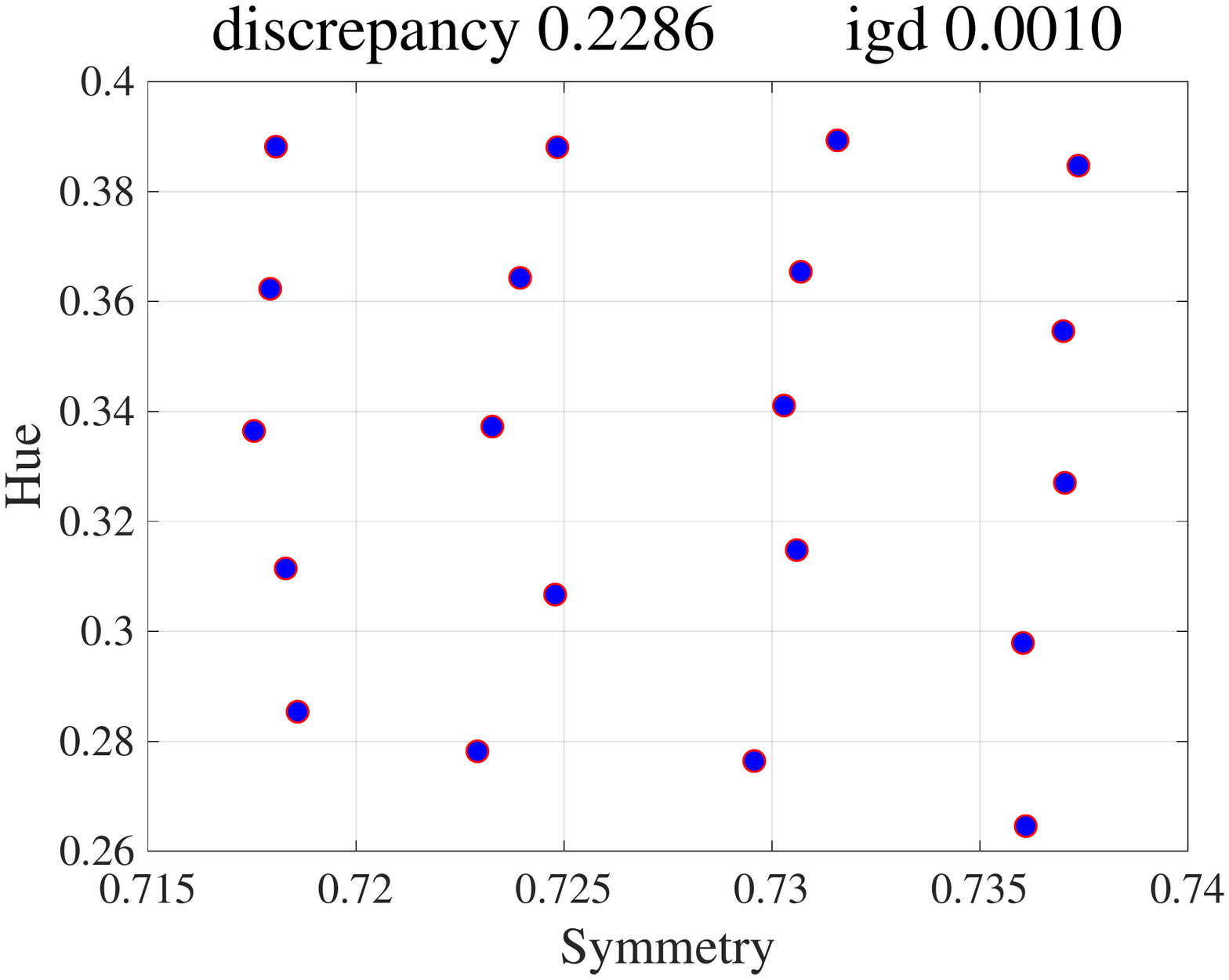}%
\includegraphics[trim={0cm 4.7cm 0.27cm 4.8cm},clip,width=0.3\textwidth]{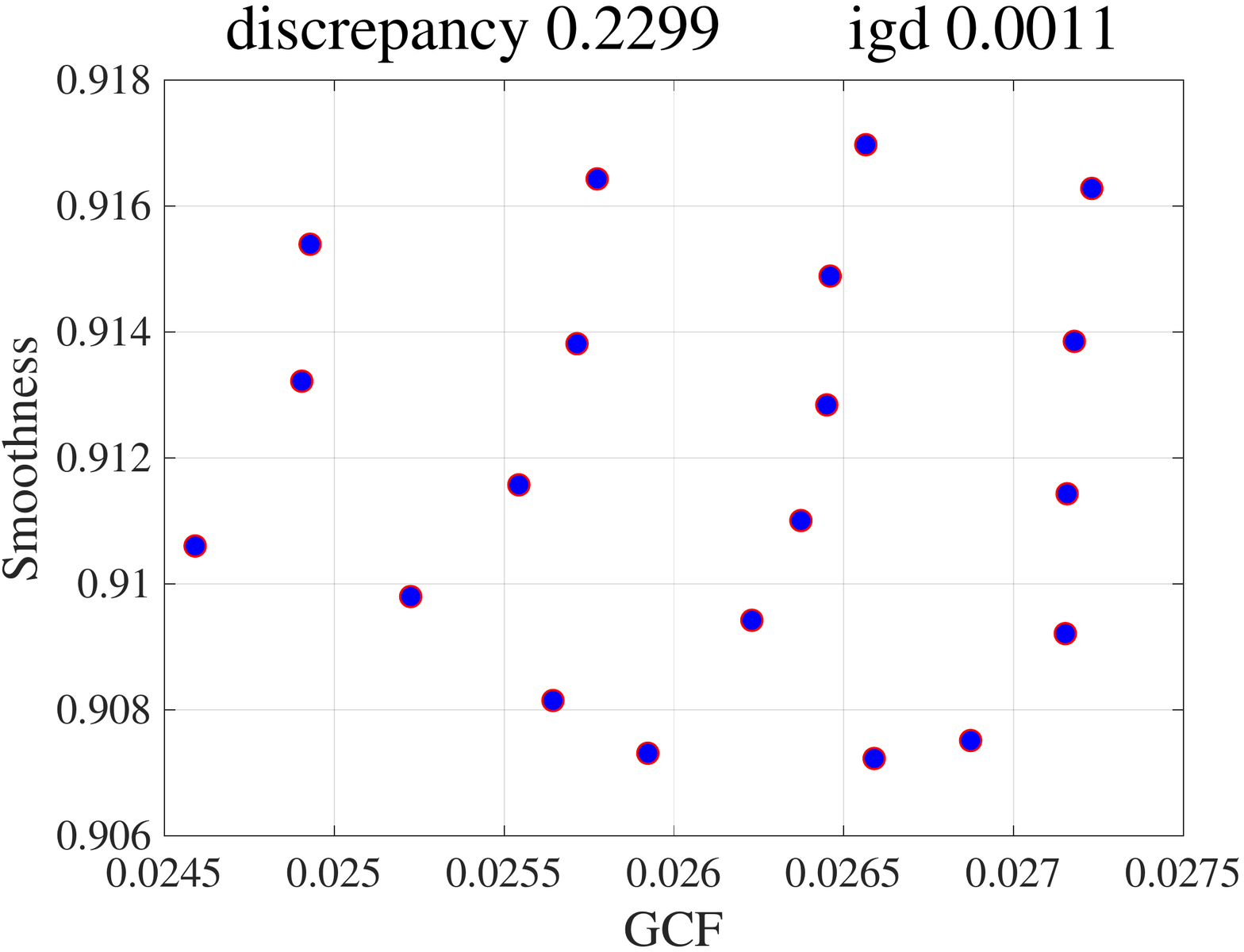}%

\rotatebox{90}{\hspace{12mm}EA$_{\text{EPS}}$} \rotatebox{90}{\hspace{4mm}\rule{26mm}{1pt}}%
\hspace{0.1cm}
\includegraphics[trim={0.002cm 4.7cm 0.02cm 4.8cm},clip,width=0.3\textwidth]{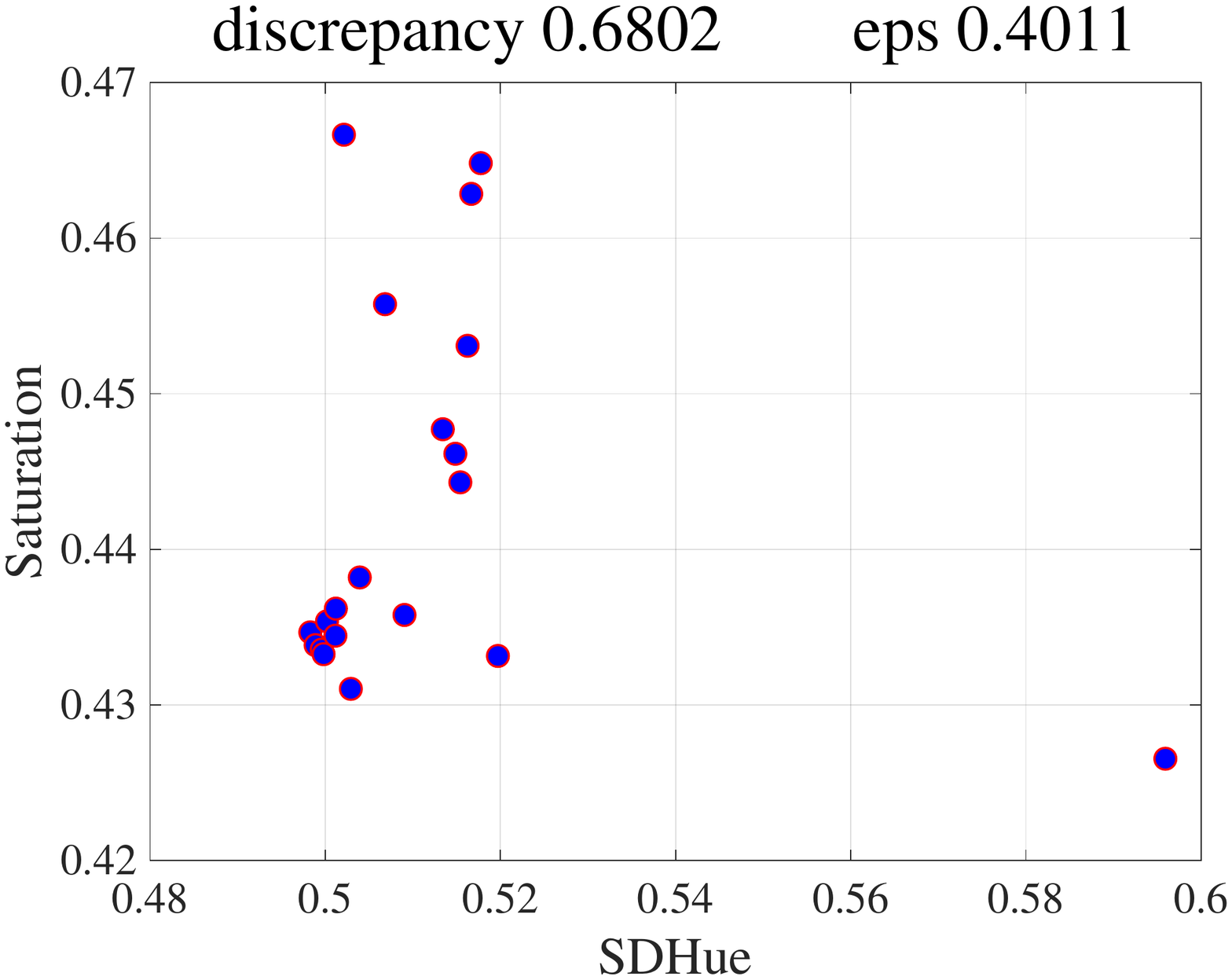}%
\includegraphics[trim={0.001cm 4.7cm 0.0002cm 4.8cm},clip,width=0.3\textwidth]{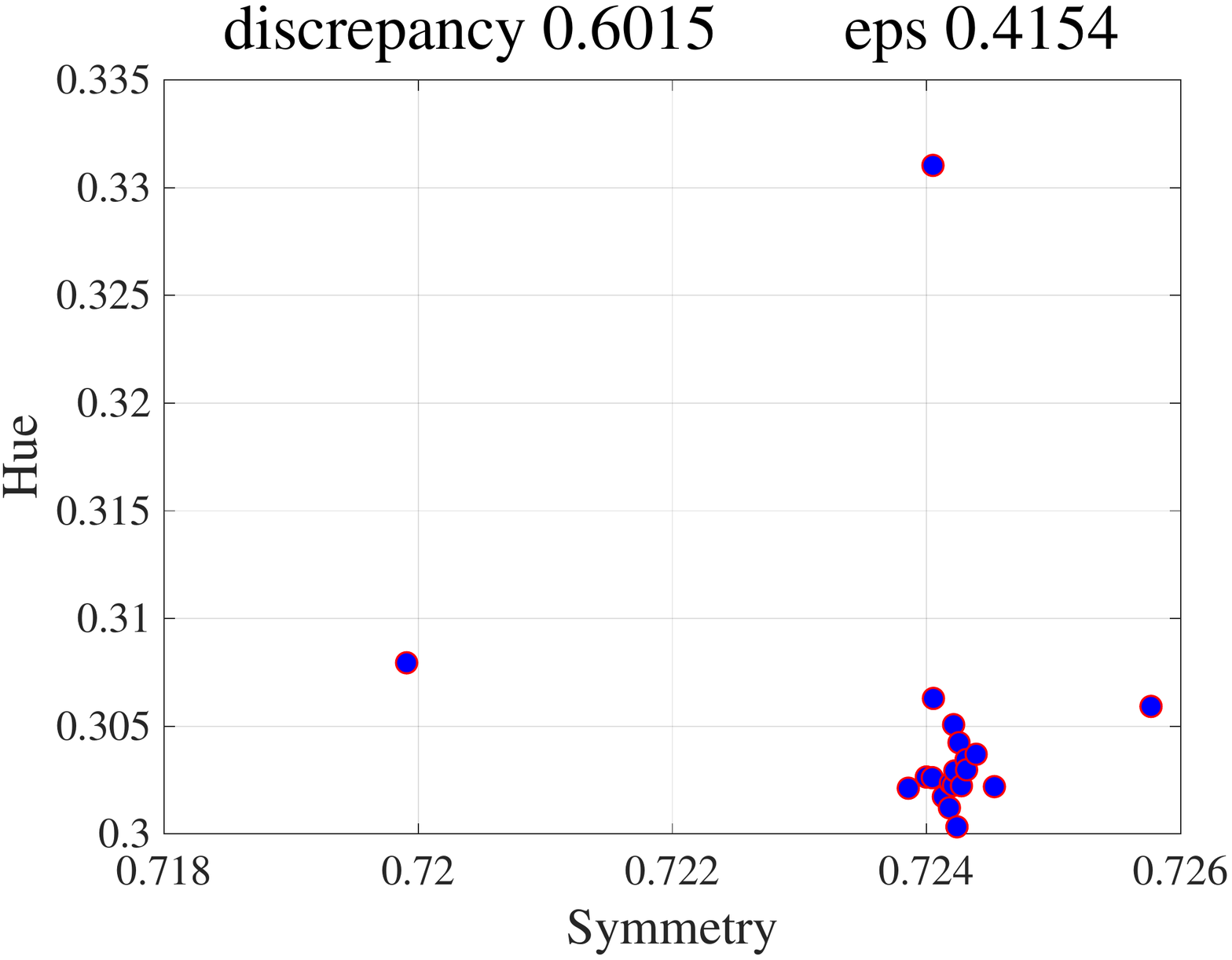}
\includegraphics[trim={0.01cm 4.7cm 0cm 4.8cm},clip,width=0.3\textwidth]{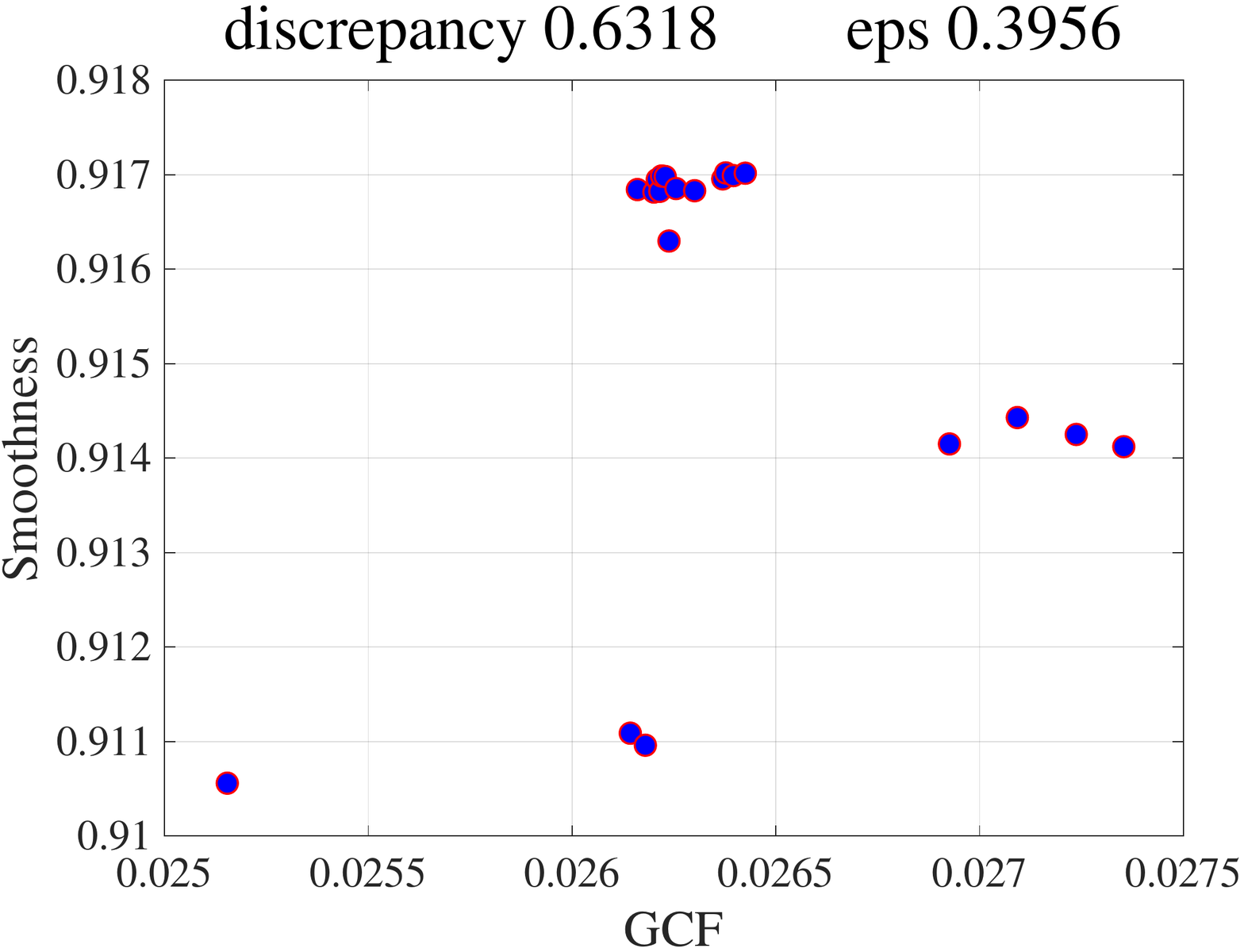}%
\caption{Feature vectors for final population of EA$_{\text{\HYPR}}$ (top), EA$_{\text{HYP}}$, EA$_{\text{IGD}}$ and EA$_{\text{EPS}}$ (bottom) for images based on pair of features from left to right: ($f_1$, $f_2$), ($f_3$, $f_4$), ($f_5$, $f_6$).}
\label{fig:images_plot} 
\end{figure}

\subsection{Experimental settings}

Now, we consider the indicator-based diversity optimization for combinations of two and three features.
We select features in order to combine different aesthetic and general features based on our initial experimental investigations and previous investigations in~\cite{DBLP:conf/evoW/NeumannAN17}. 
In this work we explore several features and features ranges described in Table~\ref{tab:features}.
We use scaled feature values while we calculate the different indicators values of a given set of points.
After having consider the combination of two features, we investigate sets of three features.
Here, we select different features combining aesthetic and general features together used in the previous experiment. 

In order to obtain a clear comparison between our present experiments and experiments based on the discrepancy-based evolutionary algorithm introduced in~\cite{DBLP:journals/corr/abs-1802-05448} we work with the same range of feature values.

We run each configuration for $2,000$ generations with a population size of $\mu = 20$ and $\lambda=1$.
To assess our results using statistical tests, we run each combination of feature-pair and indicator 30 times. 
All algorithms were implemented in $Matlab$ $(R2017b)$ and run on $48$-core compute nodes with AMD $2.80$\,GHz CPUs and $128$\,GB of RAM.
\begin{sidewaystable}
\centering
\caption{
Investigations for images with $2$ features. 
Comparison in terms of mean, standard deviation and statistical test for considered indicators.}
\label{tb:statistic-images}
\renewcommand*{\arraystretch}{1.2}\setlength{\tabcolsep}{1mm}\resizebox{\textwidth}{!}{
\begin{tabular}{lclllllllllllllll}
                     &                            & \multicolumn{3}{c}{EA$_{\text{\HYPR}}$ (1)}   
& \multicolumn{3}{c}{EA$_{\text{HYP}}$ (2)}
& \multicolumn{3}{c}{EA$_{\text{IGD}}$ (3)}                                                                     & \multicolumn{3}{c}{EA$_{\text{EPS}}$ (4)}                                                                     & \multicolumn{3}{c}{EA$_{\text{DIS}}$ (5)}                                                \\
                     &                            & \multicolumn{1}{c}{mean} & \multicolumn{1}{c}{st} & \multicolumn{1}{c}{stat}                           & \multicolumn{1}{c}{mean} & \multicolumn{1}{c}{st} & \multicolumn{1}{c}{stat}                           & 
\multicolumn{1}{c}{mean} & \multicolumn{1}{c}{st} & \multicolumn{1}{c}{stat}                           & \multicolumn{1}{c}{mean} & \multicolumn{1}{c}{st} & \multicolumn{1}{c}{stat}                           & \multicolumn{1}{c}{mean} & \multicolumn{1}{c}{st} & \multicolumn{1}{c}{stat}      \\ \cline{3-17} 

\multicolumn{1}{l}{\multirow{3}{*}{\rotatebox{90}{\HYPR}}} & \multicolumn{1}{l|}{$f_1$,$f_2$} & 0.347 & 0.004 & \multicolumn{1}{l|}{$4^{(+)}$,$5^{(+)}$} & 0.382 & 0.007 & \multicolumn{1}{l|}{$3^{(+)}$,$4^{(+)}$,$5^{(+)}$} & 0.335 & 0.003 & \multicolumn{1}{l|}{$2^{(-)}$,$5^{(+)}$} & 0.198& 0.019 & \multicolumn{1}{l|}{$1^{(-)}$,$2^{(-)}$} & 0.112 & 0.030 & $1^{(-)}$,$2^{(-)}$,$3^{(-)}$ \\

& \multicolumn{1}{l|}{$f_3$,$f_4$} & 0.344 & 0.004 & \multicolumn{1}{l|}{$2^{(+)}$,$4^{(+)}$,$5^{(+)}$} & 0.268 & 0.014 & \multicolumn{1}{l|}{$1^{(-)}$,$3^{(-)}$,$4^{(+)}$,$5^{(+)}$} & 0.339 & 0.004 & \multicolumn{1}{l|}{$2^{(+)}$,$4^{(+)}$,$5^{(+)}$} & 0.221 & 0.015 & \multicolumn{1}{l|}{$1^{(-)}$,$2^{(-)}$,$3^{(-)}$} & 0.105 & 0.025 & $1^{(-)}$,$2^{(-)}$,$3^{(-)}$ \\

& \multicolumn{1}{l|}{$f_5$,$f_6$} & 0.350 & 0.007 & \multicolumn{1}{l|}{$2^{(+)}$,$3^{(+)}$,$4^{(+)}$,$5^{(+)}$} & 0.342 & 0.004 & \multicolumn{1}{l|}{$1^{(-)}$,$4^{(+)}$,$5^{(+)}$} & 0.332 & 0.004 & \multicolumn{1}{l|}{$1^{(-)}$,$4^{(+)}$,$5^{(+)}$} & 0.220 & 0.045 & \multicolumn{1}{l|}{$1^{(-)}$,$2^{(-)}$,$3^{(-)}$} & 0.134 & 0.016 & $1^{(-)}$,$2^{(-)}$,$3^{(-)}$  \\ \cline{2-17}

\multirow{3}{*}{\rotatebox{90}{HYP}} & \multicolumn{1}{c|}{$f_1$,$f_2$}
& 0.525                   & 0.012                 & \multicolumn{1}{l|}{$3^{(+)}$,$4^{(+)}$,$5^{(+)}$}

& 0.693                   & 0.013                 & \multicolumn{1}{l|}{$3^{(+)}$,$4^{(+)}$,$5^{(+)}$} & 0.374                   & 0.006                 & \multicolumn{1}{l|}{$1^{(-)}$,$2^{(-)}$,$4^{(+)}$} & 0.344                   & 0.003                 & \multicolumn{1}{l|}{$1^{(-)}$,$2^{(-)}$,$3^{(-)}$} & 0.363                   & 0.014                 & $1^{(-)}$,$2^{(-)}$  \\
                     & \multicolumn{1}{c|}{$f_3$,$f_4$}
& 0.500                  & 0.007                 & \multicolumn{1}{l|}{$3^{(+)}$,$4^{(+)}$,$5^{(+)}$}                     
                     & 0.681                   & 0.010                 & \multicolumn{1}{l|}{$3^{(+)}$,$4^{(+)}$,$5^{(+)}$} & 0.268                   & 0.072                & \multicolumn{1}{l|}{$1^{(-)}$,$2^{(-)}$,$4^{(+)}$,$5^{(+)}$} & 0.280                   & 0.010                 & \multicolumn{1}{l|}{$1^{(-)}$,$2^{(-)}$,$3^{(-)}$} & 0.267                   & 0.014                 & $1^{(-)}$,$2^{(-)}$,$3^{(-)}$  \\
                     & \multicolumn{1}{c|}{$f_5$,$f_6$}
                     & 0.518                  & 0.012                 & \multicolumn{1}{l|}{$2^{(-)}$,$4^{(+)}$,$5^{(+)}$}
                     & 0.663                   & 0.010                 & \multicolumn{1}{l|}{$1^{(+)}$,$3^{(+)}$,$4^{(+)}$,$5^{(+)}$} & 0.335                   & 0.004                 & \multicolumn{1}{l|}{$2^{(-)}$,$4^{(+)}$} & 0.317                  & 0.006                 & \multicolumn{1}{l|}{$1^{(-)}$,$2^{(-)}$,$3^{(-)}$} & 0.327                   & 0.008                 & $1^{(-)}$,$2^{(-)}$  \\ \cline{2-17} 
\multirow{3}{*}{\rotatebox{90}{IGD}} & \multicolumn{1}{c|}{$f_1$,$f_2$}& 0.001                   & 0.335                 & \multicolumn{1}{l|}{$2^{(+)}$,$4^{(+)}$,$5^{(+)}$} & 0.003                   & 0.000                 & \multicolumn{1}{l|}{$1^{(-)}$,$3^{(-)}$}           & 0.001                   & 0.000                 & \multicolumn{1}{l|}{$2^{(+)}$,$4^{(+)}$,$5^{(+)}$}           & 0.003                   & 0.000                 & \multicolumn{1}{l|}{$1^{(-)}$,$3^{(-)}$,$5^{(+)}$} & 0.005                   & 0.001                 & $1^{(-)}$,$3^{(-)}$,$4^{(-)}$  \\
                     & \multicolumn{1}{c|}{$f_3$,$f_4$} & 0.001                   & 0.339                 & \multicolumn{1}{l|}{$2^{(+)}$,$4^{(+)}$,$5^{(+)}$}& 0.004                   & 0.000                 & \multicolumn{1}{l|}{$1^{(-)}$,$3^{(-)}$,$5^{(+)}$}           & 0.001                   & 0.000                 & \multicolumn{1}{l|}{$2^{(+)}$,$4^{(+)}$,$5^{(+)}$}           & 0.003                   & 0.000                 & \multicolumn{1}{l|}{$1^{(-)}$,$3^{(-)}$,$5^{(+)}$} & 0.005                   & 0.001                 & $1^{(-)}$,$2^{(-)}$,$3^{(-)}$,$4^{(-)}$  \\
                     & \multicolumn{1}{c|}{$f_5$,$f_6$} & 0.002                   & 0.332                & \multicolumn{1}{l|}{$2^{(+)}$,$5^{(+)}$}& 0.007                   & 0.000                 & \multicolumn{1}{l|}{$1^{(-)}$,$3^{(-)}$,$4^{(-)}$,$5^{(-)}$}           & 0.001                   & 0.000                 & \multicolumn{1}{l|}{$2^{(+)}$,$4^{(+)}$,$5^{(+)}$}           & 0.003                   & 0.001                 & \multicolumn{1}{l|}{$2^{(+)}$,$3^{(-)}$} & 0.004                   & 0.001                 & $1^{(-)}$,$2^{(+)}$,$3^{(-)}$  \\ \cline{2-17} 
\multirow{3}{*}{\rotatebox{90}{EPS}} & \multicolumn{1}{c|}{$f_1$,$f_2$}& 0.190                   & 0.198                 & \multicolumn{1}{l|}{$2^{(+)}$,$4^{(+)}$,$5^{(+)}$} & 0.498                   & 0.011                 & \multicolumn{1}{l|}{$1^{(-)}$, $3^{(-)}$}           & 0.194                   & 0.032                 & \multicolumn{1}{l|}{$2^{(+)}$,$4^{(+)}$,$5^{(+)}$}           & 0.402                   & 0.039                 & \multicolumn{1}{l|}{$1^{(-)}$,$3^{(-)}$,$5^{(+)}$} & 0.600                   & 0.106                 & $1^{(-)}$,$3^{(-)}$,$4^{(-)}$  \\
                     & \multicolumn{1}{c|}{$f_3$,$f_4$}& 0.198                   & 0.221                 & \multicolumn{1}{l|}{$2^{(+)}$,$4^{(+)}$,$5^{(+)}$} & 0.569                   & 0.016                 & \multicolumn{1}{l|}{$1^{(-)}$,$3^{(-)}$}           & 0.208                   & 0.035                 & \multicolumn{1}{l|}{$2^{(+)}$,$4^{(+)}$,$5^{(+)}$}           & 0.418                   & 0.036                 & \multicolumn{1}{l|}{$1^{(-)}$,$3^{(-)}$,$5^{(+)}$} & 0.615                   & 0.069                 & $1^{(-)}$,$3^{(-)}$,$4^{(-)}$  \\
                     & \multicolumn{1}{c|}{$f_5$,$f_6$}& 0.125                   & 0.220                 & \multicolumn{1}{l|}{$2^{(+)}$,$4^{(+)}$,$5^{(+)}$} & 0.946                   & 0.001                & \multicolumn{1}{l|}{$1^{(-)}$,$3^{(-)}$,$4^{(-)}$}           & 0.225                   & 0.064                 & \multicolumn{1}{l|}{$2^{(+)}$,$4^{(+)}$,$5^{(+)}$}           & 0.397                   & 0.110                 & \multicolumn{1}{l|}{$1^{(-)}$,$2^{(+)}$,$3^{(-)}$} & 0.587                   & 0.063                 & $1^{(-)}$,$3^{(-)}$  \\ \cline{2-17} 
\multirow{3}{*}{\rotatebox{90}{DIS}} & \multicolumn{1}{c|}{$f_1$,$f_2$}& 0.171                   & 0.018                 & \multicolumn{1}{l|}{$2^{(+)}$,$4^{(+)}$,$5^{(+)}$} & 0.257                   & 0.010                 & \multicolumn{1}{l|}{$1^{(-)}$,$4^{(+)}$}           & 0.201                   & 0.031                 & \multicolumn{1}{l|}{$4^{(+)}$,$5^{(+)}$}           & 0.686                   & 0.064                 & \multicolumn{1}{l|}{$1^{(-)}$,$2^{(-)}$,$3^{(-)}$,$5^{(-)}$} & 0.204                   & 0.116                 & $1^{(-)}$,$3^{(-)}$,$4^{(+)}$  \\
                     & \multicolumn{1}{c|}{$f_3$,$f_4$}& 0.234                   & 0.031                 & \multicolumn{1}{l|}{$4^{(+)}$} & 0.273                   & 0.041                 & \multicolumn{1}{l|}{$3^{(-)}$,$4^{(+)}$,$5^{(-)}$} & 0.198                   & 0.017                 & \multicolumn{1}{l|}{$2^{(+)}$,$4^{(+)}$} & 0.606                   & 0.054                 & \multicolumn{1}{l|}{$1^{(-)}$,$2^{(-)}$,$3^{(-)}$,$5^{(-)}$} & 0.228                   & 0.059                 & $2^{(+)}$,$4^{(+)}$  \\
                     & \multicolumn{1}{c|}{$f_5$,$f_6$}& 0.221                   & 0.026                 & \multicolumn{1}{l|}{$4^{(+)}$} & 0.263                   & 0.070                 & \multicolumn{1}{l|}{$3^{(-)}$,$4^{(+)}$,$5^{(-)}$} & 0.205                   & 0.055                 & \multicolumn{1}{l|}{$2^{(+)}$,$4^{(+)}$} & 0.633                   & 0.158                 & \multicolumn{1}{l|}{$1^{(-)}$,$2^{(-)}$,$3^{(-)}$,$5^{(-)}$} & 0.203                   & 0.054                 & $2^{(+)}$,$4^{(+)}$ 
\end{tabular} }
\end{sidewaystable}

\subsection{Experimental results and analysis}

We present a series of experiments for two- and three-feature combinations in order to evaluate our evolutionary diversity algorithms based on the use of indicators from multi-objective optimization described in Section~\ref{sec:ind}.

\subsubsection{Two-feature combinations}

Our results are summarized in Table~\ref{tb:statistic-images} and Table~\ref{tb:statistic-images-new}. The columns represent the algorithms with the corresponding mean value and standard deviation. The rows represent the indicators \HYPR, HYP, IGD, EPS and discrepancy (DIS). For each indicator, we obtained results for all sets of features. 

Additionally, we use the Kruskal-Wallis test for statistical validation with $95$\% confidence and subsequently apply the Bonferroni post-hoc statistical procedure. For a detailed description of the statistical tests we refer the reader to~\cite{Corder09}.
Our experimental analysis characterizes the behavior of the four examined indicator-based evolutionary algorithms and discrepancy-based evolutionary algorithm. In the statistical tests shown in Table~\ref{tb:statistic-images} and Table~\ref{tb:statistic-images-new}, $A^{(+)}$ is equivalent to the statement that the algorithm in this column outperformed algorithm $A$, and $A^{(-)}$ is equivalent to the statement that $A$ outperformed the algorithm given in the column. If the algorithm $A$ does not appear, this means that no significant difference was determined.

\begin{figure}[!t]
\centering
\includegraphics[width=45mm]
{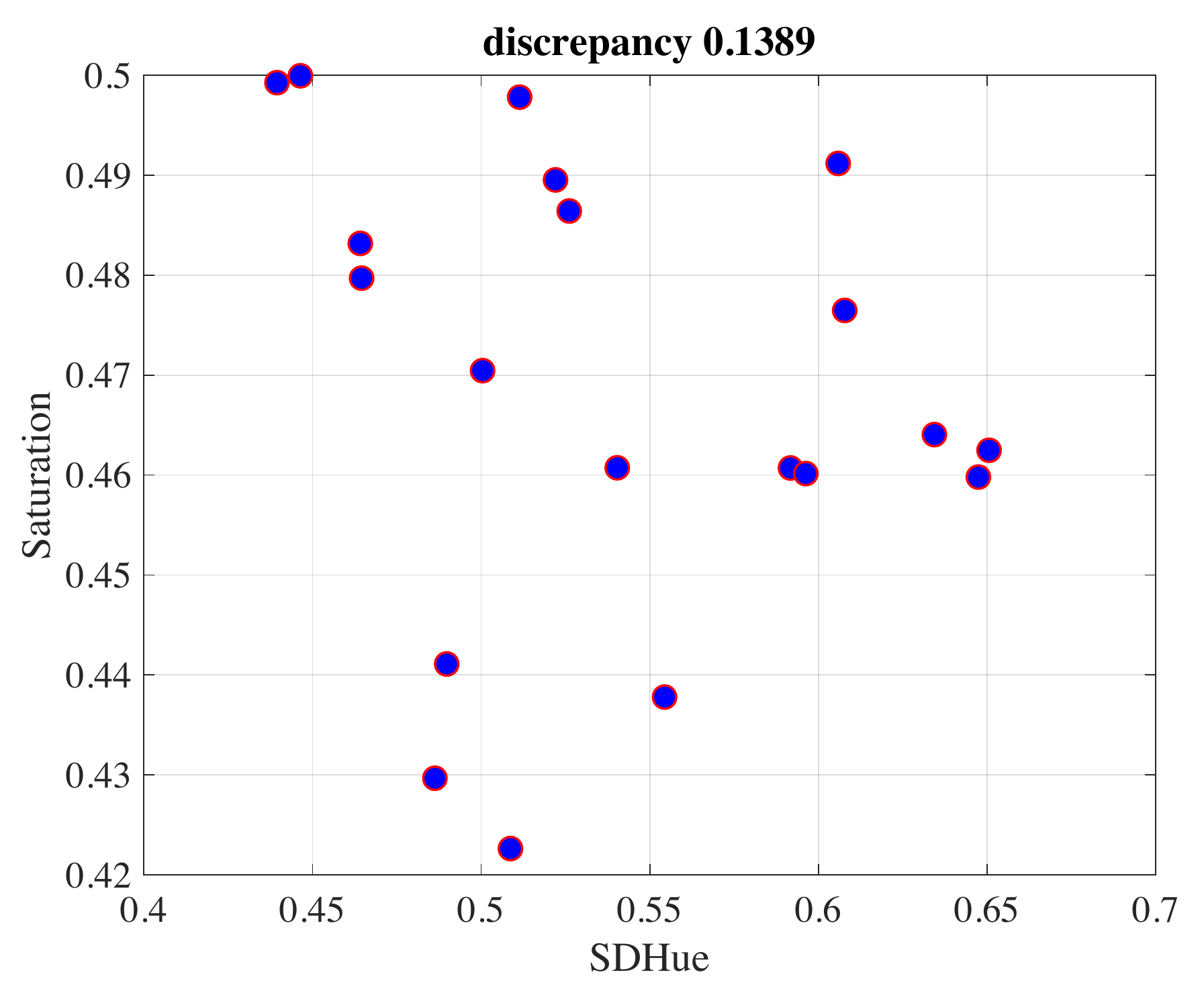}
\caption{Feature vectors for final population of EA$_{\text{DIS}}$~\cite{DBLP:journals/corr/abs-1802-05448} for images based on ($f_1$, $f_2$).
} 
\label{plot:ead:example}\vspace{-3mm}
\end{figure}
Figure~\ref{fig:images_plot} illustrates feature plots of (randomly selected) final populations of EA$_{\text{\HYPR}}$ (top), EA$_{\text{HYP}}$, EA$_{\text{IGD}}$ and EA$_{\text{EPS}}$ (bottom) for three pairs of feature combinations. In the first column, we see the feature vectors for the final population of the four algorithms for image based on pairs of features ($f_1$,$f_2$). 
It can be observed that the discrepancy value for EA$_{\text{\HYPR}}$ is $0.1767$. This is significantly smaller than the one for EA$_{\text{EPS}}$ at $0.6802$. Note that smaller discrepancy values are considered to be better.
The middle column shows the combination of the feature pair ($f_3$,$f_4$). The discrepancy value for feature pair ($f_3$,$f_4$) for EA$_{\text{IGD}}$ is $0.2286$ whereas it is $0.6015$ for EA$_{\text{EPS}}$.
The last column shows the final populations of the diversity optimization when considering feature pair ($f_5$,$f_6$). The discrepancy value for feature pair ($f_5$,$f_6$) is the smallest among all algorithms for EA$_{\text{\HYPR}}$ at $0.2182$ and the highest for EA$_{\text{EPS}}$ at $0.6318$.

In summary, we observe that EA$_{\text{\HYPR}}$, EA$_{\text{HYP}}$ and EA$_{\text{IGD}}$ achieve a good and even coverage of the feature space, especially in comparison with the discrepancy-based diversification (see Figure~\ref{plot:ead:example} for an example from~\cite{DBLP:journals/corr/abs-1802-05448}). Interestingly, EA$_{\text{EPS}}$ appears to experience difficulties, and it achieves the worst coverage in the search space in all scenarios.

Moreover, in Table~\ref{tb:statistic-images}, we observe that the EA$_{\text{HYP}}$ algorithm has the best performance among all algorithms.
It has the highest hypervolume values for all features combinations, and this is also statistically significant. 
Also, due to the statistical tests we can say that EA$_{\text{\HYPR}}$ outperforms EA$_{\text{EPS}}$ and EA$_{\text{DIS}}$ with respect to the inverted generational distance and additive epsilon approximation indicator measurements values for all sets of features. We observe that EA$_{\text{\HYPR}}$ considering IGD and EPS values has no significant differences to EA$_{\text{IGD}}$. In terms of discrepancy, the EA$_{\text{\HYPR}}$ has a following characteristic: 
for set of features ($f_1$,$f_2$) the EA$_{\text{\HYPR}}$ outperforms EA$_{\text{HYP}}$, EA$_{\text{EPS}}$ and EA$_{\text{DIS}}$, however, it only outperforms the EA$_{\text{EPS}}$ for the set of features ($f_3$,$f_4$) and ($f_5$,$f_6$).

Furthermore, EA$_{\text{IGD}}$ outperforms the EA$_{\text{HYP}}$, EA$_{\text{EPS}}$ and the EA$_{\text{DIS}}$ with respect to IGD, EPS and DIS indicators measurements in most of the cases and achieves the lowest values for IGD measurements among all others algorithms for all sets of features. The best performance achieves EA$_{\text{IGD}}$ for discrepancy measurements for the combinations of features ($f_3$,$f_4$) and ($f_5$,$f_6$) with values $0.198$ and $0.205$. The hypervolume-based approaches EA$_{\text{\HYPR}}$ and EA$_{\text{HYP}}$ outperform EA$_{\text{IGD}}$ for all sets of features.

Among all others algorithms EA$_{\text{EPS}}$ shows the worst performance. Especially, according to all indicators measurements and all sets of features, the EA$_{\text{EPS}}$ is dominated by EA$_{\text{HYP}}$ and EA$_{\text{IGD}}$, and this difference is statistically significant.

Finally, the EA$_{\text{DIS}}$ is dominated by EA$_{\text{\HYPR}}$ and EA$_{\text{HYP}}$, EA$_{\text{IGD}}$ and EA$_{\text{EPS}}$ with respect to the \HYPR, HYP, IGD and EPS indicator values. Also, most results are significantly different with respect to the HYP, IGD, EPS indicators.
EA$_{\text{DIS}}$ achieves the best performance with respect to the DIS indicator for the combinations of features ($f_3$,$f_4$) and ($f_5$,$f_6$). The EA$_{\text{DIS}}$ outperforms the EA$_{\text{HYP}}$ and EA$_{\text{EPS}}$ in this case. For the combinations ($f_1$,$f_2$) with respect to the DIS indicator, the EA$_{\text{DIS}}$ is dominated by EA$_{\text{\HYPR}}$ and EA$_{\text{IGD}}$. 
\\

\subsubsection{Three-feature combinations}

\begin{figure*}[t]
\label{fig:Images3D3}
\rotatebox{90}{\hspace{10mm}EA$_{\text{HYP}}$} \rotatebox{90}{\rule{28mm}{1pt}}%
\hspace{0.1cm}
\includegraphics[width=0.31\textwidth]
{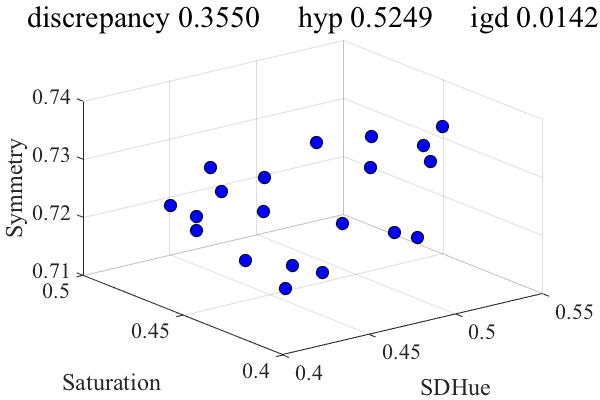}
\includegraphics[width=0.31\textwidth]
{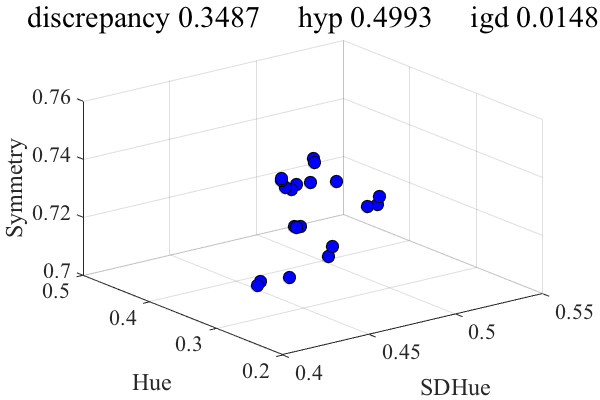}
\includegraphics[width=0.31\textwidth]
{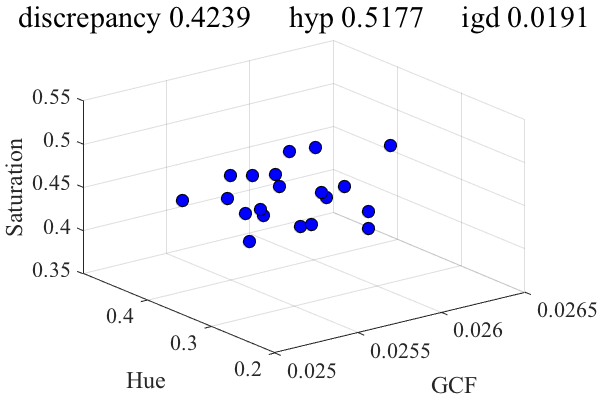}
\vspace{0.3cm}
\\
\rotatebox{90}{\hspace{10mm}EA$_{IGD}$} \rotatebox{90}{\rule{28mm}{1pt}}%
\hspace{0.1cm}
\includegraphics[width=0.31\textwidth]
{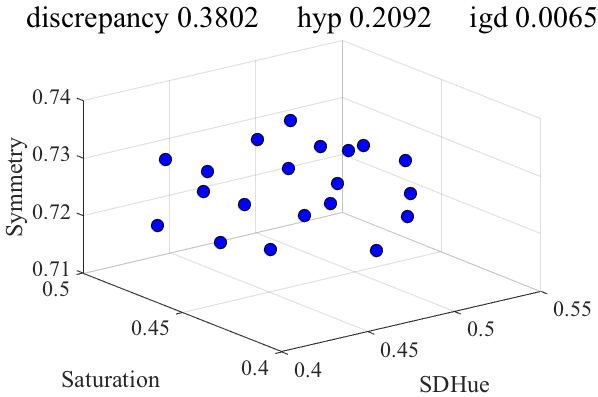}
\includegraphics[width=0.31\textwidth]
{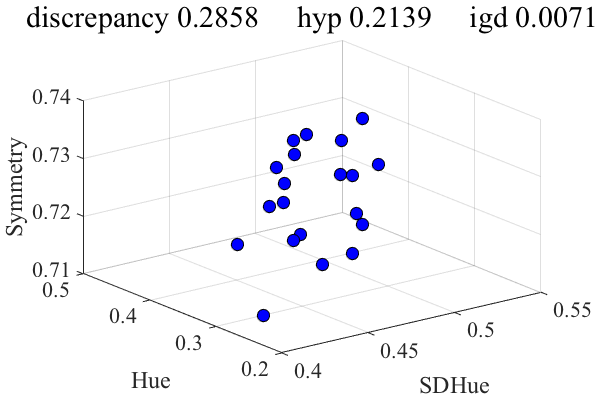}
\includegraphics[width=0.31\textwidth]
{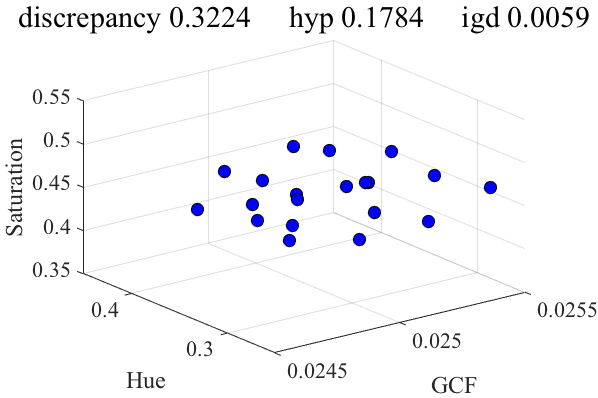}
\vspace{0.3cm}\\
\rotatebox{90}{\hspace{10mm}EA$_{\text{DIS}}$} \rotatebox{90}{\rule{28mm}{1pt}}%
\hspace{0.1cm}
\includegraphics[width=0.31\textwidth]
{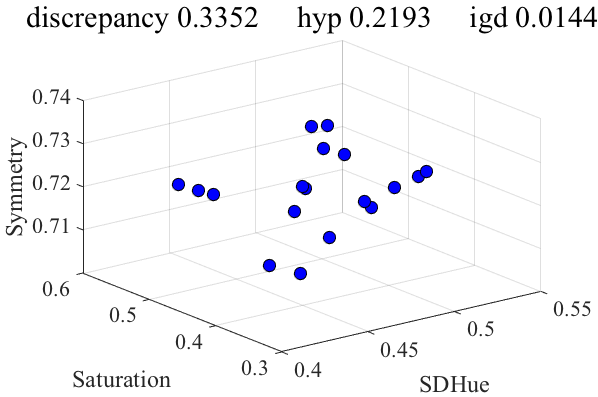}
\includegraphics[width=0.31\textwidth]
{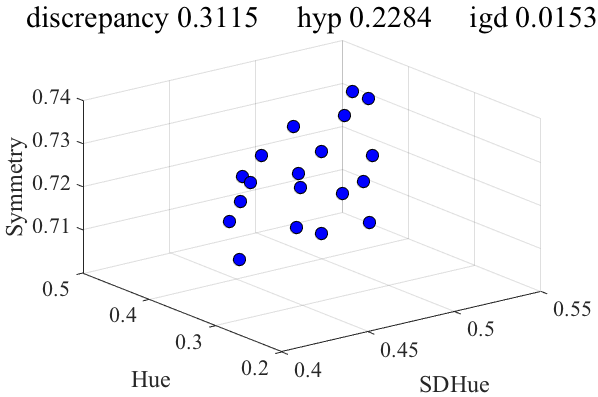}
\includegraphics[width=0.31\textwidth]
{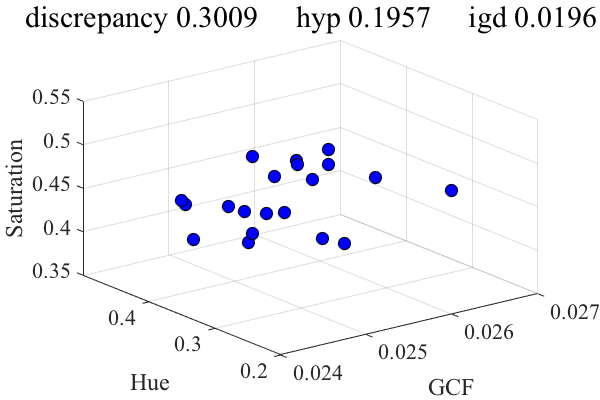}
\caption{Feature vectors for final population of EA$_{\text{HYP}}$ (top), EA$_{\text{IGD}}$ (middle) and EA$_{\text{DIS}}$ (bottom) for images based on three features from left to right: ($f_1$, $f_2$, $f_3$), ($f_1$, $f_4$, $f_3$), ($f_5$, $f_4$, $f_2$).}
\label{fig:images_plot_for3d}
\end{figure*}

\begin{table*}[t]
\centering
\caption{Investigations for images with $3$ features. 
Comparison in terms of mean, standard deviation and statistical test for considered indicators.}
\vspace{4mm}
\renewcommand*{\arraystretch}{1.3}\setlength{\tabcolsep}{1.2mm}\resizebox{1.05\textwidth}{!}{

\begin{tabular}{lcllllllllllll}
                     &                            & \multicolumn{3}{c}{EA$_{\text{HYP}}$ (1)}                                                                     & \multicolumn{3}{c}{EA$_{\text{IGD}}$ (2)}                                                                     & \multicolumn{3}{c}{EA$_{\text{DIS}}$ (3)}                                                                     & \multicolumn{3}{c}{}                                                \\
                     &                            & \multicolumn{1}{c}{mean} & \multicolumn{1}{c}{st} & \multicolumn{1}{c}{stat}                           & \multicolumn{1}{c}{mean} & \multicolumn{1}{c}{st} & \multicolumn{1}{c}{stat}                           & \multicolumn{1}{c}{mean} & \multicolumn{1}{c}{st} & \multicolumn{1}{c}{stat}                           &  
                     \\ \cline{3-11} 
\multirow{3}{*}{\rotatebox{90}{HYP}} & \multicolumn{1}{c|}{$f_1$,$f_2$,$f_3$} & 0.5251                  & 0.0122                 & \multicolumn{1}{l|}{$2^{(+)}$,$3^{(+)}$} & 0.2096                  & 0.0018                 & \multicolumn{1}{l|}{$1^{(-)}$,$3^{(-)}$} & 0.2196                  & 0.0110                 & \multicolumn{1}{l}{$1^{(-)}$,$2^{(+)}$} &                    &                 & \\
                     & \multicolumn{1}{c|}{$f_1$,$f_4$,$f_3$} & 0.4998                  & 0.0071                 & \multicolumn{1}{l|}{$2^{(+)}$,$3^{(+)}$} &                   0.2142&  0.0036                & \multicolumn{1}{l|}{$1^{(-)}$,$3^{(-)}$} & 0.2286                & 0.0034                 & \multicolumn{1}{l}{$1^{(-)}$,$2^{(+)}$} &                    &                 &  \\
                     & \multicolumn{1}{c|}{$f_5$,$f_4$,$f_2$} &  0.5181                 &   0.0122             & \multicolumn{1}{l|}{$2^{(+)}$,$3^{(+)}$} & 0.1785                   & 0.0017                 & \multicolumn{1}{l|}{$1^{(-)}$,$3^{(-)}$} & 0.1961                  & 0.0023                 & \multicolumn{1}{l}{$1^{(-)}$,$2^{(+)}$} &                   &               &  \\ \cline{2-11} 
\multirow{3}{*}{\rotatebox{90}{IGD}} & \multicolumn{1}{c|}{$f_1$,$f_2$,$f_3$} & 0.0146                    & 0.0001                & \multicolumn{1}{l|}{$2^{(-)}$,$3^{(+)}$}           &       0.0067              &      0.0003           & \multicolumn{1}{l|}{$1^{(+)}$,$3^{(+)}$}           & 0.0148                  & 0.0003               & \multicolumn{1}{l}{$1^{(-)}$,$2^{(-)}$} &  \\
                     & \multicolumn{1}{c|}{$f_1$,$f_4$,$f_3$} & 0.0150                 & 0.0001               & \multicolumn{1}{l|}{$2^{(-)}$}           & 0.0074                  & 0.0002                 & \multicolumn{1}{l|}{$1^{(+)}$,$3^{(+)}$}           & 0.0151                   & 0.0001                 & \multicolumn{1}{l}{$2^{(-)}$} &  \\
                     & \multicolumn{1}{c|}{$f_5$,$f_4$,$f_2$} & 0.0193                  & 0.0001                 & \multicolumn{1}{l|}{$2^{(-)}$,$3^{(+)}$}           & 0.0062                   & 0.0002                 & \multicolumn{1}{l|}{$1^{(+)}$,$3^{(+)}$}           &  0.0199                   & 0.0007                                    & \multicolumn{1}{l}{$1^{(-)}$,$2^{(-)}$} &  \\ \cline{2-11} 
\multirow{3}{*}{\rotatebox{90}{DIS}} & \multicolumn{1}{c|}{$f_1$,$f_2$,$f_3$} &0.3554                  & 0.0458                 & \multicolumn{1}{l|}{$2^{(+)}$,$3^{(-)}$}           &   0.3809                  &    0.0522               & \multicolumn{1}{l|}{$1^{(-)}$,$3^{(-)}$}           & 0.3350                  & 0.1002                 & \multicolumn{1}{l}{$1^{(+)}$,$2^{(+)}$} & \\
                     & \multicolumn{1}{c|}{$f_1$,$f_4$,$f_3$} & 0.3493                  & 0.0532                 & \multicolumn{1}{l|}{$2^{(-)}$}           &                    0.2860 &          0.0342         & \multicolumn{1}{l|}{$1^{(+)}$,$3^{(+)}$}           & 0.3118                  & 0.1309                & \multicolumn{1}{l}{$2^{(-)}$} &  \\
                     & \multicolumn{1}{c|}{$f_5$,$f_4$,$f_2$} & 0.4237                  & 0.0643                & \multicolumn{1}{l|}{$2^{(-)}$,$3^{(-)}$}           &                    0.3227&       0.0557           & \multicolumn{1}{l|}{$1^{(+)}$,$3^{(-)}$}           & 0.3007                  & 0.1467                & \multicolumn{1}{l}{$1^{(+)}$,$2^{(+)}$} & \\ 

\end{tabular} }
\label{tb:statistic-images-new}
\end{table*}

The triplets of features are described in Table~\ref{tab:features} and the results are summarized in Table~\ref{tb:statistic-images-new}. 
As before, the columns represent the algorithms with the corresponding mean value and standard deviation, and the rows represent the indicators.

Figure~\ref{fig:images_plot_for3d} shows feature plots of (randomly selected) final populations of EA$_{\text{HYP}}$ (top), EA$_{\text{IGD}}$ and EA$_{\text{DIS}}$ (bottom) for all sets of features.
We can observe that the HYP value for EA$_{\text{HYP}}$ is $0.5249$, which is significantly higher than the ones for EA$_{\text{IGD}}$ at $0.2092$ and EA$_{\text{DIS}}$ at $0.2193$.
The IGD value for EA$_{\text{IGD}}$ is the lowest (and best) at $0.0065$. The EA$_{\text{DIS}}$ achieves the lowest (and best) discrepancy value $0.3352$. The situation is similar for the other two triplets.
The HYP values for EA$_{\text{HYP}}$ $0.4993$ and $0.5177$ are significantly higher than the ones for EA$_{\text{IGD}}$ at $0.2139$ and $0.1784$, and accordantly for EA$_{\text{DIS}}$ at $0.2284$ and $0.1957$. In contrast, EA$_{\text{IGD}}$ obtains the smallest discrepancy values  at $0.2858$ for the second set of features.

In Table~\ref{tb:statistic-images-new}, we compare EA$_{\text{HYP}}$ and EA$_{\text{IGD}}$ with EA$_{\text{DIS}}$ algorithm with respect to two multi-objective indicators and the discrepancy measurement. 
Table~\ref{tb:statistic-images-new} shows that EA$_{\text{HYP}}$ outperforms EA$_{\text{IGD}}$ and EA$_{\text{DIS}}$ for all three sets of features with respect to the HYP indicator. In particular, for the first set of features ($f_1$,$f_2$,$f_3$) the EA$_{\text{HYP}}$ algorithm obtains the value $0.5251$, and only $0.2096$ for IGD, and $0.2196$ for discrepancy.

Comparing EA$_{\text{IGD}}$ to EA$_{\text{HYP}}$ and EA$_{\text{DIS}}$ with respect to the IGD indicator, we find a similar picture as for the EA$_{\text{HYP}}$ algorithm. EA$_{\text{IGD}}$ clearly outperforms the EA$_{\text{HYP}}$ and EA$_{\text{DIS}}$ for all three sets of features. The EA$_{\text{DIS}}$ algorithm also clearly outperforms EA$_{\text{HYP}}$ and EA$_{\text{IGD}}$ with respect to discrepancy. Overall, EA$_{\text{DIS}}$ achieves improvements in terms of discrepancy value among another two algorithms for all sets of features apart from one exception. It can be observed that for the set of feature ($f_1$,$f_4$,$f_3$) EA$_{\text{DIS}}$ does not have a major advantage over the EA$_{\text{IGD}}$.

In a nutshell, according to our statistical tests the EA$_{\text{HYP}}$ outperforms all examined algorithms with respect to the HYP indicator values for all sets of features in case of two-feature combination. Moreover, EA$_{\text{IGD}}$ outperforms EA$_{\text{HYP}}$, EA$_{\text{EPS}}$ and EA$_{\text{DIS}}$ with respect to the IGD indicator, which was expected, but it shows no significant difference to EA$_{\text{\HYPR}}$ for the first two sets of features. The EA$_{\text{EPS}}$ algorithm has the worst performance, no matter the indicator considered. Similarly, considering our experiments for three-feature combinations, EA$_{\text{HYP}}$ and EA$_{\text{IGD}}$ achieve the best results, which are also statistically significant.


\section{Traveling Salesperson Problem}
\label{sec:tsp}

We also test our newly introduced approach on the feature-based diversity maximization problem of Traveling Salesperson Problem (TSP) instances. The TSP is one of the well-known NP-hard combinatorial optimization problems with many real-world applications. The TSP we consider in this research is the classical Euclidean TSP with multiple cities in the $[0,1]^2$ Euclidean plane as input and a Hamiltonian cycle with the minimal total
distance as output. TSP instances can be characterized by different sets of features, and in this research we select a set of feature combinations studied in~\cite{Mersmann2013}.

\begin{table}[!t]
\centering
\caption{Description of features for TSP instances.} \label{tab:featuresTSP}
\vspace{1.8mm}
  \renewcommand{\arraystretch}{1.5}\setlength{\tabcolsep}{1.2mm}\resizebox{1.0\textwidth}{!}{
\begin{tabular}{lllll}
\hline
      &  \large Notation                                  & $f^{min}$ & $f^{max}$ & \large Description            \\ \hline
$f_1$ & \large angle\_mean                                    & $0.70$ & $2.90$ & \large mean value of the angles made by each point with its two nearest neighbor points \\
$f_2$ & \large centroid\_mean\_distance\_to\_centroid                                      & $0.24$ & $0.70$ & \large mean value of the distances from the points to the centroid \\
$f_3$ & \large nnds\_mean & $0.10$ & $0.70$ & \large mean distance between nearest neighbours        \\
$f_4$ & \large mst\_dists\_mean                                      & $0.06$ & $0.15$ &  \large mean distance of the minimum spanning trees \\
\hline
\end{tabular} 
\vspace{3mm}
}
\end{table}

\begin{figure}[!t]
\rotatebox{90}{\hspace{9mm}EA$_{\text{\HYPR}}$} \rotatebox{90}{\rule{31mm}{1pt}} 
\includegraphics[trim={0.5cm 7.7cm 0.7cm 6.8cm},clip,width=0.32\textwidth]{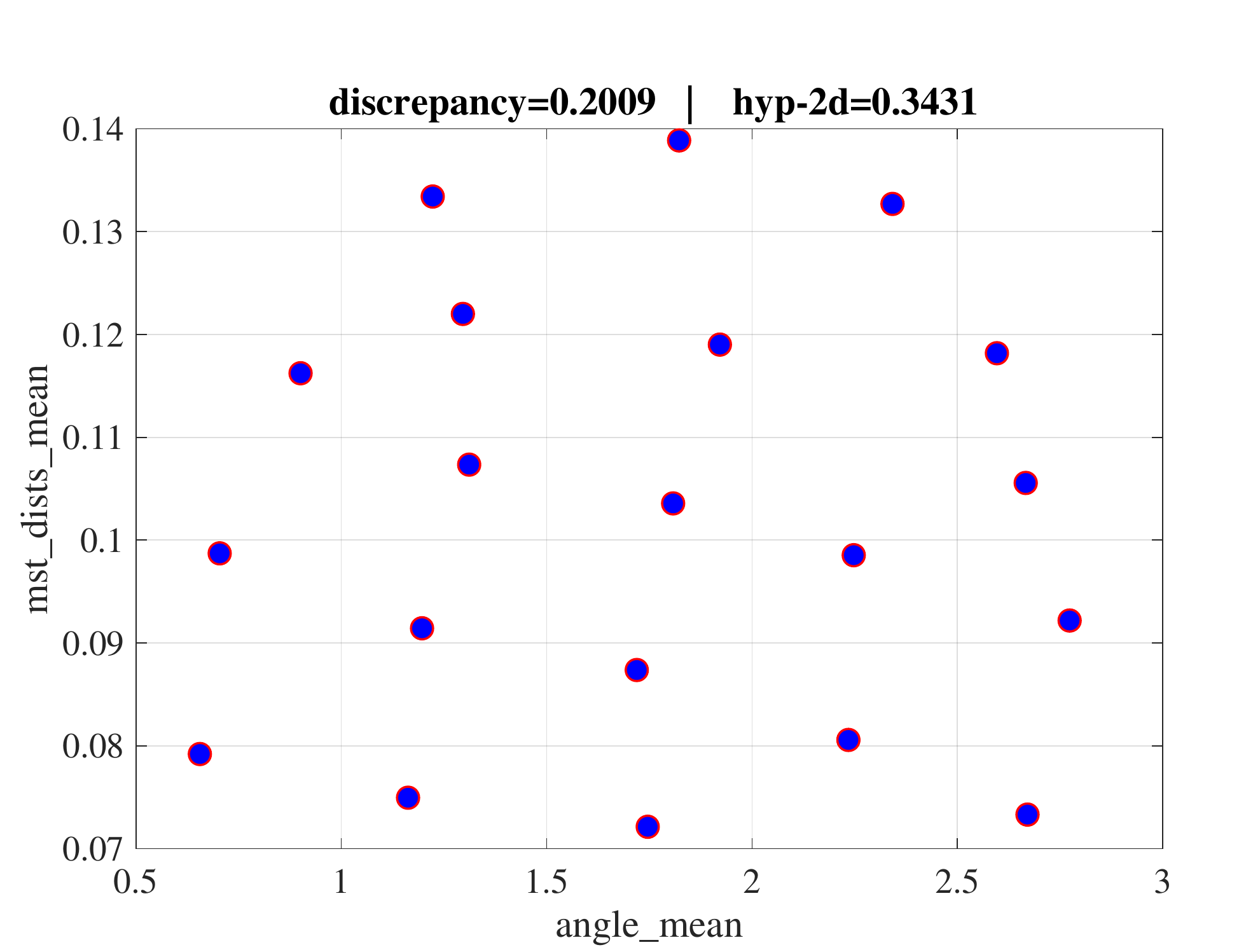}%
\hspace{-0.3cm}
\includegraphics[trim={0.5cm 7.7cm 0.7cm 6.8cm},clip,width=0.32\textwidth]{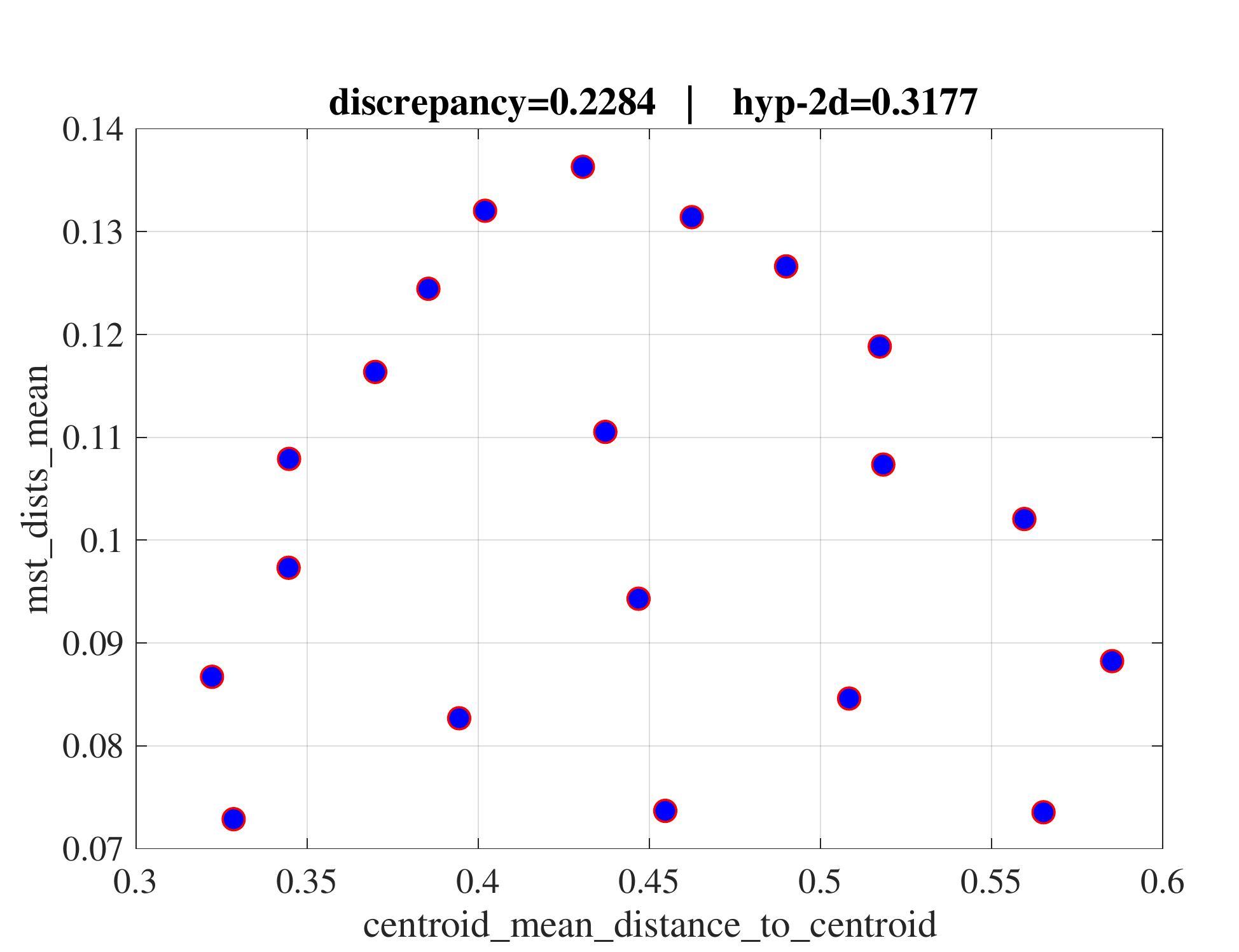}%
\includegraphics[trim={0.5cm 7.7cm 0.7cm 6.8cm},clip,width=0.32\textwidth]{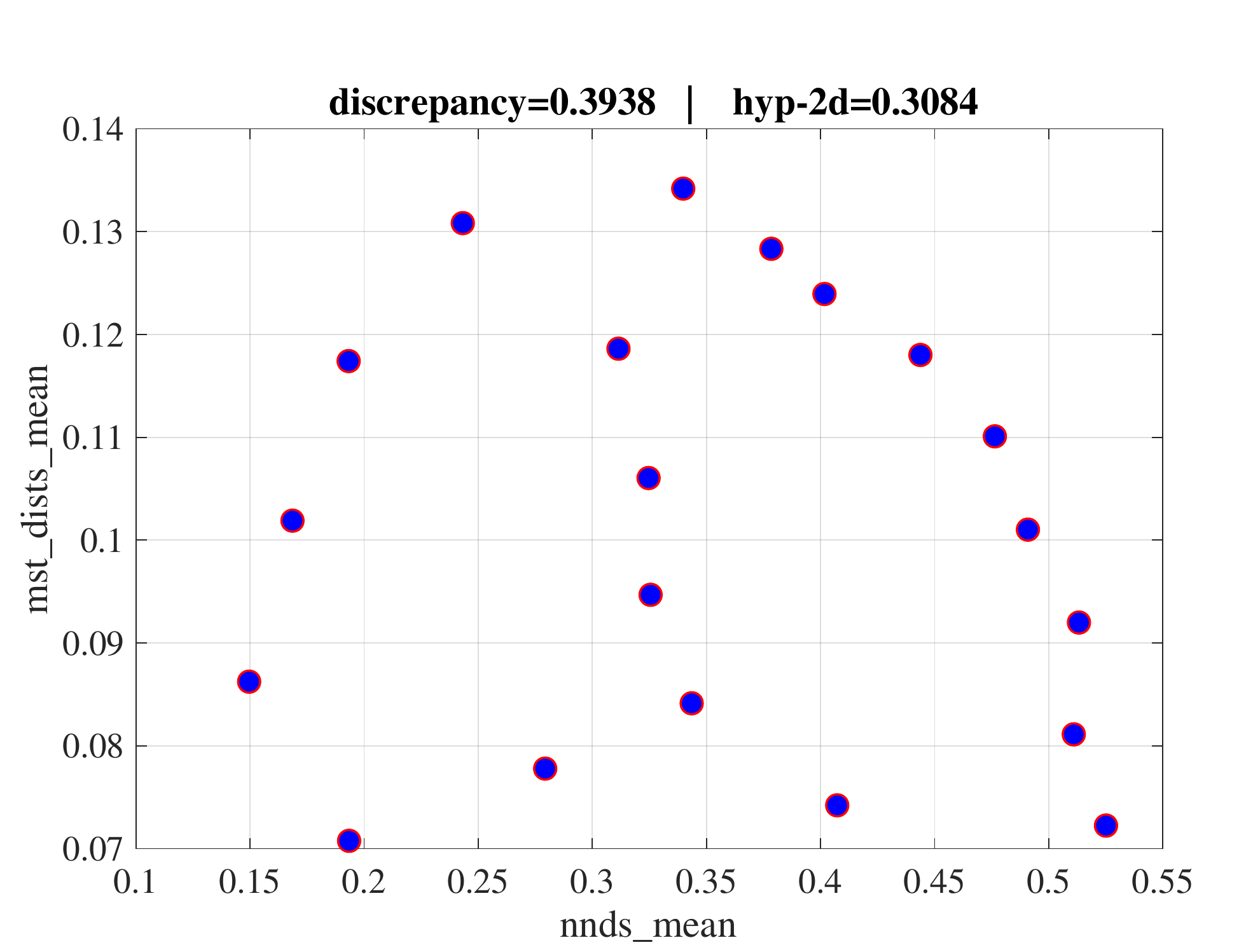}%
\\
\rotatebox{90}{\hspace{11mm}EA$_{\text{HYP}}$} \rotatebox{90}{\rule{31mm}{1pt}} 
\includegraphics[trim={0.5cm 7.7cm 0.7cm 6.8cm},clip,width=0.32\textwidth]{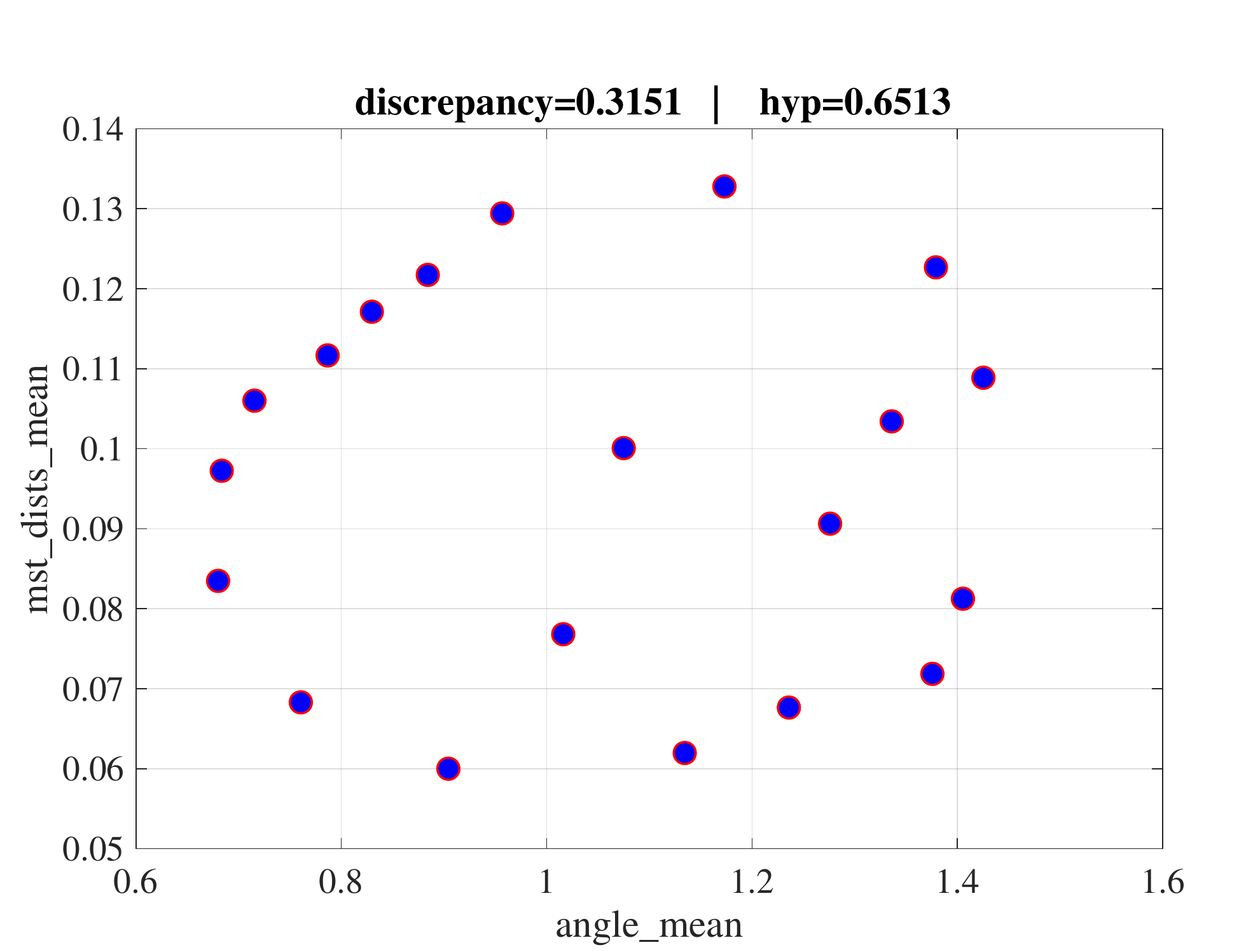}%
\hspace{-0.3cm}
\includegraphics[trim={0.5cm 7.7cm 0.7cm 6.8cm},clip,width=0.32\textwidth]{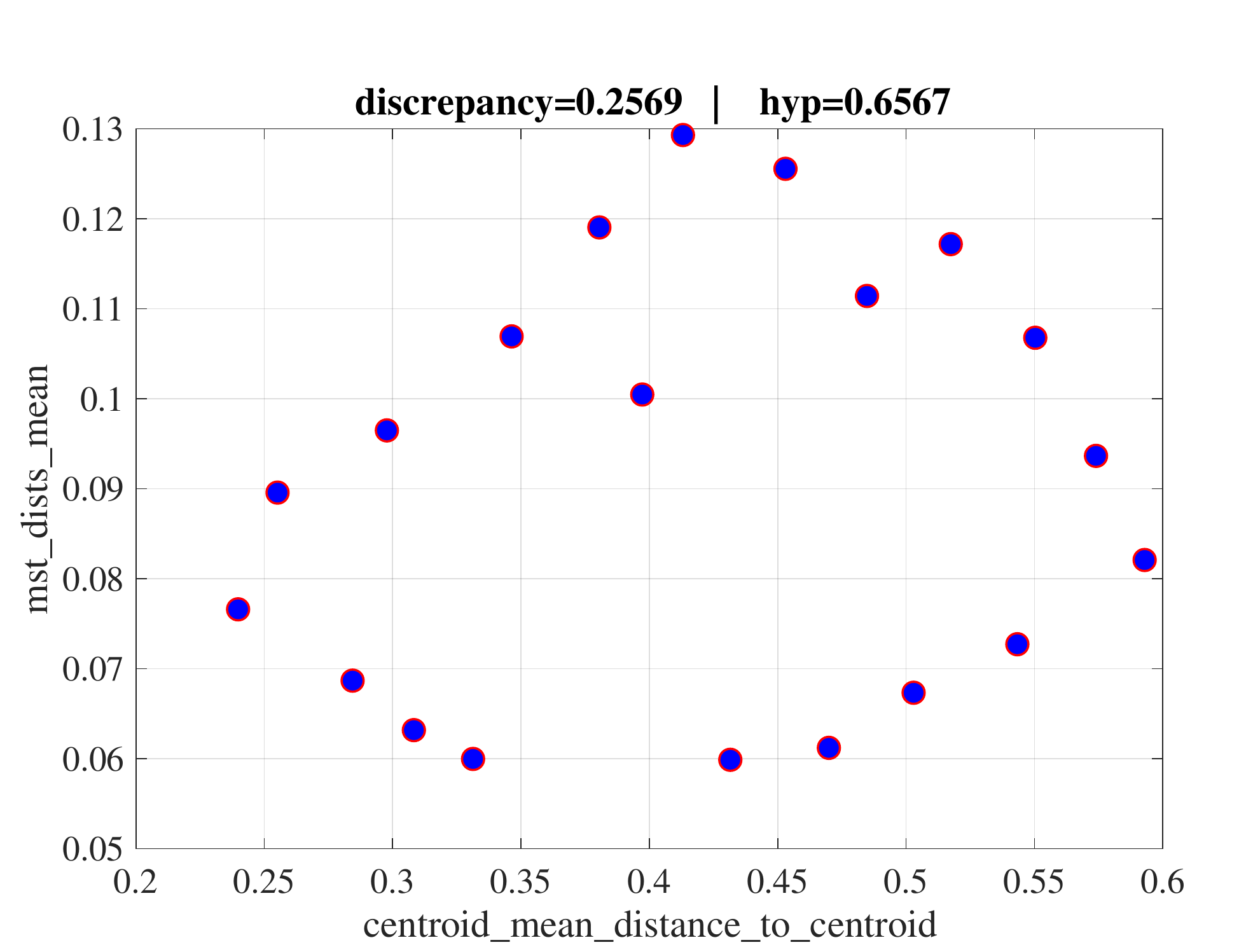}%
\includegraphics[trim={0.5cm 7.7cm 0.7cm 6.8cm},clip,width=0.32\textwidth]{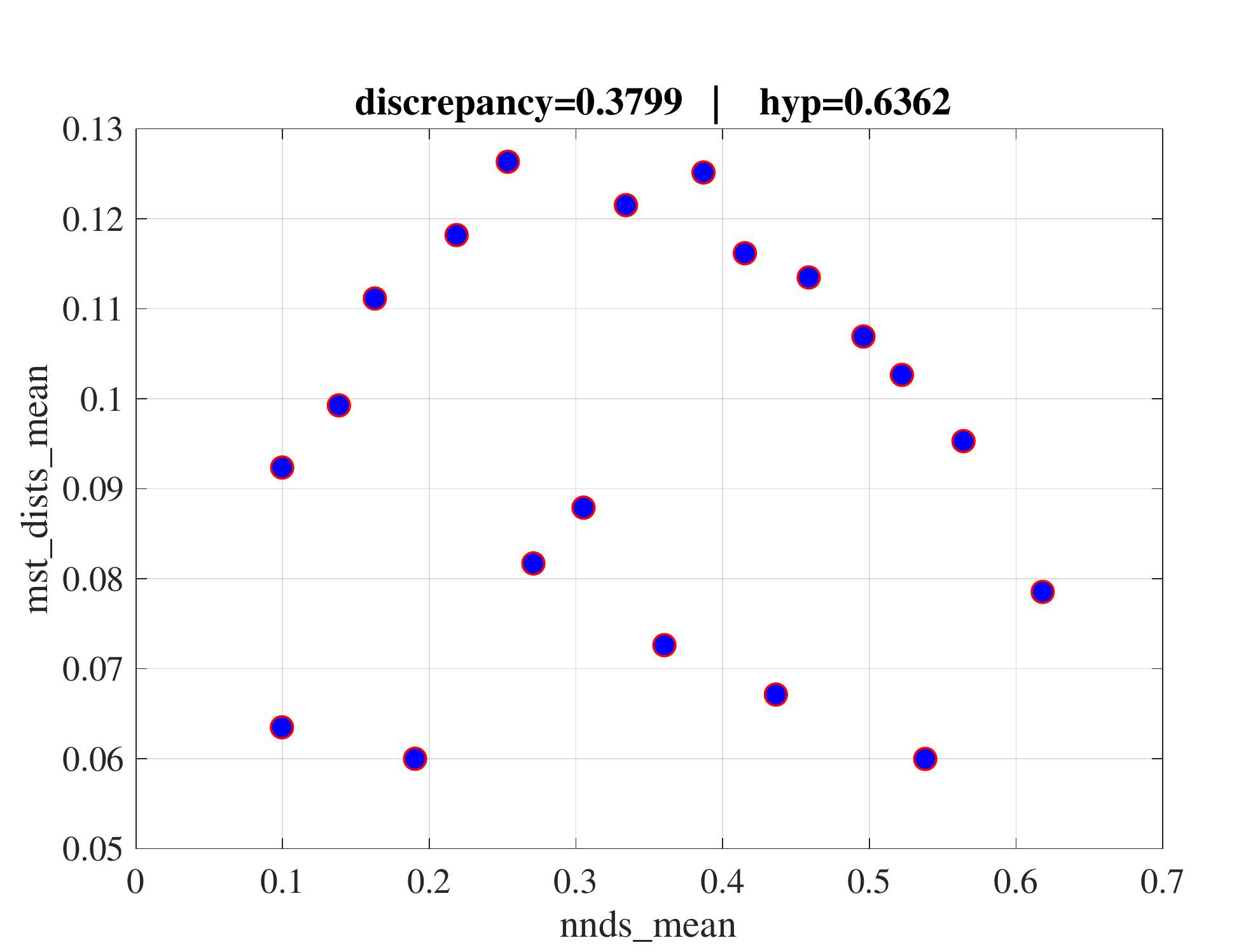}%
\\
\rotatebox{90}{\hspace{12mm}EA$_{\text{IGD}}$} \rotatebox{90}{\rule{31mm}{1pt}} 
\includegraphics[trim={0.5cm 7.7cm 0.7cm 6.8cm},clip,width=0.32\textwidth]{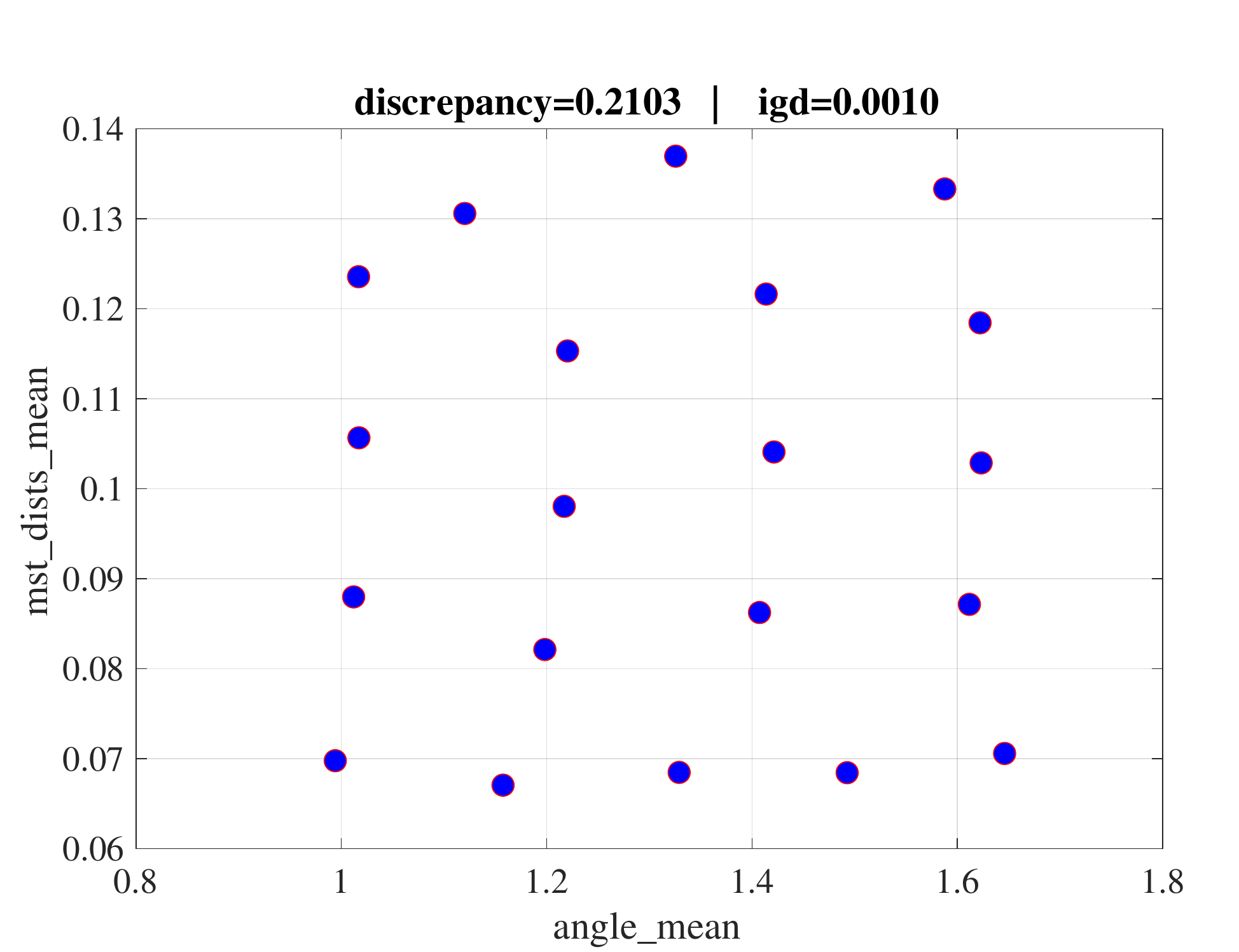}%
\hspace{-0.3cm}
\includegraphics[trim={0.5cm 7.7cm 0.7cm 6.8cm},clip,width=0.32\textwidth]{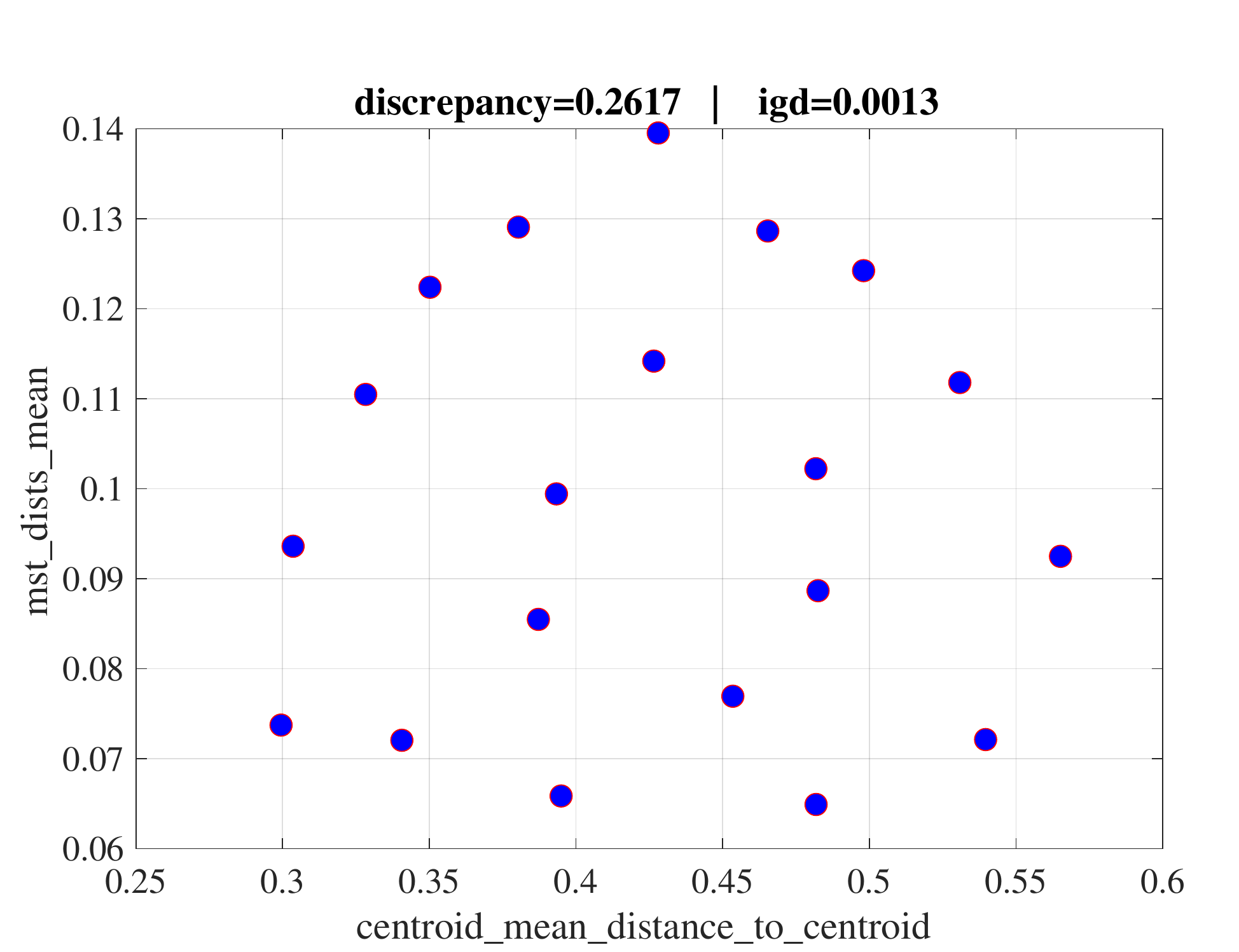}%
\includegraphics[trim={0.5cm 7.7cm 0.7cm 6.8cm},clip,width=0.32\textwidth]{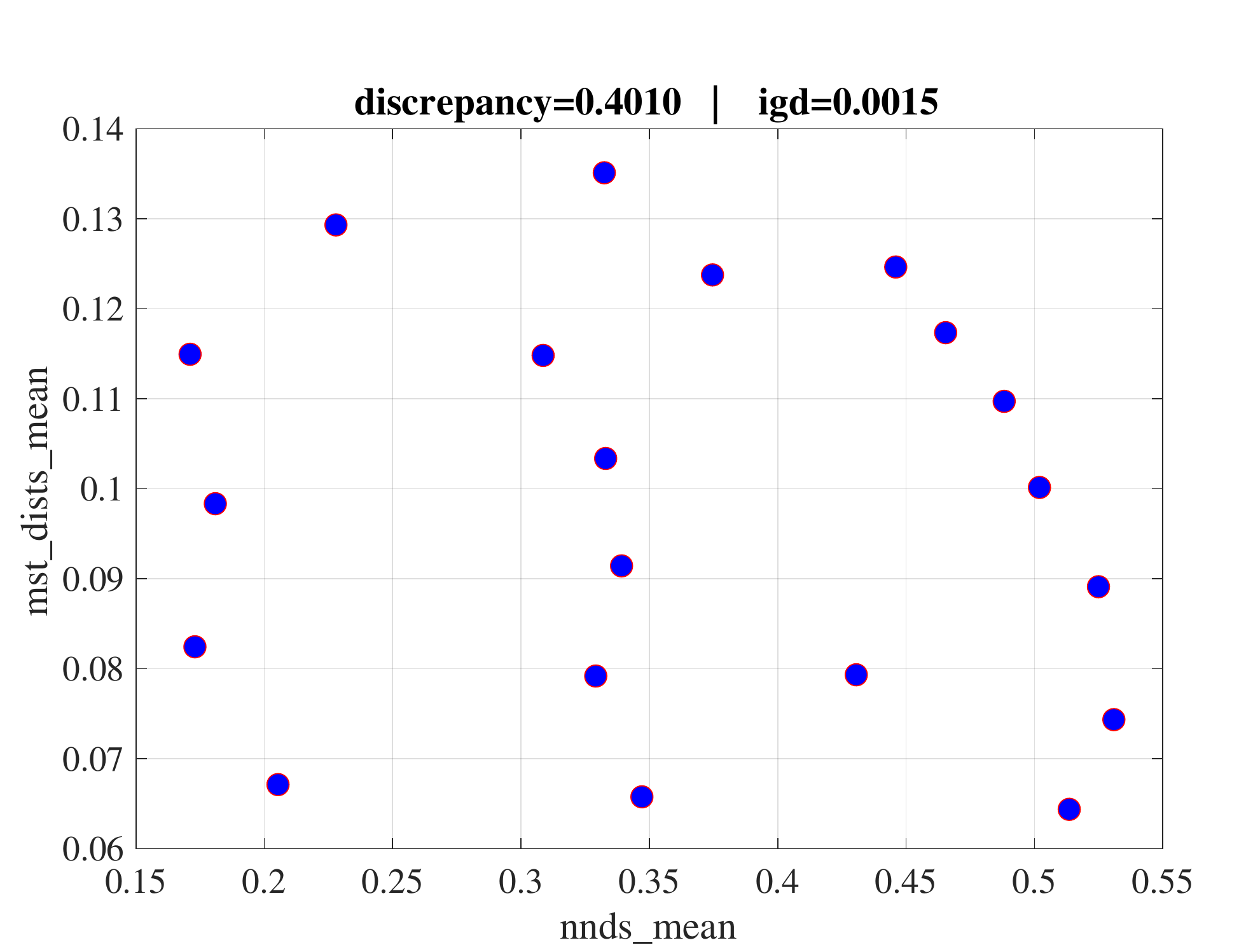}%
\\
\rotatebox{90}{\hspace{12mm}EA$_{\text{EPS}}$} \rotatebox{90}{\rule{31mm}{1pt}} 
\includegraphics[trim={0.5cm 7.7cm 0.7cm 6.8cm},clip,width=0.32\textwidth]{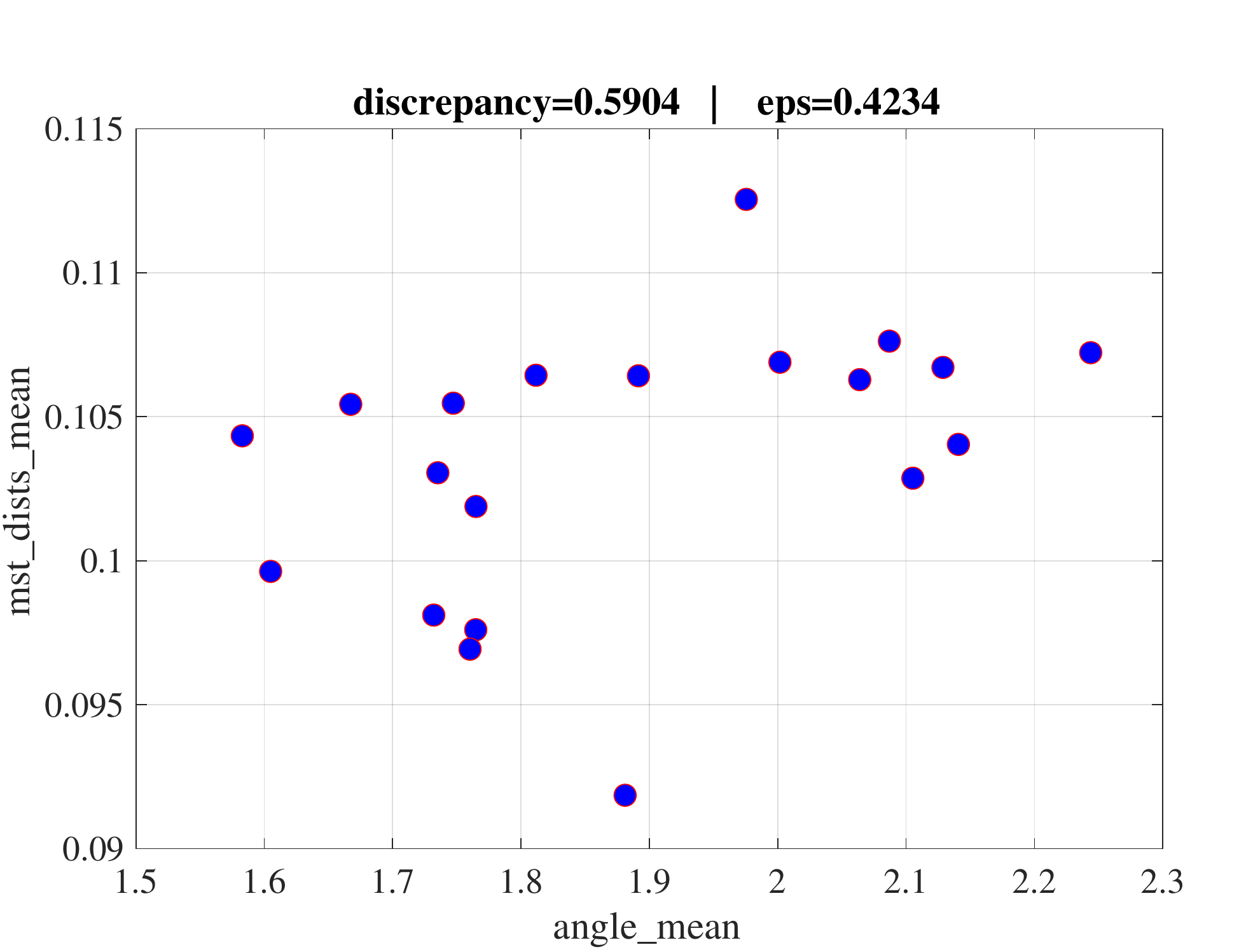}
\hspace{-0.3cm}
\includegraphics[trim={0.5cm 7.7cm 0.7cm 6.8cm},clip,width=0.32\textwidth]{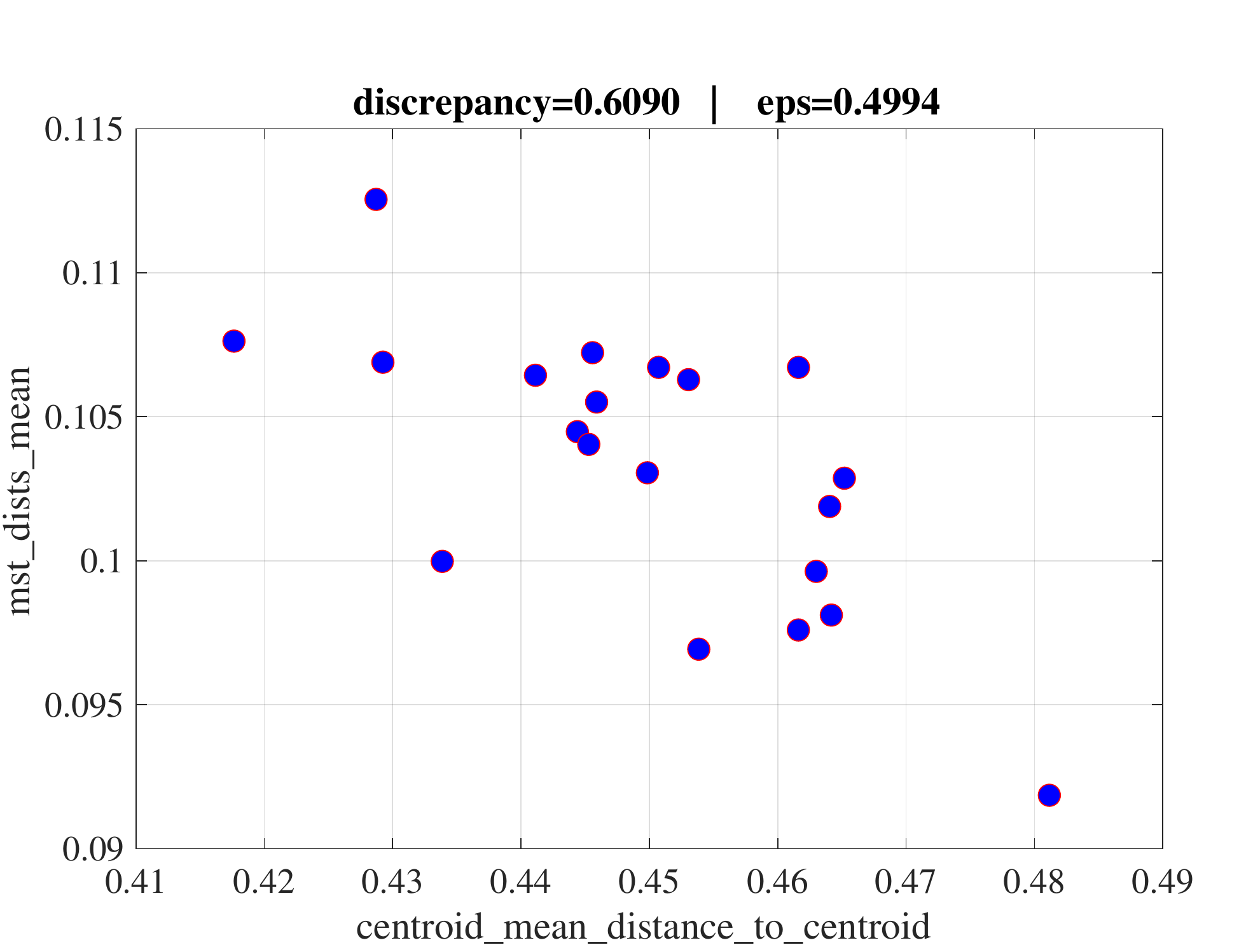}%
\includegraphics[trim={0.5cm 7.7cm 0.7cm 6.8cm},clip,width=0.32\textwidth]{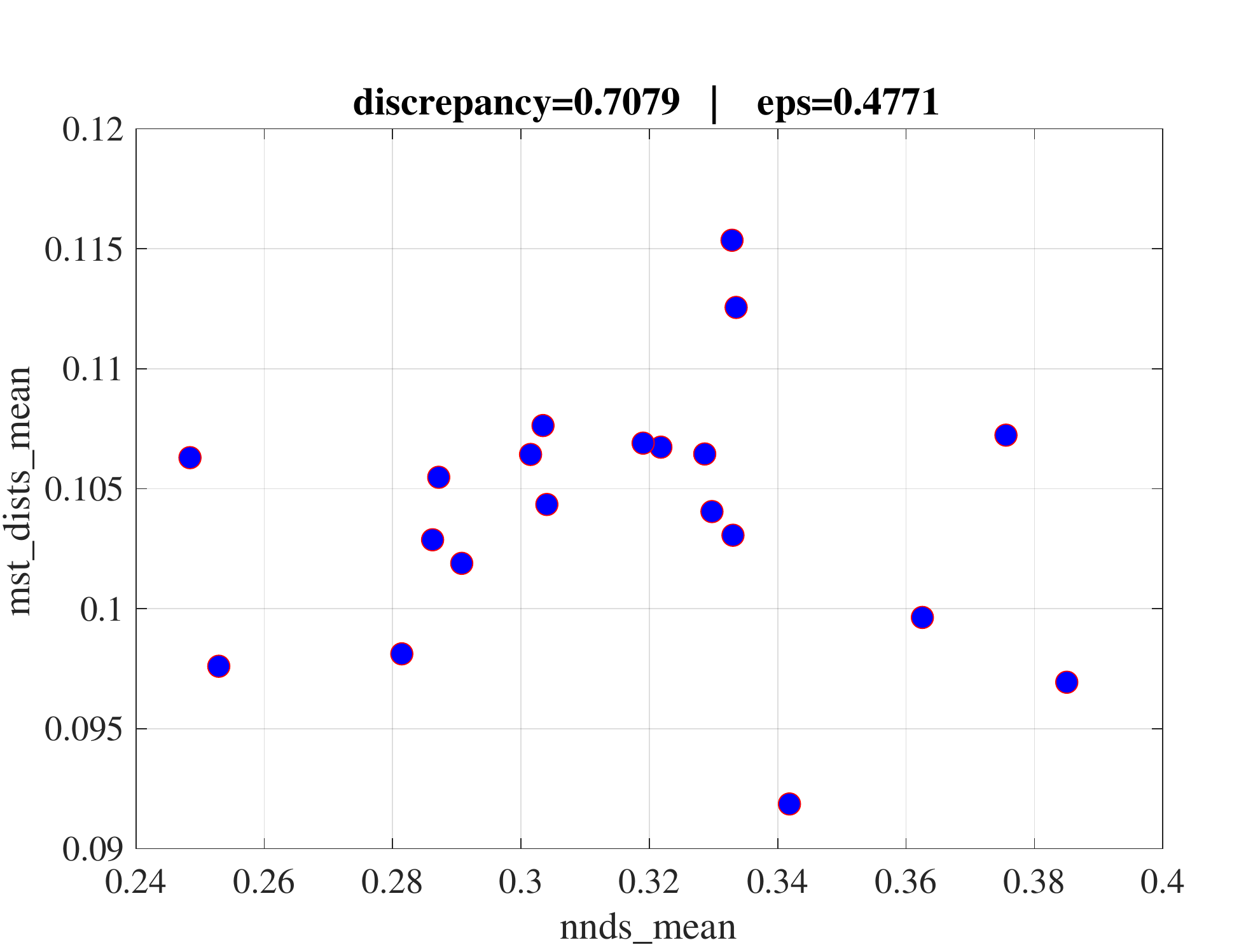}%
\caption{Feature vectors for final population of EA$_{\text{\HYPR}}$ (top), EA$_{\text{HYP}}$ (2nd), EA$_{\text{IGD}}$ (3rd) and EA$_{\text{EPS}}$ (bottom) for TSP instances based on two features from left to right: ($f_1$, $f_4$), ($f_2$, $f_4$), ($f_3$, $f_4$).}
\label{fig:tsp-2d}
\end{figure}

In this study, our goal is to generate diverse sets of TSP instances with 50 cities in the space of $[0,1]^2$, which is a reasonable size of problem for feature analysis of TSP. The instance quality is evaluated by the approximation ratio, which is calculated by  
\[
  \alpha_A(I) = A(I) / OPT(I),
\]
where $A(I)$ is the fitness value of the solution found by algorithm $A$ for the given instance $I$, and $OPT(I)$ is the size of an optimal solution for instance $I$ which in our case is calculated using the exact TSP solver Concorde~\cite{Applegate02}. Within this study, $A(I)$ is the minimum tour length obtained by three independent repeated runs of the 2-OPT algorithm for a given TSP instance $I$. As the number of cities in an instance is $50$, our algorithm chooses $1.18$ as threshold for approximation ratio, which means only TSP instances with approximation ratios equal to or greater than $1.18$ are accepted; this follows the setting in~\cite{DBLP:conf/ppsn/GaoNN16}.

\begin{sidewaystable}
\centering
\caption{Investigations for TSP instances with $2$ features. 
Comparison in terms of mean, standard deviation and statistical test for considered indicators.}
\label{tb:tsp-stat}
\renewcommand*{\arraystretch}{1.2}\setlength{\tabcolsep}{1mm}\resizebox{\textwidth}{!}{ 
\begin{tabular}{lclllllllllllllll}
                     &                            & \multicolumn{3}{c}{EA$_{\text{\HYPR}}$ (1)}                                                                     & \multicolumn{3}{c}{EA$_{\text{HYP}}$ (2)}                                                                     & \multicolumn{3}{c}{EA$_{\text{IGD}}$ (3)}                                                                     & \multicolumn{3}{c}{EA$_{\text{EPS}}$ (4)} & \multicolumn{3}{c}{EA$_{\text{DIS}}$ (5)}                     \\
                     &                            & \multicolumn{1}{c}{mean} & \multicolumn{1}{c}{st} & \multicolumn{1}{c}{stat}                           & \multicolumn{1}{c}{mean} & \multicolumn{1}{c}{st} & \multicolumn{1}{c}{stat}                           & \multicolumn{1}{c}{mean} & \multicolumn{1}{c}{st} & \multicolumn{1}{c}{stat}     & \multicolumn{1}{c}{mean} & \multicolumn{1}{c}{st} & \multicolumn{1}{c}{stat}                      & \multicolumn{1}{c}{mean} & \multicolumn{1}{c}{st} & \multicolumn{1}{c}{stat}      \\ 
\cline{3-17} 
\multicolumn{1}{l}{\multirow{3}{*}{\rotatebox{90}{\HYPR}}} & \multicolumn{1}{l|}{$f_1$,$f_4$} & 0.338 & 2E-3 & \multicolumn{1}{l|}{$2^{(+)}$,$4^{(+)}$,$5^{(+)}$} & 0.309 & 4E-3 & \multicolumn{1}{l|}{$1^{(-)}$,$4^{(+)}$} & 0.331 & 3E-3 & \multicolumn{1}{l|}{$4^{(+)}$,$5^{(+)}$} & 0.190 & 1E-3 & \multicolumn{1}{l|}{$1^{(-)}$,$2^{(-)}$,$3^{(-)}$} & 0.256 & 1E-2 & $1^{(-)}$,$3^{(-)}$ \\

\multicolumn{1}{l}{} & \multicolumn{1}{l|}{$f_2$,$f_4$} & 0.317 & 3E-3 & \multicolumn{1}{l|}{$2^{(+)}$,$4^{(+)}$,$5^{(+)}$}  & 0.303 & 5E-3 & \multicolumn{1}{l|}{$1^{(-)}$,$3^{(-)}$,$4^{(+)}$} & 0.316 & 3E-3 & \multicolumn{1}{l|}{$2^{(+)}$,$4^{(+)}$,$5^{(+)}$} & 0.178 & 1E-7 & \multicolumn{1}{l|}{$1^{(-)}$,$2^{(-)}$,$3^{(-)}$} & 0.252 & 1E-2 & $1^{(-)}$,$3^{(-)}$  \\

\multicolumn{1}{l}{} & \multicolumn{1}{l|}{$f_3$,$f_4$} & 0.303 & 2E-2 & \multicolumn{1}{l|}{$2^{(+)}$,$4^{(+)}$,$5^{(+)}$}  & 0.296 & 5E-3& \multicolumn{1}{l|}{$1^{(-)}$,$3^{(-)}$,$4^{(+)}$,$5^{(+)}$} & 0.304 & 2E-2 & \multicolumn{1}{l|}{$2^{(+)}$,$4^{(+)}$,$5^{(+)}$} & 0.190 & 2E-3 & \multicolumn{1}{l|}{$1^{(-)}$,$2^{(-)}$,$3^{(-)}$} & 0.238 & 2E-2 & $1^{(-)}$,$2^{(-)}$,$3^{(-)}$ 

\\ \cmidrule(l){2-17}

\multicolumn{1}{l}{\multirow{3}{*}{\rotatebox{90}{HYP}}} & \multicolumn{1}{l|}{$f_1$,$f_4$} & 0.645 & 5E-3 & \multicolumn{1}{l|}{$4^{(+)}$,$5^{(+)}$} & 0.638 & 7E-3 & \multicolumn{1}{l|}{$4^{(+)}$,$5^{(+)}$}  & 0.639 & 6E-3 & \multicolumn{1}{l|}{$4^{(+)}$,$5^{(+)}$} & 0.424 & 2E-3 & \multicolumn{1}{l|}{$1^{(-)}$,$2^{(-)}$,$3^{(-)}$} & 0.529 & 3E-2 & $1^{(-)}$,$2^{(-)}$,$3^{(-)}$  \\

\multicolumn{1}{l}{} & \multicolumn{1}{l|}{$f_2$,$f_4$} & 0.609 & 7E-3 & \multicolumn{1}{l|}{$2^{(-)}$,$4^{(+)}$,$5^{(+)}$} & 0.632 & 1E-2 & \multicolumn{1}{l|}{$1^{(+)}$,$4^{(+)}$,$5^{(+)}$} & 0.621 & 6E-3 & \multicolumn{1}{l|}{$4^{(+)}$,$5^{(+)}$} & 0.398 & 1E-6 & \multicolumn{1}{l|}{$1^{(-)}$,$2^{(-)}$,$3^{(-)}$} & 0.505 & 2E-2 & $1^{(-)}$,$2^{(-)}$,$3^{(-)}$ \\

\multicolumn{1}{l}{} & \multicolumn{1}{l|}{$f_3$,$f_4$} & 0.584 & 3E-2 & \multicolumn{1}{l|}{$2^{(-)}$,$4^{(+)}$} & 0.621 & 9E-3 & \multicolumn{1}{l|}{$1^{(+)}$,$3^{(+)}$,$4^{(+)}$,$5^{(+)}$} & 0.595 & 4E-2 & \multicolumn{1}{l|}{$2^{(-)}$,$4^{(+)}$,$5^{(+)}$} & 0.410 & 2E-3 & \multicolumn{1}{l|}{$1^{(-)}$,$2^{(-)}$,$3^{(-)}$} & 0.485 & 3E-2 &$2^{(-)}$,$3^{(-)}$
\\\cmidrule(l){2-17} 
\multicolumn{1}{l}{\multirow{3}{*}{\rotatebox{90}{IGD}}} & \multicolumn{1}{l|}{$f_1$,$f_4$} & 0.001 & 2E-5 & \multicolumn{1}{l|}{$4^{(+)}$,$5^{(+)}$}  & 0.001 & 6E-5 &  \multicolumn{1}{l|}{$3^{(-)}$,$4^{(+)}$} & 0.001 & 4E-5 & \multicolumn{1}{l|}{$2^{(+)}$,$4^{(+)}$,$5^{(+)}$} & 0.003 & 2E-5 & \multicolumn{1}{l|}{$1^{(-)}$,$2^{(-)}$,$3^{(-)}$} & 0.002 & 2E-4 & $1^{(-)}$,$3^{(-)}$ \\

\multicolumn{1}{l}{} & \multicolumn{1}{l|}{$f_2$,$f_4$} & 0.001 & 3E-5 & \multicolumn{1}{l|}{$2^{(+)}$,$4^{(+)}$,$5^{(+)}$} & 0.002 & 6E-5 & \multicolumn{1}{l|}{$1^{(-)}$,$3^{(-)}$,$4^{(+)}$} & 0.001 & 3E-5 & \multicolumn{1}{l|}{$2^{(+)}$,$4^{(+)}$,$5^{(+)}$} & 0.003 & 2E-10 & \multicolumn{1}{l|}{$1^{(-)}$,$2^{(-)}$,$3^{(-)}$} & 0.002 & 2E-4 & $1^{(-)}$,$3^{(-)}$ \\

\multicolumn{1}{l}{} & \multicolumn{1}{l|}{$f_3$,$f_4$} & 0.002 & 3E-4 & \multicolumn{1}{l|}{$4^{(+)}$,$5^{(+)}$} & 0.002 & 6E-5 & \multicolumn{1}{l|}{$3^{(-)}$,$4^{(+)}$,$5^{(+)}$} & 0.002 & 3E-4 & \multicolumn{1}{l|}{$2^{(+)}$,$4^{(+)}$,$5^{(+)}$} & 0.003 & 3E-5 & \multicolumn{1}{l|}{$1^{(-)}$,$2^{(-)}$,$3^{(-)}$} & 0.003 & 3E-4 & $1^{(-)}$,$2^{(-)}$,$3^{(-)}$ \\ \cmidrule(l){2-17}

\multicolumn{1}{l}{\multirow{3}{*}{\rotatebox{90}{EPS}}} & \multicolumn{1}{l|}{$f_1$,$f_4$} & 0.196 & 2E-2 & \multicolumn{1}{l|}{$2^{(+)}$,$4^{(+)}$,$5^{(+)}$} & 0.249 & 2E-2 & \multicolumn{1}{l|}{$1^{(-)}$,$3^{(-)}$,$4^{(+)}$} & 0.189 & 2E-2 & \multicolumn{1}{l|}{$2^{(+)}$,$4^{(+)}$,$5^{(+)}$} & 0.423 & 1E-3 & \multicolumn{1}{l|}{$1^{(-)}$,$2^{(-)}$,$3^{(-)}$} & 0.345 & 4E-2 & $1^{(-)}$,$3^{(-)}$  \\

\multicolumn{1}{l}{} & \multicolumn{1}{l|}{$f_2$,$f_4$} & 0.226 & 8E-3 & \multicolumn{1}{l|}{$2^{(+)}$,$4^{(+)}$,$5^{(+)}$} & 0.256 & 2E-2 & \multicolumn{1}{l|}{$1^{(-)}$,$3^{(-)}$,$4^{(+)}$,$5^{(+)}$} & 0.228 & 1E-2 & \multicolumn{1}{l|}{$2^{(+)}$,$4^{(+)}$,$5^{(+)}$} & 0.499 & 2E-16 & \multicolumn{1}{l|}{$1^{(-)}$,$2^{(-)}$,$3^{(-)}$} & 0.360 & 5E-2 & $1^{(-)}$,$2^{(-)}$,$3^{(-)}$ \\

\multicolumn{1}{l}{} & \multicolumn{1}{l|}{$f_3$,$f_4$} & 0.260 & 4E-2 & \multicolumn{1}{l|}{$4^{(+)}$,$5^{(+)}$} & 0.278 & 2E-2 & \multicolumn{1}{l|}{$4^{(+)}$,$5^{(+)}$} & 0.265 & 4E-2 & \multicolumn{1}{l|}{$4^{(+)}$,$5^{(+)}$} & 0.477 & 3E-3 & \multicolumn{1}{l|}{$1^{(-)}$,$2^{(-)}$,$3^{(-)}$} & 0.368 & 5E-2 & $1^{(-)}$,$2^{(-)}$,$3^{(-)}$
\\ \cmidrule(l){2-17}

\multicolumn{1}{l}{\multirow{3}{*}{\rotatebox{90}{DIS}}} & \multicolumn{1}{l|}{$f_1$,$f_4$} & 0.222 & 2E-2 & \multicolumn{1}{l|}{$2^{(+)}$,$4^{(+)}$,$5^{(+)}$} & 0.353 & 2E-2 & \multicolumn{1}{l|}{$1^{(-)}$,$3^{(-)}$,$4^{(+)}$} & 0.249 & 2E-2 & \multicolumn{1}{l|}{$2^{(+)}$,$4^{(+)}$} & 0.589 & 4E-3 & \multicolumn{1}{l|}{$1^{(-)}$,$2^{(-)}$,$3^{(-)}$,$5^{(-)}$} & 0.292 & 5E-2 & $1^{(-)}$,$4^{(+)}$  \\

\multicolumn{1}{l}{} & \multicolumn{1}{l|}{$f_2$,$f_4$} & 0.230 & 2E-2 & \multicolumn{1}{l|}{$2^{(+)}$,$4^{(+)}$,$5^{(+)}$}  & 0.274 & 2E-2 & \multicolumn{1}{l|}{$1^{(-)}$,$4^{(+)}$,$5^{(+)}$} & 0.252 & 1E-3 & \multicolumn{1}{l|}{$4^{(+)}$,$5^{(+)}$} & 0.609 & 1E-16 & \multicolumn{1}{l|}{$1^{(-)}$,$2^{(-)}$,$3^{(-)}$,$5^{(-)}$} & 0.336 & 4E-2 &  $1^{(-)}$,$2^{(-)}$,$3^{(-)}$,$4^{(+)}$ \\

\multicolumn{1}{l}{} & \multicolumn{1}{l|}{$f_3$,$f_4$} & 0.418 & 6E-2 & \multicolumn{1}{l|}{$4^{(+)}$}  & 0.416 & 3E-2 &  \multicolumn{1}{l|}{$4^{(+)}$} & 0.401 & 7E-2 & \multicolumn{1}{l|}{$4^{(+)}$,$5^{(+)}$} & 0.719 & 6E-3 & \multicolumn{1}{l|}{$1^{(-)}$,$2^{(-)}$,$3^{(-)}$,$5^{(-)}$} & 0.448 & 9E-2 &$3^{(-)}$,$4^{(+)}$

\end{tabular}
}
\end{sidewaystable}

\subsection{Experimental settings}

The algorithm is implemented in R and run in R environment~\cite{rManual}. The feature vectors are calculated using the tspmeta package~\cite{Mersmann2013}. The hardware is identical to that used in the image-related experiments.
The features we use to characterize TSP instances are as follows: angle\_mean, centroid\_mean\_distance\_to\_centroid, nnds\_mean, and mst\_dists\_mean (see Table~\ref{tab:featuresTSP}).
The parameter setting follows the same setting as in~\cite{DBLP:journals/corr/abs-1802-05448}. The population size $\mu$ and number of offspring generated $\lambda$ of EA is set to $20$ and $1$ respectively. 

As mentioned before in Section~\ref{sec:ind}, we normalize feature values before indicator calculations. Based on the results gathered from some initial runs of feature-based diversity maximization algorithm, the maximum and minimum values $f^{max}$ and $f^{min}$ for each feature are determined (see Table~\ref{tab:featuresTSP}). Each algorithm setting is repeated independently for $30$ times. Each experiment is run for $20,000$ generations and the values of all proposed indicators and discrepancy values are reported in the following section.

\subsection{Experimental results and analysis}

As before, three pairs of features and three triplets of features are examined. The results are compared with those from the discrepancy minimization algorithm. 

\subsubsection{Two-feature combinations}

Figure~\ref{fig:tsp-2d} shows some (randomly drawn) populations in the feature space after running the corresponding algorithms with consideration of certain two-feature combinations. In these figures, the populations after optimizing the hypervolume and inverted generational distance show good coverage and distribution over the whole space. Compared to the 2D plots from previous research~\cite{DBLP:journals/corr/GaoNN15,DBLP:journals/corr/abs-1802-05448}, the EA maximizing \HYPR is able to generate individuals with feature vectors that are not found in previous research. The feature vectors obtained from EA$_{\text{\HYPR}}$, EA$_{\text{HYP}}$ and EA$_{\text{IGD}}$ are -- in our opinion -- nicely distributed in the space. In respect of indicator values, the population discrepancies of the sample populations from EA$_{\text{\HYPR}}$, EA$_{\text{HYP}}$ and EA$_{\text{IGD}}$ are comparable to those from the algorithm minimizing discrepancy value. Although the discrepancy values are similar, the individuals from these three algorithms are better distributed than the previous results~\cite{DBLP:journals/corr/abs-1802-05448}.

Table~\ref{tb:tsp-stat} lists the results of $30$ independent runs, following the same layout as Table~\ref{tb:statistic-images}. The statistics are gathered from the final populations after running each algorithm on the three different two-feature combinations. The statistical values in the first three large columns are from the EA maximizing \HYPR, HYP and minimizing IGD respectively. The results show that they outperform the evolutionary algorithms minimizing EPS and discrepancy in all four indicators. Both EA$_{\text{\HYPR}}$ and EA$_{\text{IGD}}$ achieve significant improvements in all four indicators after running for $20,000$ generations. It is not a surprise that EA$_{\text{\HYPR}}$ outperforms the other three algorithms in terms of hypervolume covered. It also shows comparable performance in optimizing IGD and other indicators. The same behavior is observed for EA$_{\text{IGD}}$, which outperform EA$_{\text{EPS}}$ and EA$_{\text{DIS}}$ and maximizes HYP relatively well. EA$_{\text{DIS}}$ is designed for the purpose of minimizing the population discrepancy value. However, based on the statistical analysis, it does not obtain better population discrepancy than EA$_{\text{\HYPR}}$ or EA$_{\text{IGD}}$ after $20,000$ generations. 
   
Similar to what we have observed in the image-based study in Section~\ref{sec:images}, the results of EA$_{\text{EPS}}$ are not as good as those from the algorithms optimizing \HYPR, HYP and IGD. No significant improvement in population diversity is achieved using this algorithm. 
We have experimented with target grids of higher resolution to mitigate local-sensitivity issues that exist despite the use of the vector $S_\alpha(R)$, however, the computational costs have been prohibitively high. We conjecture that EA$_{\text{EPS}}$ needs to grow its reference set just like the approximation-guided algorithm AGE~\cite{Wagner2015ageejor} does.

\begin{figure}[!t]
\rotatebox{90}{\hspace{11mm}EA$_{\text{HYP}}$} \rotatebox{90}{\rule{30mm}{1pt}} 
\includegraphics[trim={0cm 7.7cm 0cm 6.8cm},clip,width=0.32\textwidth]{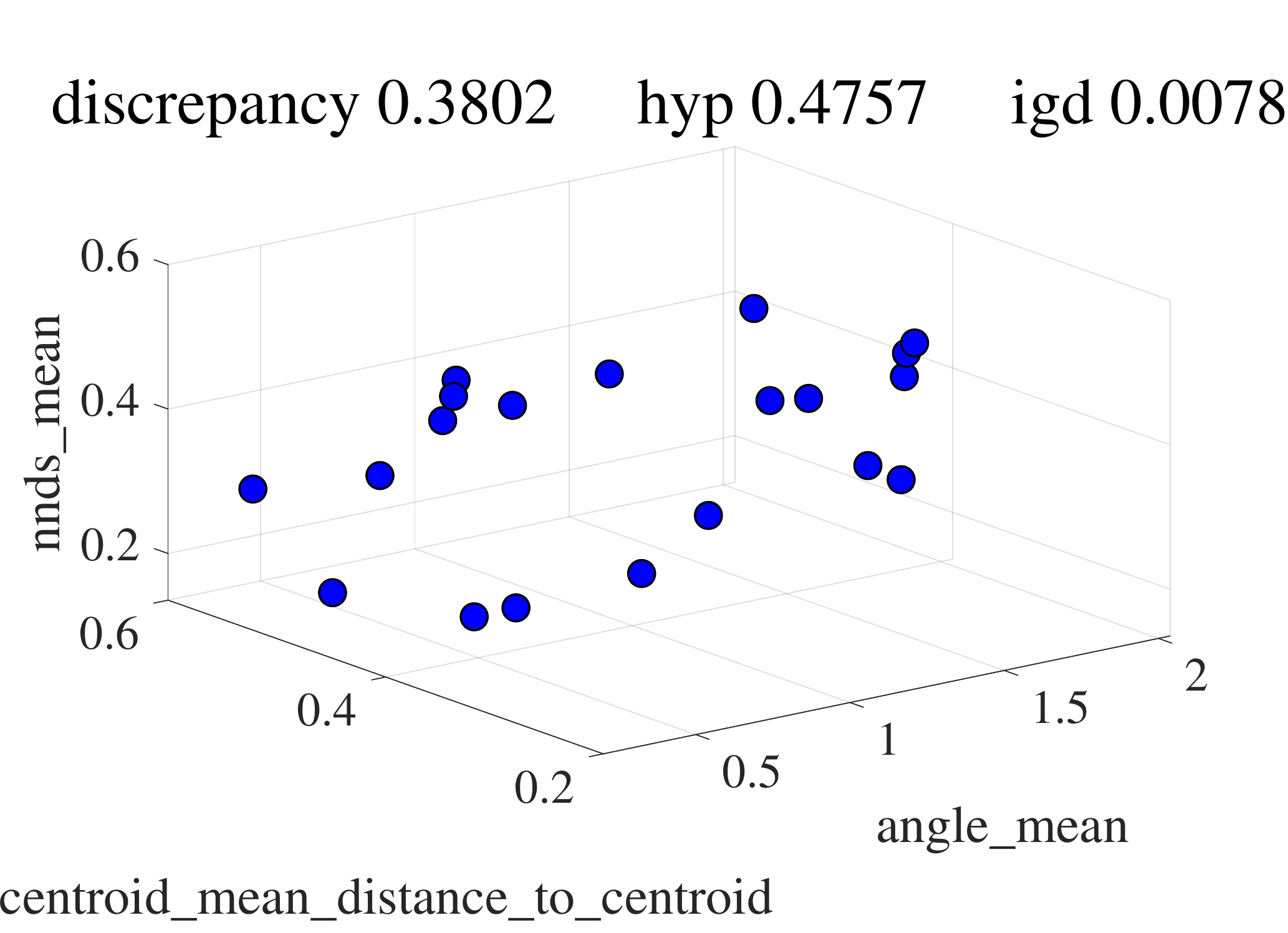}%
\includegraphics[trim={0cm 7.7cm 0cm 6.8cm},clip,width=0.32\textwidth]{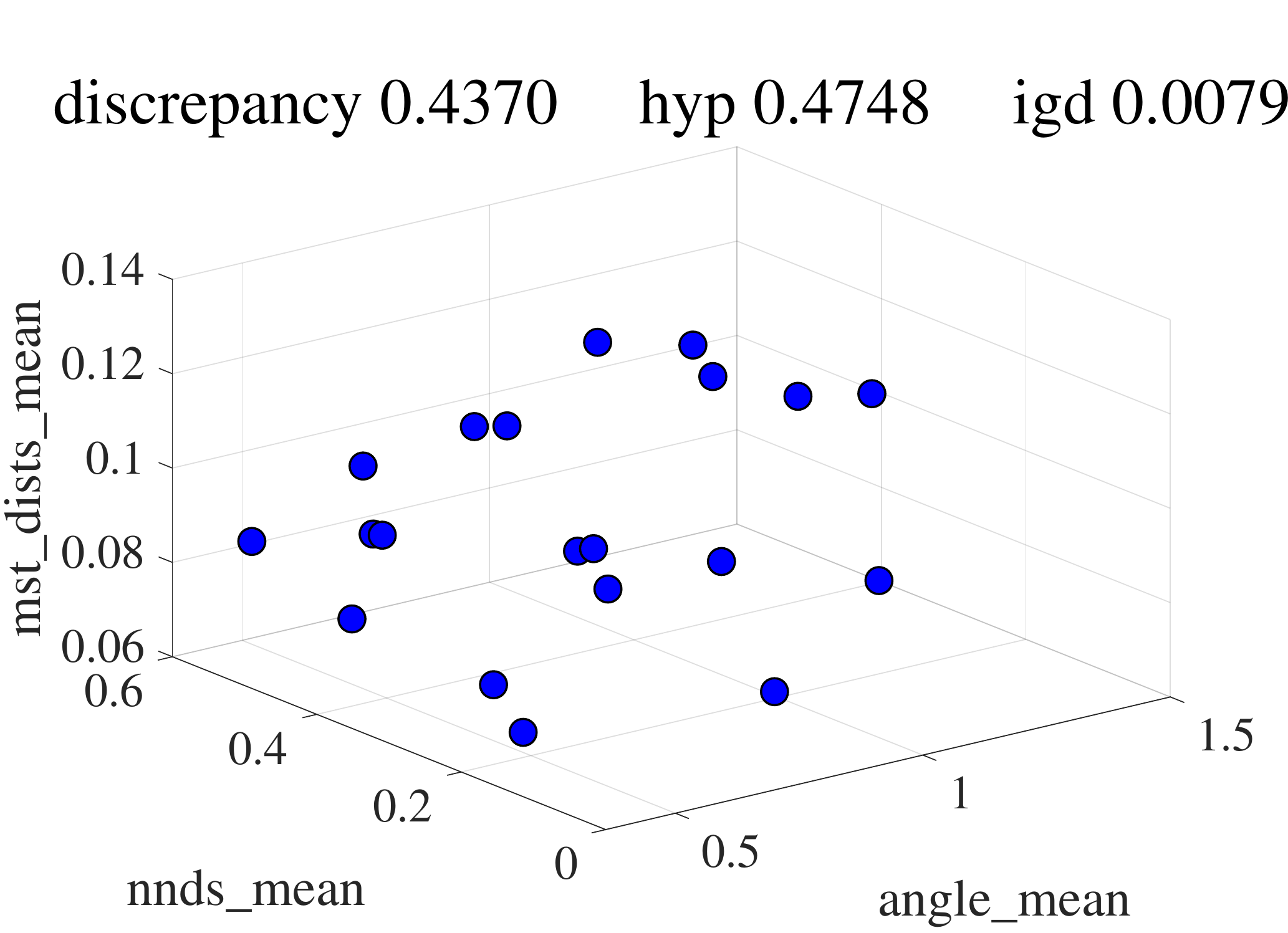}%
\includegraphics[trim={0cm 7.7cm 0cm 6.8cm},clip,width=0.32\textwidth]{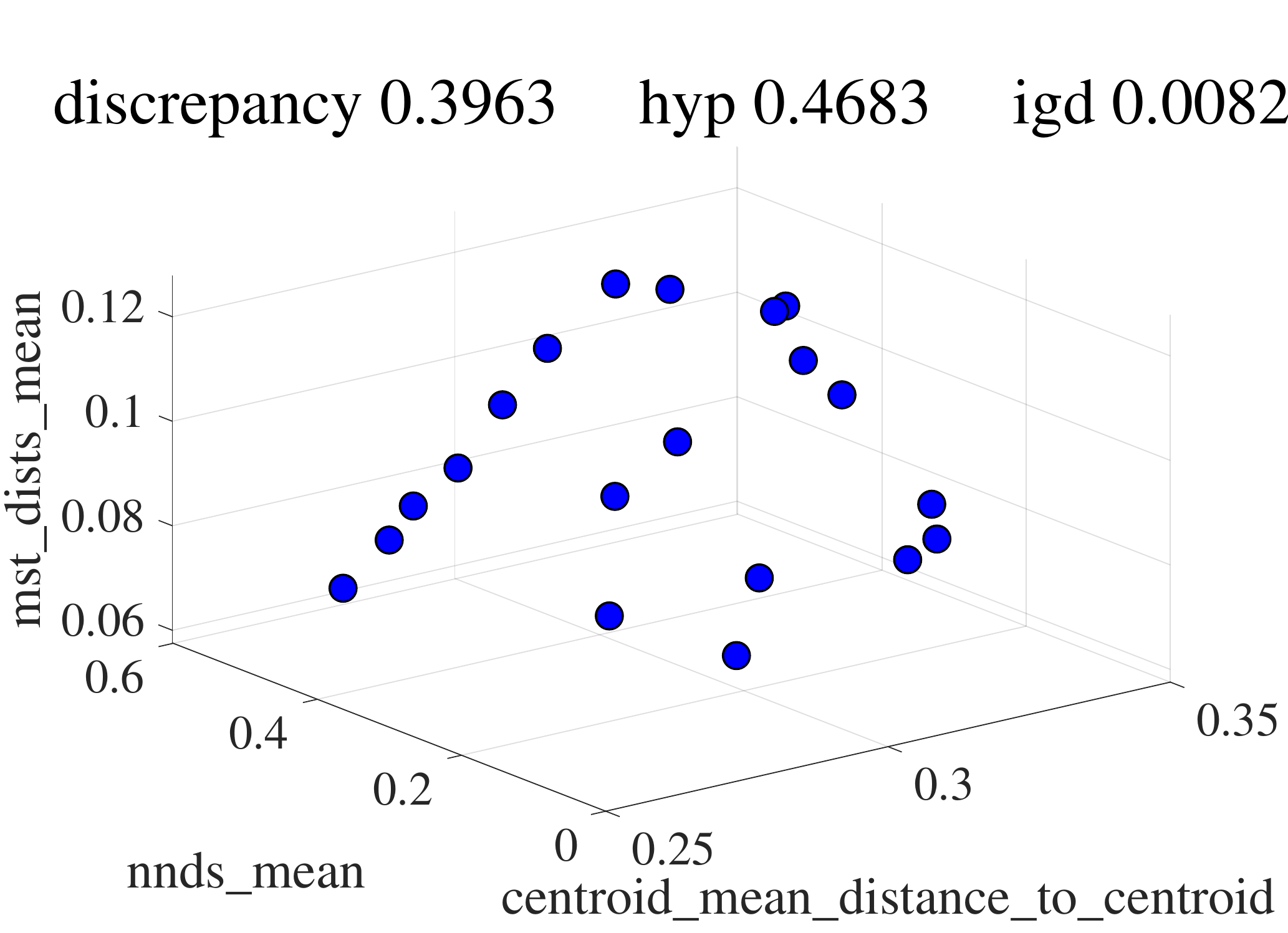}%
\vspace{0.3cm}
\rotatebox{90}{\hspace{11mm}EA$_{\text{IGD}}$} \rotatebox{90}{\rule{30mm}{1pt}} 
\includegraphics[trim={0cm 7.7cm 0cm 6.8cm},clip,width=0.32\textwidth]{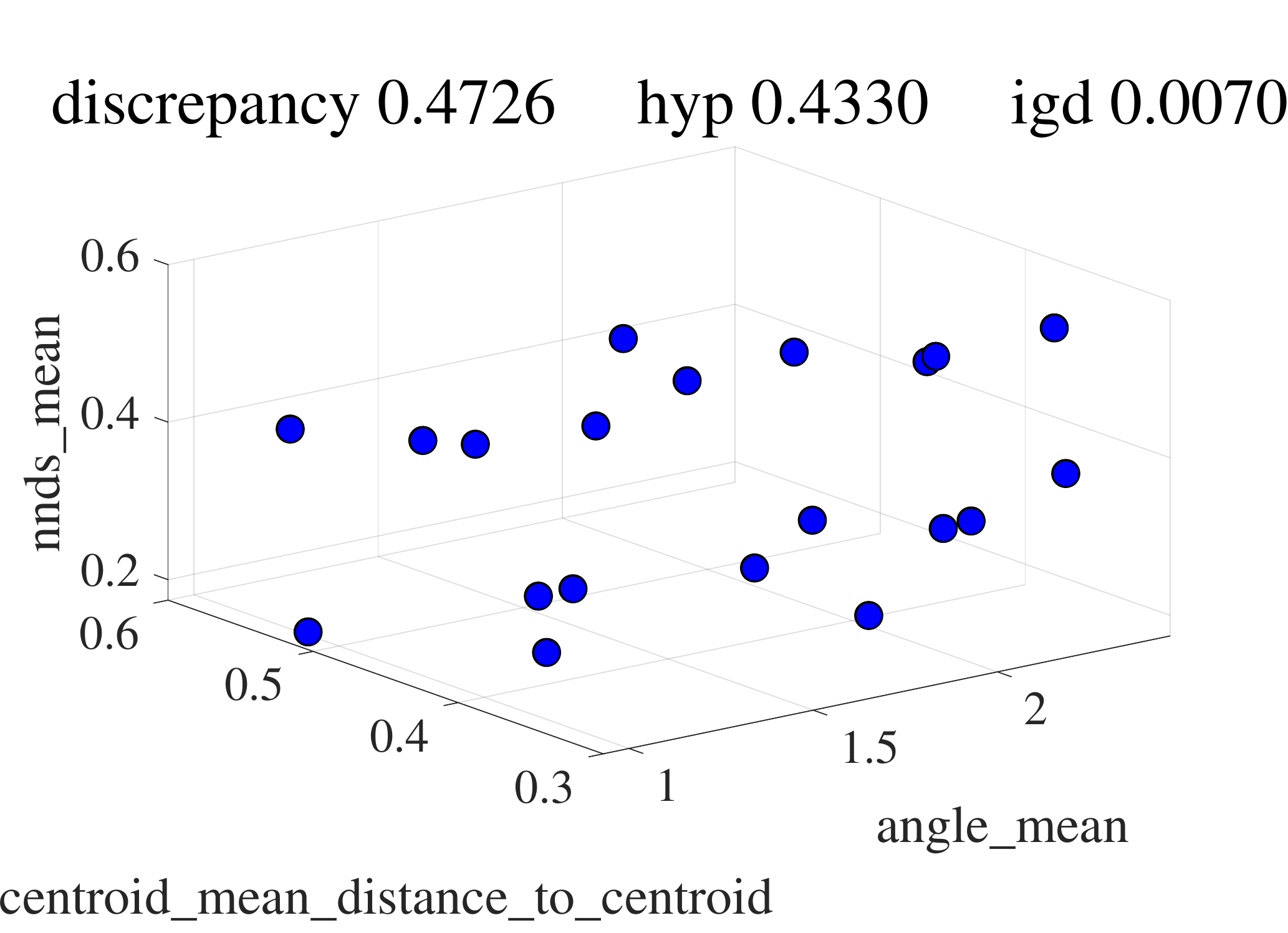}%
\includegraphics[trim={0cm 7.7cm 0cm 6.8cm},clip,width=0.32\textwidth]{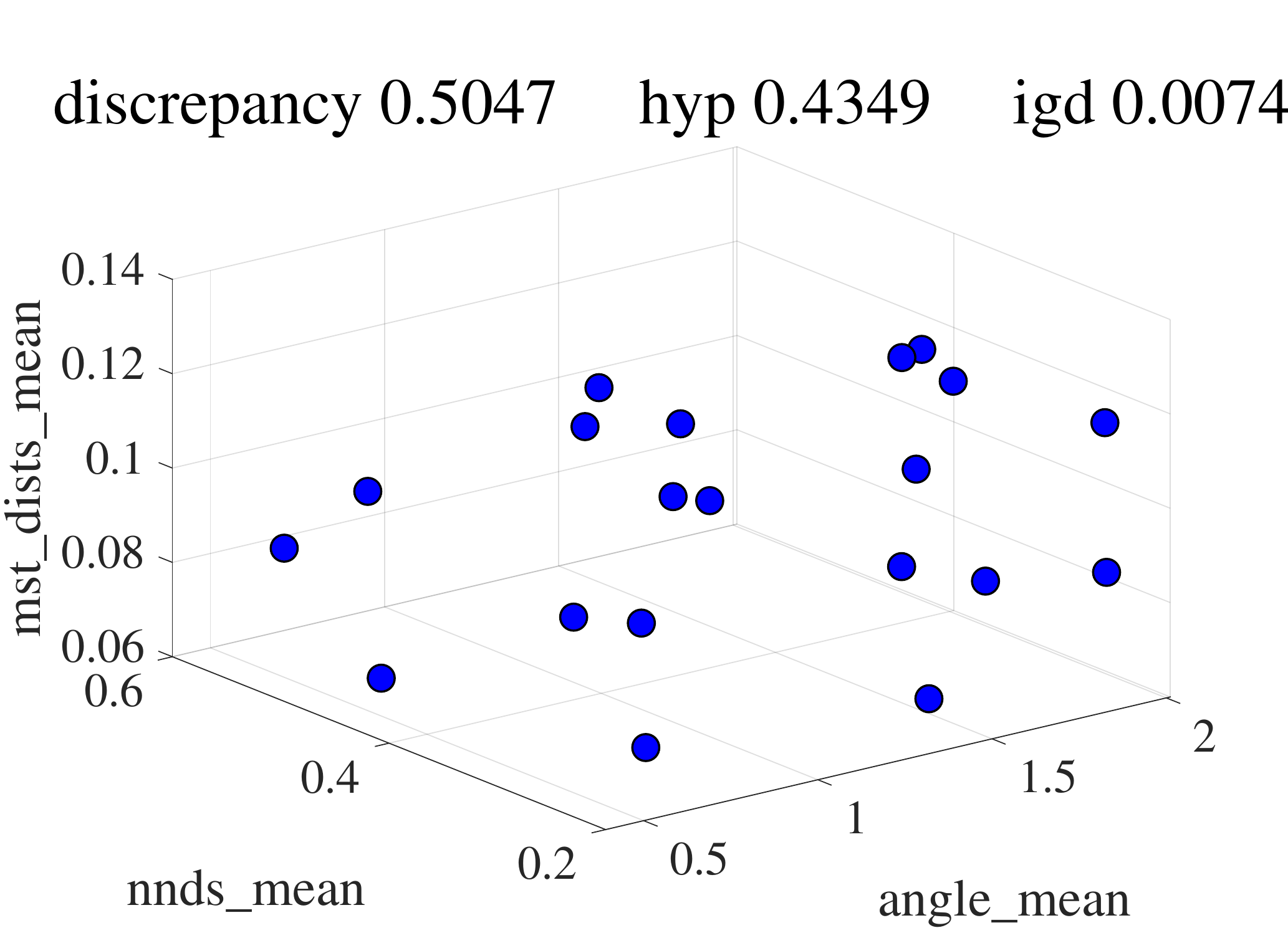}%
\includegraphics[trim={0cm 7.7cm 0cm 6.8cm},clip,width=0.32\textwidth]{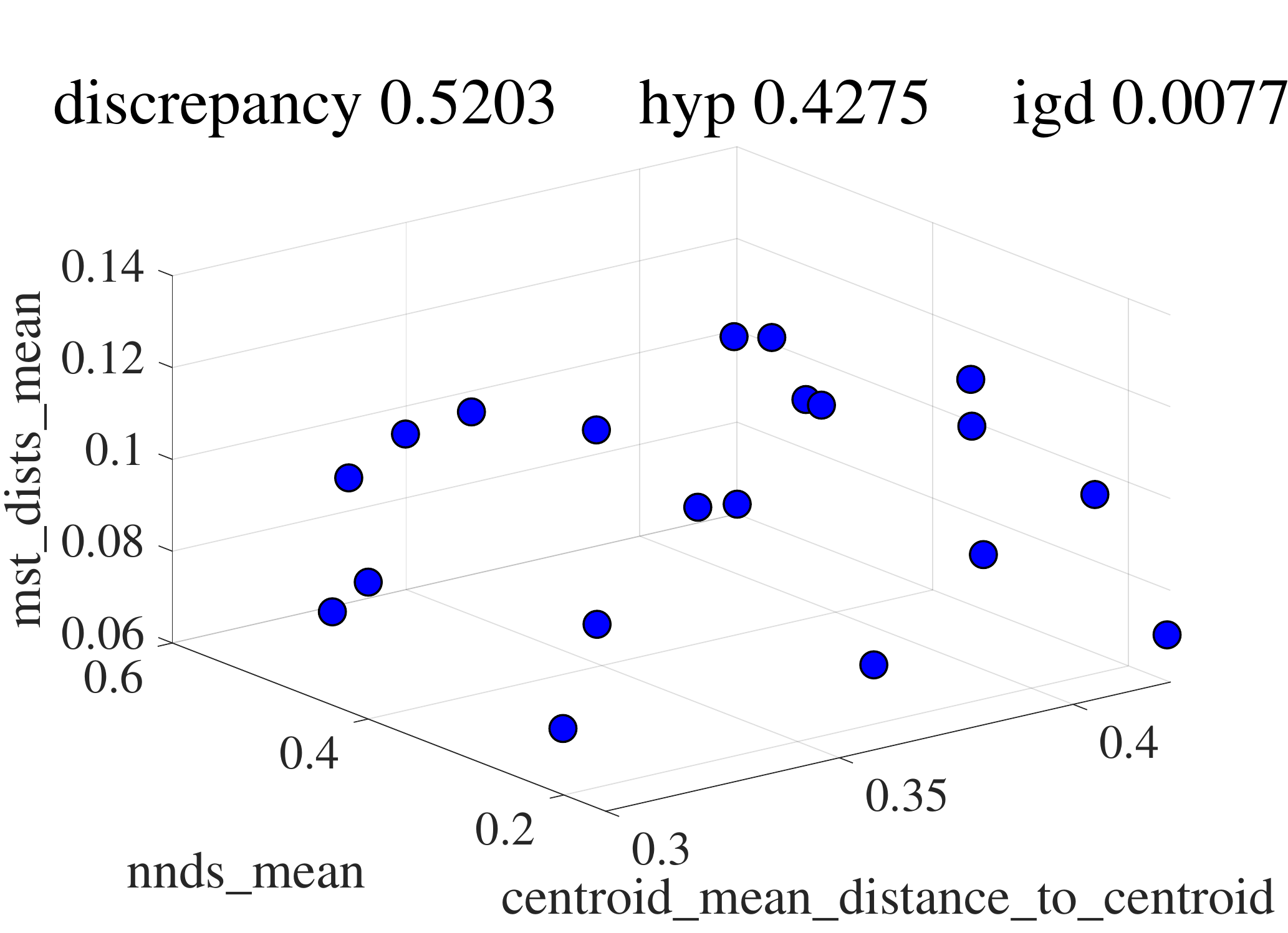}%
\vspace{0.3cm}
\rotatebox{90}{\hspace{11mm}EA$_{\text{DIS}}$} \rotatebox{90}{\rule{30mm}{1pt}} 
\includegraphics[trim={0cm 7.7cm 0cm 6.8cm},clip,width=0.32\textwidth]{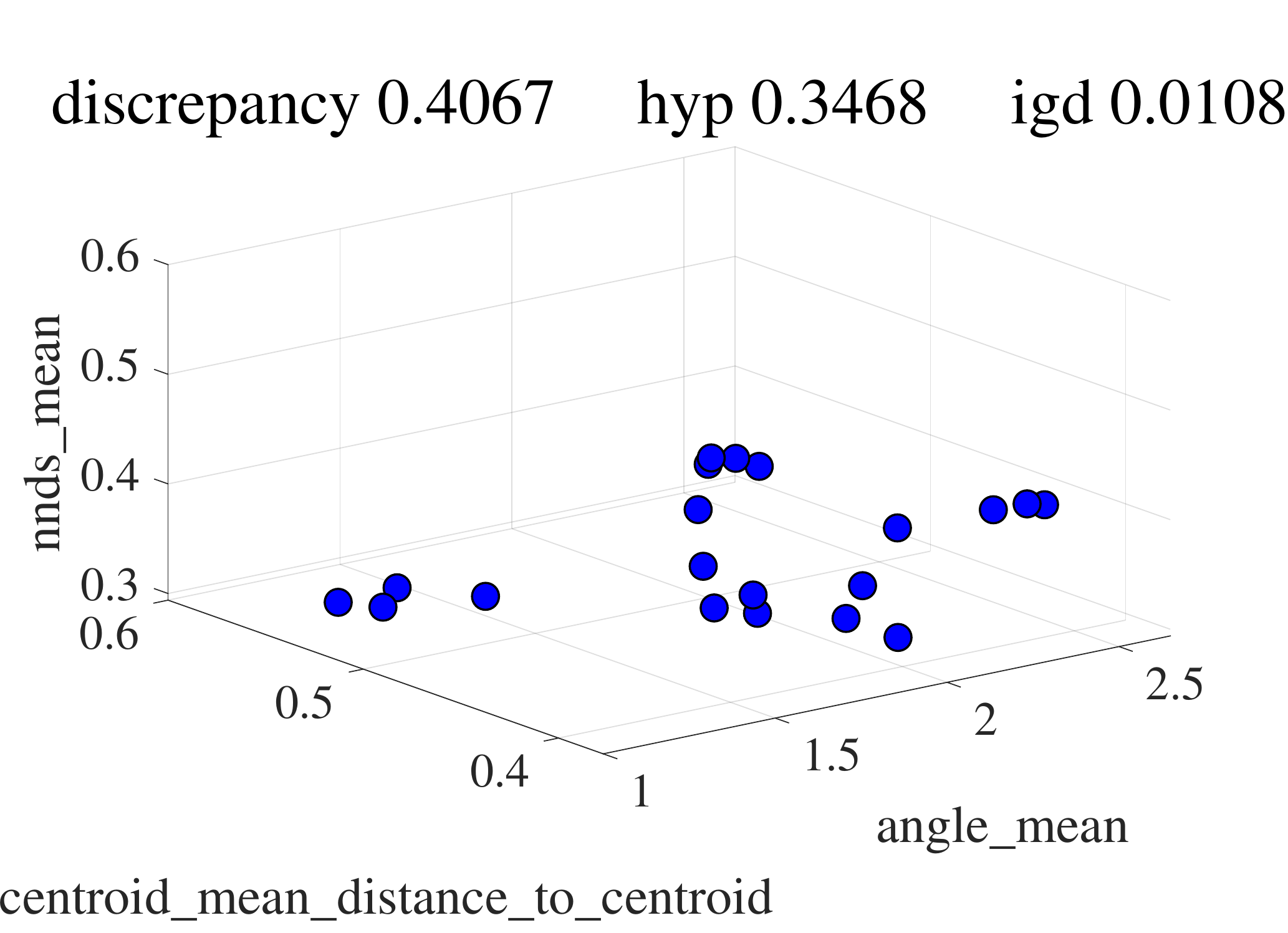}%
\includegraphics[trim={0cm 7.7cm 0cm 6.8cm},clip,width=0.32\textwidth]{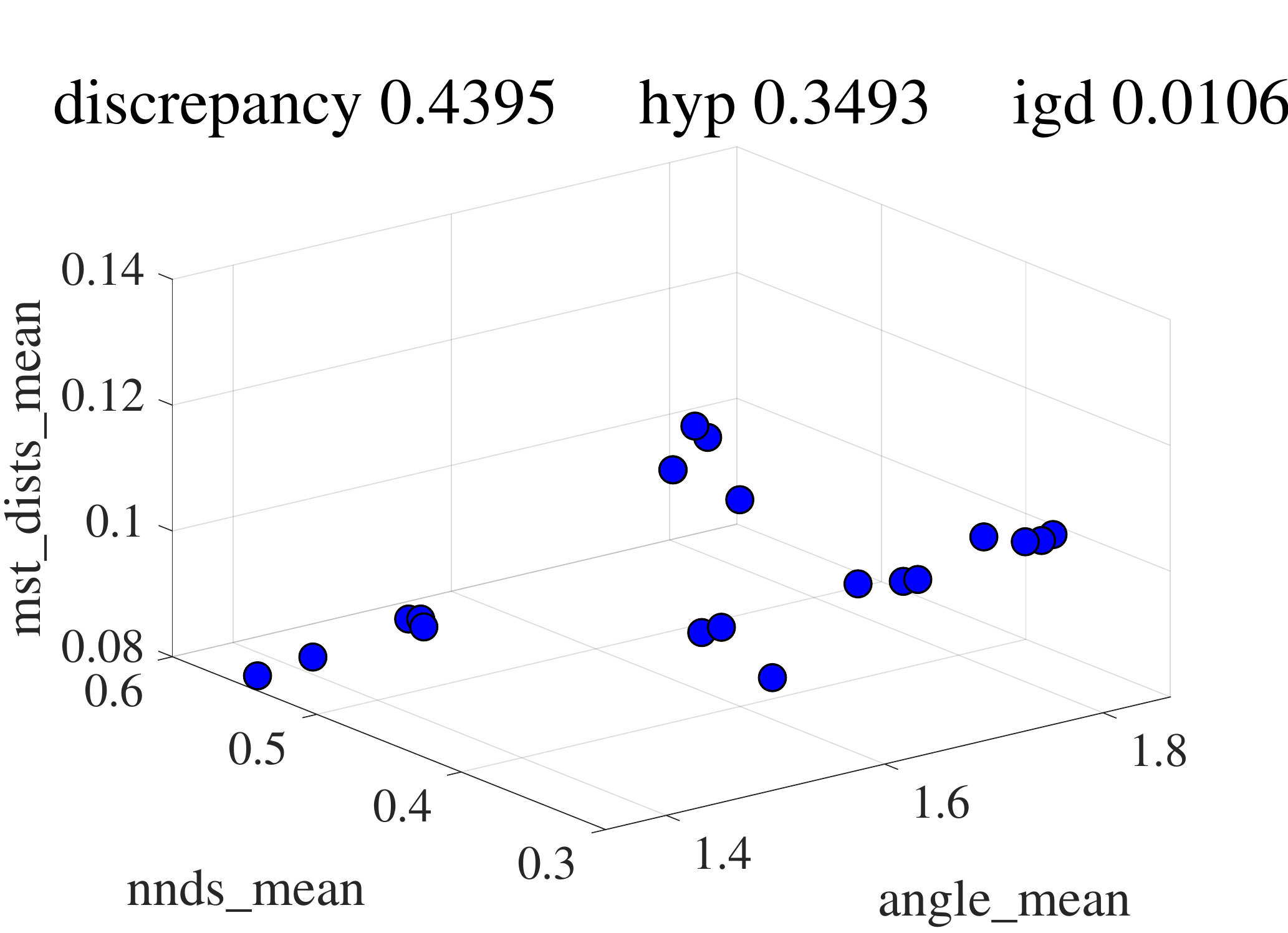}%
\includegraphics[trim={0cm 7.7cm 0cm 6.8cm},clip,width=0.32\textwidth]{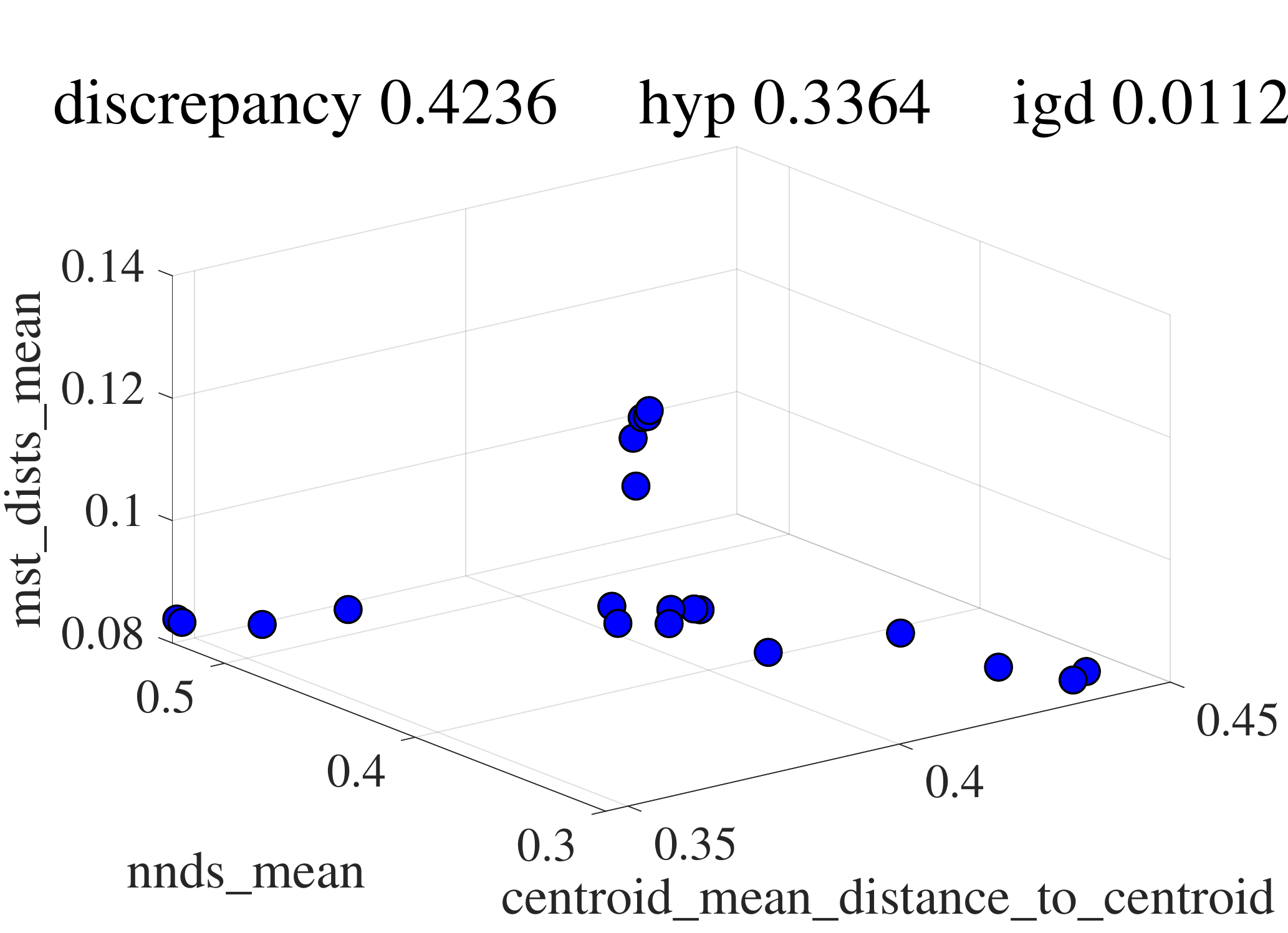}%
\caption{Feature vectors for final population of EA$_{\text{HYP}}$ (top), EA$_{\text{IGD}}$ and EA$_{\text{DIS}}$ (bottom) for TSP instances based on three features from left to right: ($f_1$, $f_2$, $f_3$), ($f_1$, $f_3$, $f_4$), ($f_2$, $f_3$, $f_4$).}
\label{fig:tsp-3d}
\end{figure}

\begin{table}[!t]
\centering
\caption{Investigations for TSP instances with $3$ features. 
Comparison in terms of mean, standard deviation and statistical test for considered indicators.}
\label{tb:tsp-stat-3d}
\vspace{4mm}
\renewcommand*{\arraystretch}{1.3}\setlength{\tabcolsep}{1.2mm}\resizebox{1.0\textwidth}{!}{

\begin{tabular}{lclllllllll}
                     &                            & \multicolumn{3}{c}{EA$_{\text{HYP}}$ (1)}                                                                     & \multicolumn{3}{c}{EA$_{\text{IGD}}$ (2)}                                                                     & \multicolumn{3}{c}{EA$_{\text{DIS}}$ (3)}                                                                                       \\
                     &                            & \multicolumn{1}{c}{mean} & \multicolumn{1}{c}{st} & \multicolumn{1}{c}{stat}                           & \multicolumn{1}{c}{mean} & \multicolumn{1}{c}{st} & \multicolumn{1}{c}{stat}                           & \multicolumn{1}{c}{mean} & \multicolumn{1}{c}{st} & \multicolumn{1}{c}{stat}          \\ 
\cline{3-11} 
\multicolumn{1}{l}{\multirow{3}{*}{\rotatebox{90}{HYP}}} & \multicolumn{1}{l|}{$f_1$,$f_2$,$f_3$} & 0.4511 & 1E-2 & \multicolumn{1}{l|}{$2^{(+)}$,$3^{(+)}$} & 0.4261 & 7E-3 & \multicolumn{1}{l|}{$1^{(-)}$,$3^{(+)}$} & 0.3385 & 6E-3 & $1^{(-)}$,$2^{(-)}$\\

\multicolumn{1}{l}{} & \multicolumn{1}{l|}{$f_1$,$f_3$,$f_4$} & 0.4579 & 8E-3 & \multicolumn{1}{l|}{$2^{(+)}$,$3^{(+)}$} & 0.4260 & 6E-3 & \multicolumn{1}{l|}{$1^{(-)}$,$3^{(+)}$} & 0.3430 & 6E-3 & $1^{(-)}$,$2^{(-)}$ \\

\multicolumn{1}{l}{} & \multicolumn{1}{l|}{$f_2$,$f_3$,$f_4$} & 0.4478 & 8E-3 & \multicolumn{1}{l|}{$2^{(+)}$,$3^{(+)}$} & 0.4262 & 6E-3 & \multicolumn{1}{l|}{$1^{(-)}$,$3^{(+)}$} & 0.3430 & 6E-3 & $1^{(-)}$,$2^{(-)}$ \\ 

\cmidrule(l){2-11} 

\multicolumn{1}{l}{\multirow{3}{*}{\rotatebox{90}{IGD}}} & \multicolumn{1}{l|}
{$f_1$,$f_2$,$f_3$} & 0.0083 & 3E-4 & \multicolumn{1}{l|}{$2^{(-)}$,$3^{(+)}$} & 0.0075 & 2E-4 & \multicolumn{1}{l|}{$1^{(+)}$,$3^{(+)}$} & 0.0110 & 1E-4 & $1^{(-)}$,$2^{(-)}$\\

\multicolumn{1}{l}{} & \multicolumn{1}{l|}{$f_1$,$f_3$,$f_4$} & 0.0082 & 2E-4 & \multicolumn{1}{l|}{$2^{(-)}$,$3^{(+)}$} & 0.0077 & 1E-4 & \multicolumn{1}{l|}{$2^{(+)}$,$3^{(+)}$} & 0.0107 & 1E-4 &  $1^{(-)}$,$2^{(-)}$\\

\multicolumn{1}{l}{} & \multicolumn{1}{l|}{$f_2$,$f_3$,$f_4$} & 0.0086 & 2E-4 & \multicolumn{1}{l|}{$2^{(-)}$,$3^{(+)}$} & 0.0080 & 2E-2 & \multicolumn{1}{l|}{$2^{(+)}$,$3^{(+)}$} & 0.0112 & 8E-5 & $1^{(-)}$,$2^{(-)}$ \\

\cmidrule(l){2-11} 

\multicolumn{1}{l}{\multirow{3}{*}{\rotatebox{90}{DIS}}} & \multicolumn{1}{l|}{$f_1$,$f_2$,$f_3$} & 0.4115 & 3E-2 & \multicolumn{1}{l|}{$2^{(+)}$,$3^{(+)}$} & 0.4839 & 3E-2 & \multicolumn{1}{l|}{$1^{(-)}$,$3^{(-)}$} & 0.4399 & 2E-2 & $1^{(-)}$,$2^{(+)}$\\

\multicolumn{1}{l}{} & \multicolumn{1}{l|}{$f_1$,$f_3$,$f_4$} & 0.5220 & 4E-2 & \multicolumn{1}{l|}{$3^{(-)}$} & 0.5474 & 3E-2 & \multicolumn{1}{l|}{$3^{(-)}$} & 0.4757 & 2E-2 & $1^{(+)}$,$2^{(+)}$\\

\multicolumn{1}{l}{} & \multicolumn{1}{l|}{$f_2$,$f_3$,$f_4$} & 0.4669 & 3E-2 & \multicolumn{1}{l|}{$2^{(+)}$} & 0.5111 & 3E-2 & \multicolumn{1}{l|}{$1^{(-)}$,$3^{(-)}$} & 0.4667 & 2E-2 &  $2^{(+)}$ \\ 
\end{tabular}
}
\end{table}

\subsubsection{Three-feature combinations}

For three-feature combinations, the indicators examined are the hypervolume and the inverted generational distance. The results from optimizing these two indicators are compared with those from minimizing the discrepancy value. The statistics gathered from 30 repeated runs of each setting are included in Table~\ref{tb:tsp-stat-3d}.

The three-feature combinations under examination in this paper are the same as in~\cite{DBLP:journals/corr/abs-1802-05448}. The plots in Figure~\ref{fig:tsp-3d} show some (again randomly drawn) final populations in the feature space as examples. Compared to the figures obtained after minimizing discrepancy, those from minimizing IGD or maximizing HYP show better coverage of the whole feature space. The figures showing the final population from EA$_{\text{DIS}}$ often contain some clusters of points, which means the feature vectors are not very diverse. The discrepancy values in the examples from EA$_{\text{HYP}}$ are comparable or even smaller than those of the corresponding examples of EA$_{\text{DIS}}$. By observation, the sets of feature vectors obtained by EA$_{\text{IGD}}$ nicely spread out over the feature space even when the discrepancy values are not smaller than those from EA$_{\text{DIS}}$.

Table~\ref{tb:tsp-stat-3d} summarizes the indicator values of the final populations after running the three algorithms on the three three-feature combinations. Both of the IGD values and HYP values of the final populations from EA$_{\text{IGD}}$ and EA$_{\text{HYP}}$ are better than those from EA$_{\text{DIS}}$. Although both algorithms do not perform very well in minimizing discrepancy for most three-feature combinations, EA$_{\text{HYP}}$ is able to achieve a smaller average discrepancy value than EA$_{\text{DIS}}$ in feature combination ($f_1$,$f_3$,$f_4$) and a comparable average value in feature combination ($f_2$,$f_3$,$f_4$). The minimum discrepancy values obtained by EA$_{\text{HYP}}$ for the three different feature combinations are all smaller than the corresponding values from EA$_{\text{DIS}}$.

\section{Conclusions}
\label{sec:conclusions}

We have proposed a new approach for evolutionary diversity optimization. It bridges the areas of evolutionary diversity optimization and evolutionary multi-objective optimization and shows how techniques developed in evolutionary multi-objective optimization can be used to come up with diverse sets of solutions of high quality for a given single-objective problem.
Our investigations demonstrated that well-established multi-objective performance indicators can be used to achieve a good diversity of sets of solutions according to a given set of features. The advantages of our approaches are (i) their simplicity and (ii) the quality of diversity achieved as measured by the respective indicators. The best performing approaches use HYP or IGD as indicators. We have shown that they achieve excellent results in terms of all indicators and often even outperform the discrepancy-based approach~\cite{DBLP:journals/corr/abs-1802-05448} when measuring quality in terms of discrepancy, which is surprising as they are not tailored towards this measure. 

In this work, we concentrated on using popular multi-objective indicators in existing diversity optimization approaches.
For future work, it would be interesting to use popular evolutionary multi-objective approaches such as MOEA/D, IBEA or NSGA-II/III for evolutionary diversity optimization.

.

\bibliographystyle{abbrv}

\bibliography{references}

\begin{thebibliography}{10}

\bibitem{DBLP:conf/gecco/AlexanderKN17}
B.~Alexander, J.~Kortman, and A.~Neumann.
\newblock Evolution of artistic image variants through feature based diversity
  optimisation.
\newblock In {\em Genetic and Evolutionary Computation Conference, {GECCO}},
  pages 171--178. ACM, 2017.

\bibitem{Applegate02}
D.~Applegate, W.~Cook, S.~Dash, and A.~Rohe.
\newblock Solution of a min-max vehicle routing problem.
\newblock {\em INFORMS Journal on Computing}, 14(2):132--143, Apr. 2002.

\bibitem{DBLP:journals/tcs/BerghammerFN12}
R.~Berghammer, T.~Friedrich, and F.~Neumann.
\newblock Convergence of set-based multi-objective optimization, indicators and
  deteriorative cycles.
\newblock {\em Theor. Comput. Sci.}, 456:2--17, 2012.

\bibitem{Chand2015manyEmo}
S.~Chand and M.~Wagner.
\newblock Evolutionary many-objective optimization: A quick-start guide.
\newblock {\em Surveys in Operations Res. and Management Science}, 20(2):35 --
  42, 2015.

\bibitem{Corder09}
G.~W. Corder and D.~I. Foreman.
\newblock {\em {Nonparametric Statistics for Non-Statisticians: A Step-by-Step
  Approach}}.
\newblock Wiley, 2009.

\bibitem{deb2001a}
K.~Deb.
\newblock {\em {Multi-objective optimization using evolutionary algorithms}}.
\newblock Wiley, Chichester, UK, 2001.

\bibitem{DBLP:journals/tec/DebAPM02}
K.~Deb, S.~Agrawal, A.~Pratap, and T.~Meyarivan.
\newblock A fast and elitist multiobjective genetic algorithm: {NSGA-II}.
\newblock {\em {IEEE} Trans. Evolutionary Computation}, 6(2):182--197, 2002.

\bibitem{den2014investigating}
E.~den Heijer and A.~Eiben.
\newblock Investigating aesthetic measures for unsupervised evolutionary art.
\newblock {\em Swarm and Evolutionary Computation}, 16:52--68, 2014.

\bibitem{DobkinEM96}
D.~P. Dobkin, D.~Eppstein, and D.~P. Mitchell.
\newblock Computing the discrepancy with applications to supersampling
  patterns.
\newblock {\em ACM Trans. Graph.}, 15:354--376, 1996.

\bibitem{DBLP:journals/corr/GaoNN15}
W.~Gao, S.~Nallaperuma, and F.~Neumann.
\newblock Feature-based diversity optimization for problem instance
  classification.
\newblock {\em CoRR}, abs/1510.08568, 2015.

\bibitem{DBLP:conf/ppsn/GaoNN16}
W.~Gao, S.~Nallaperuma, and F.~Neumann.
\newblock Feature-based diversity optimization for problem instance
  classification.
\newblock In {\em Parallel Problem Solving from Nature, PPSN}, volume 9921 of
  {\em LNCS}, pages 869--879. Springer, 2016.

\bibitem{hughes2014computer}
J.~F. Hughes, A.~Van~Dam, J.~D. Foley, M.~McGuire, S.~K. Feiner, D.~F. Sklar,
  and K.~Akeley.
\newblock {\em Computer graphics: principles and practice}.
\newblock Pearson Education, 2014.

\bibitem{DBLP:journals/tec/LiDZK15}
K.~Li, K.~Deb, Q.~Zhang, and S.~Kwong.
\newblock An evolutionary many-objective optimization algorithm based on
  dominance and decomposition.
\newblock {\em {IEEE} Trans. Evolutionary Computation}, 19(5):694--716, 2015.

\bibitem{matkovic2005global}
K.~Matkovic, L.~Neumann, A.~Neumann, T.~Psik, and W.~Purgathofer.
\newblock Global contrast factor-a new approach to image contrast.
\newblock {\em Computational Aesthetics}, 2005:159--168, 2005.

\bibitem{Mersmann2013}
O.~Mersmann, B.~Bischl, H.~Trautmann, M.~Wagner, J.~Bossek, and F.~Neumann.
\newblock A novel feature-based approach to characterize algorithm performance
  for the traveling salesperson problem.
\newblock {\em Annals of Mathematics and Artificial Intelligence},
  69(2):151--182, Oct 2013.

\bibitem{DBLP:conf/evoW/NeumannAN17}
A.~Neumann, B.~Alexander, and F.~Neumann.
\newblock Evolutionary image transition using random walks.
\newblock In {\em Computational Intelligence in Music, Sound, Art and Design -
  6th International Conference, EvoMUSART}, volume 10198 of {\em Lecture Notes
  in Computer Science}, pages 230--245, 2017.

\bibitem{DBLP:journals/corr/abs-1802-05448}
A.~Neumann, W.~Gao, C.~Doerr, F.~Neumann, and M.~Wagner.
\newblock Discrepancy-based evolutionary diversity optimization.
\newblock In {\em Genetic and Evolutionary Computation Conference, {GECCO}},
  pages 991--998. {ACM}, 2018.

\bibitem{DBLP:conf/gecco/NeumannSCN17}
A.~Neumann, Z.~L. Szpak, W.~Chojnacki, and F.~Neumann.
\newblock Evolutionary image composition using feature covariance matrices.
\newblock In {\em Genetic and Evolutionary Computation Conference, {GECCO}},
  pages 817--824. {ACM}, 2017.

\bibitem{Nixon:2008:FEI:1571711}
M.~Nixon and A.~S. Aguado.
\newblock {\em Feature Extraction \& Image Processing, Second Edition}.
\newblock Academic Press, 2nd edition, 2008.

\bibitem{rManual}
{R Core Team}.
\newblock {\em R: A Language and Environment for Statistical Computing}.
\newblock R Foundation for Statistical Computing, Vienna, Austria, 2015.

\bibitem{DBLP:conf/gecco/RisiVHS09}
S.~Risi, S.~D. Vanderbleek, C.~E. Hughes, and K.~O. Stanley.
\newblock How novelty search escapes the deceptive trap of learning to learn.
\newblock In {\em Genetic and Evolutionary Computation Conference, {GECCO}},
  pages 153--160. {ACM}, 2009.

\bibitem{DBLP:books/sp/StanleyL15}
K.~O. Stanley and J.~Lehman.
\newblock {\em Why Greatness Cannot Be Planned - The Myth of the Objective}.
\newblock Springer, 2015.

\bibitem{Thimard2001AnAT}
E.~Thi{\'e}mard.
\newblock An algorithm to compute bounds for the star discrepancy.
\newblock {\em J. Complexity}, 17:850--880, 2001.

\bibitem{DBLP:conf/gecco/UlrichBZ10}
T.~Ulrich, J.~Bader, and E.~Zitzler.
\newblock Integrating decision space diversity into hypervolume-based
  multiobjective search.
\newblock In {\em Genetic and Evolutionary Computation Conference, {GECCO}},
  pages 455--462. {ACM}, 2010.

\bibitem{DBLP:conf/gecco/UlrichT11}
T.~Ulrich and L.~Thiele.
\newblock Maximizing population diversity in single-objective optimization.
\newblock In {\em Genetic and Evolutionary Computation Conference, {GECCO}},
  pages 641--648. ACM, 2011.

\bibitem{Wagner2015ageejor}
M.~Wagner, K.~Bringmann, T.~Friedrich, and F.~Neumann.
\newblock Efficient optimization of many objectives by approximation-guided
  evolution.
\newblock {\em European Journal of Operational Research}, 243(2):465 -- 479,
  2015.

\bibitem{DBLP:journals/tec/ZhangL07}
Q.~Zhang and H.~Li.
\newblock {MOEA/D:} {A} multiobjective evolutionary algorithm based on
  decomposition.
\newblock {\em {IEEE} Trans. Evolutionary Computation}, 11(6):712--731, 2007.

\bibitem{DBLP:conf/ppsn/ZitzlerK04}
E.~Zitzler and S.~K{\"{u}}nzli.
\newblock Indicator-based selection in multiobjective search.
\newblock In {\em Parallel Problem Solving from Nature, PPSN}, volume 3242 of
  {\em LNCS}, pages 832--842. Springer, 2004.

\bibitem{DBLP:journals/tec/ZitzlerTLFF03}
E.~Zitzler, L.~Thiele, M.~Laumanns, C.~M. Fonseca, and V.~G. da~Fonseca.
\newblock Performance assessment of multiobjective optimizers: an analysis and
  review.
\newblock {\em {IEEE} Trans. Evolutionary Computation}, 7(2):117--132, 2003.

\end{thebibliography}

\end{document}